\documentclass{article}

\PassOptionsToPackage{numbers, compress}{natbib}

\usepackage[main, final]{neurips_2026}

\usepackage[T1]{fontenc}
\usepackage{hyperref}
\usepackage{booktabs}
\usepackage{microtype}
\usepackage{multirow}
\usepackage[table]{xcolor}
\usepackage{graphicx}
\usepackage{pifont}
\usepackage{amsmath}
\usepackage{wrapfig}
\usepackage{tabularx}
\usepackage{amssymb}
\usepackage{caption}
\usepackage{outlines}

\definecolor{prolaSky}{RGB}{225,242,250}
\definecolor{prolaSection}{RGB}{240,240,240}

\newcommand{\ftcell}[1]{\textcolor{gray}{#1}}
\newcommand{\cmark}{\ding{51}}
\newcommand{\prolaStackStat}[2]{\raisebox{\dimexpr .5\depth-.5\height+.5ex\relax}{\shortstack[r]{#1\\[-1pt]\scriptsize\ensuremath{\pm}\,#2}}}
\newcommand{\prolaStackBest}[2]{\prolaStackStat{\textbf{#1}}{#2}}
\newcommand{\abcheck}{\ensuremath{\checkmark}}
\newcommand{\prolaMeanStd}[2]{#1\nobreak\hspace{0.08em}\scalebox{0.80}{\ensuremath{\pm}\,#2}}
\newcommand{\prolaBestMeanStd}[2]{\prolaMeanStd{\textbf{#1}}{#2}}

\title{PROLA: Principal-Orthogonal Low-rank Adaptation for Predictive Spatiotemporal Weather Downscaling}

\author{%
  \textbf{Minseo Yoon} \quad
  \textbf{Minseong Bae} \quad
  \textbf{Sojin Lee} \quad
  \textbf{Hyunwoo J.~Kim}\thanks{Corresponding author.} \\
  Korea Advanced Institute of Science and Technology (KAIST) \\
  \texttt{\{yms020615, bms2002, sojin.lee, hyunwoojkim\}@kaist.ac.kr}
}

\begin{document}

\maketitle

\begin{abstract}
High-resolution weather prediction is important for resolving local atmospheric patterns missed by coarse global forecasts.
Large pretrained weather backbones have improved global forecasting, but adapting them to produce future high-resolution fields from recent coarse states remains challenging and costly.
We formulate this setting as Predictive Spatiotemporal Weather Downscaling (\textbf{PSWD}), a task that jointly refines spatial and temporal resolution while predicting future high-resolution trajectories.
To this end, we propose a framework that decomposes prediction into spatial and temporal stages using a shared pretrained backbone, fixed numerical scaffolds, and lightweight residual heads.
For backbone adaptation, we introduce \textbf{PROLA} (PRincipal-Orthogonal Low-rank Adaptation), which allocates a fixed low-rank budget between pretrained weights' principal directions and their orthogonal complement.
PROLA further rescales rank-1 optimizer updates using gradient signals, improving adaptation without increasing trainable rank.
Experiments in the PSWD setting show that PROLA outperforms representative low-rank adaptation baselines.
Further experiments across diverse datasets and tasks support the broader applicability of PROLA, showing that its benefits extend beyond the primary weather downscaling setting.
\end{abstract}

\section{Introduction}

Machine learning-based weather modeling has primarily developed along two paradigms: weather forecasting~\citep{GraphCast:lam2023learning,bi2023accurate,lang2024aifs,bodnar2025foundation}, which predicts future states at fixed spatial resolution, and weather downscaling~\citep{vandal2017deepsd,oyama2023deep,glawion2025global,mardani2025residual}, which refines either spatial resolution at a fixed time or temporal resolution at a fixed grid.
These standard formulations typically emphasize a subset of three axes: future prediction, spatial refinement, and temporal refinement.
However, real-world applications that forecast a high-resolution target from a different, lower-resolution modality naturally involve all three axes.
For example, minute-scale wind-power forecasting for renewable integration~\citep{wurth2019minute},
sub-hourly precipitation nowcasting for flood-risk monitoring~\citep{ravuri2021skillful},
and fine-scale aviation turbulence guidance for flight safety~\citep{sharman2014description}
all require future predictions at fine spatial and temporal resolution,
a regime that no prior work has explored.
This raises a natural question:
\textit{Can a single model jointly refine spatial and temporal resolution while extending predictions into the future?}

To answer this, we introduce \textbf{Predictive Spatiotemporal Weather Downscaling} (PSWD), the task of mapping past low-resolution states to future high-resolution states, with resolution gaps in both space and time.
PSWD formalizes joint spatiotemporal refinement and future prediction as a unified learning task.
It extends task-specific adaptation benchmarks for weather foundation models (WFMs)~\citep{cao2026taskadaptive} to a setting that requires all three capabilities jointly.
PSWD requires a model to infer fine-scale spatial structure and its evolution from the same coarse history. We address this task with a \textit{scaffold-and-correct} pipeline built around a shared pretrained weather backbone. The spatial stage predicts and refines a high-resolution state at the final lead time. The temporal stage then uses this predicted endpoint and the upsampled present state to construct the target trajectory. Fixed spatial upsampling and temporal interpolation provide the scaffolds, while lightweight correction heads learn the residual details.

In the scaffold-and-correct pipeline, the shared backbone provides the forecast that anchors both spatial and temporal refinement.
To align this forecast with the final trajectory objective, we adapt the backbone jointly with the correction heads.
Since the backbone already contains useful forecasting knowledge, this adaptation should build on its pretrained structure.
At the same time, predicting high-resolution future trajectories may require corrections beyond the dominant pretrained directions.
A single low-rank update provides a compact parameterization for parameter-efficient fine-tuning (PEFT)~\citep{hu2022lora}, but it does not explicitly separate these two needs under a fixed rank budget.
This motivates an adaptation design that initializes one update path from the pretrained singular structure~\citep{meng2024pissa} while reserving another path for corrections outside the selected principal directions~\citep{xiong2026oplora}.

To this end, we propose \textbf{PROLA} (\textbf{PR}incipal-\textbf{O}rthogonal \textbf{L}ow-rank \textbf{A}daptation), a dual-branch low-rank adaptation method for the shared backbone.
PROLA divides the fixed rank budget between a principal branch initialized from dominant singular components and an orthogonal branch projected outside the selected principal subspaces.
The principal factors remain freely trainable after initialization, while the orthogonal branch uses fixed projectors defined by the pretrained weight.
Beyond this allocation, PROLA rescales rank-1 optimizer updates using gradient-based branch and component scores, without increasing trainable rank.
Branch scores adjust the relative update strength between the two paths, while component scores modulate individual updates within each path.

Experiments on WeatherBench~2~\citep{rasp2024weatherbench} with the pretrained WFM backbone~\citep{bodnar2025foundation} show that PROLA improves over representative low-rank PEFT baselines under matched low-rank budgets for multi-variable, multi-lead global weather downscaling.
Ablations and diagnostics further support the scaffold-and-correct pipeline, the dual-branch design of PROLA, and gradient-guided scaling of optimizer updates.
Together, these results support PSWD as a challenging evaluation setting and PROLA as an effective adaptation method for pretrained weather backbones, with broader applicability across datasets and tasks.

Our contributions are fourfold:
\begin{itemize}
    \item We formulate \textbf{Predictive Spatiotemporal Weather Downscaling} (PSWD) to predict future high-resolution weather trajectories from recent coarse-resolution states, and introduce a \textbf{scaffold-and-correct} pipeline to address it.

    \item We propose \textbf{PROLA} (\textbf{PR}incipal-\textbf{O}rthogonal \textbf{L}ow-rank \textbf{A}daptation), a low-rank PEFT method that allocates a fixed rank budget between principal and orthogonal branches and rescales optimizer updates using gradient-based component scores.

    \item We show that PROLA outperforms representative PEFT baselines under a matched low-rank budget and narrows the performance gap between low-rank PEFT and full fine-tuning in our PSWD setting.

    \item We demonstrate PROLA's effectiveness across datasets and the transferability of its adaptation design beyond weather downscaling.

\end{itemize}

\section{Predictive Spatiotemporal Weather Downscaling}

In Section~\ref{sec:task}, we first formulate predictive spatiotemporal weather downscaling (PSWD) as a challenging task that jointly requires forecasting, spatial downscaling, and temporal downscaling.
We then introduce a scaffold-and-correct pipeline designed to address this task in Section~\ref{sec:pipeline}.

\subsection{Task Formulation}
\label{sec:task}

\textbf{Notation.}
Let \(\mathbf{x}\) denote a context input and \(\mathbf{y}\) a prediction target.
For an atmospheric state with \(V\) variable channels, the input on the low-resolution grid is represented by
\(\mathbf{x}^{\mathrm{LR}}_t \in \mathbb{R}^{V\times H_{\mathrm{LR}}\times W_{\mathrm{LR}}}\)
and the target on the high-resolution grid by
\(\mathbf{y}^{\mathrm{HR}}_t \in \mathbb{R}^{V\times H_{\mathrm{HR}}\times W_{\mathrm{HR}}}\).
Let \(\mathcal{H}_c\) denote the set of input lags at the coarse sampling interval \(\delta_c\), and \(\mathcal{H}_f\) the set of target leads at the fine sampling interval \(\delta_f\).
Relative to the anchor time \(t\), the input at lag \(h\in\mathcal{H}_c\) is \(\mathbf{x}^{\mathrm{LR}}_{t-h}\), and the target at lead \(h\in\mathcal{H}_f\) is \(\mathbf{y}^{\mathrm{HR}}_{t+h}\).
With this indexing, \(\mathcal{H}_c\) consists of the lags in \(\{k\delta_c \mid k\in\mathbb{Z}_{\ge 0}\}\) within the context window, and \(\mathcal{H}_f\) consists of the leads in \(\{k\delta_f \mid k\in\mathbb{Z}_{>0}\}\) within the forecast window.

\begin{figure*}[t]
    \centering
    \includegraphics[width=0.95\textwidth]{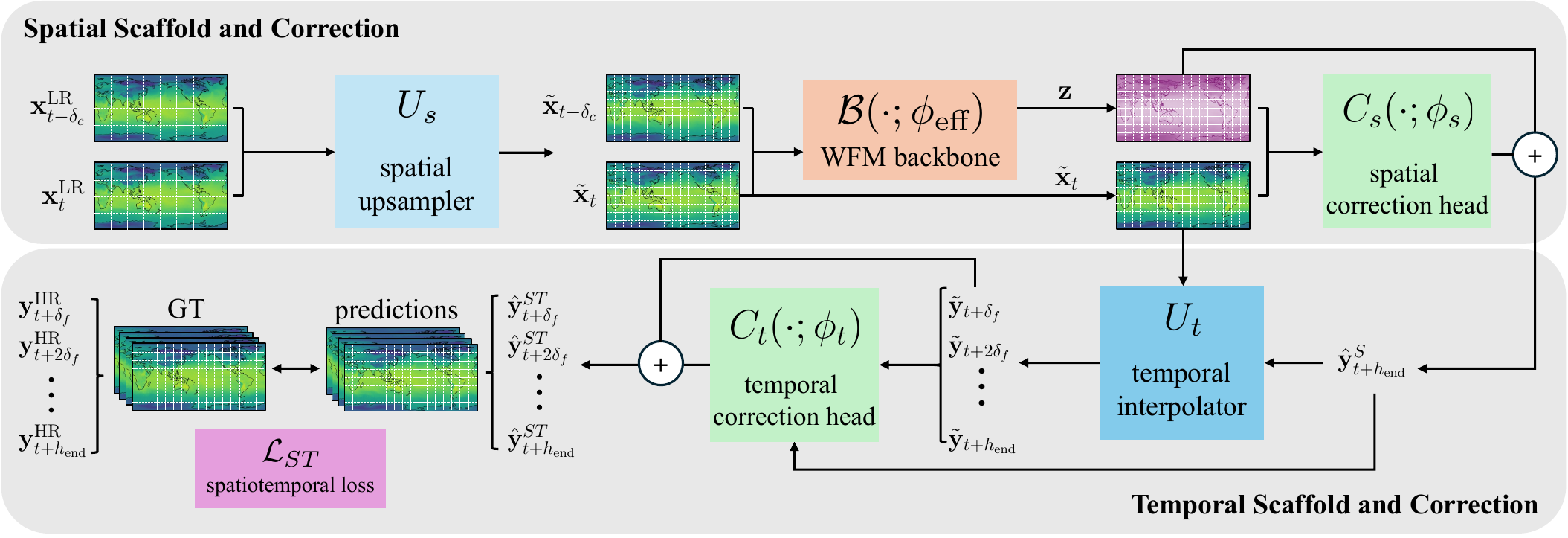}
    \caption{\textbf{Overview of the proposed scaffold-and-correct pipeline for PSWD.}
    Fixed numerical operators \(U_s\) and \(U_t\) and WFM backbone \(\mathcal{B}\) build spatial and temporal scaffolds, while the residual heads \(C_s\) and \(C_t\) provide learned corrections.}
    \label{fig:overview_scaffold}
\end{figure*}

\textbf{Comparison with existing weather tasks.}
We compare PSWD with related weather problem settings along three axes: forecasting, spatial downscaling and temporal downscaling.
\textbf{\emph{Forecasting}} predicts future states from past observations, changing only the forecast horizon while typically keeping the spatial and temporal grids fixed.
\textbf{\emph{Spatial downscaling}} reconstructs an HR state from its LR counterpart at the same valid time,
\(\mathbf{x}^{\mathrm{LR}}_t \mapsto \mathbf{y}^{\mathrm{HR}}_t\).
It refines spatial resolution by recovering unresolved local structure, but does not change the temporal horizon.
\textbf{\emph{Temporal downscaling}} maps a coarsely sampled trajectory to a denser one on the same spatial grid over the same time interval.
Existing settings usually address a single axis, whereas our target setting, \textbf{\emph{Predictive Spatiotemporal Weather Downscaling} (PSWD)}, combines all three.
At forecast initialization time \(t\), PSWD takes coarse-step past and present LR inputs and predicts fine-step future HR targets:
\begin{equation}
\mathcal{X}^\mathrm{LR}_t
=
(\mathbf{x}^\mathrm{LR}_{t-h})_{h\in\mathcal{H}_c},
\qquad
\mathcal{Y}^\mathrm{HR}_t
=
(\mathbf{y}^\mathrm{HR}_{t+h})_{h\in\mathcal{H}_f},
\end{equation}
where both sequences are ordered by increasing valid time and the targets are sampled more frequently than the inputs, \textit{i.e.,} \(\delta_f<\delta_c\).
Thus, PSWD requires learning the mapping \(\mathcal{X}^\mathrm{LR}_t \mapsto \mathcal{Y}^\mathrm{HR}_t\) to predict a future, higher-resolution, and temporally denser target trajectory.

\textbf{Why PSWD is challenging.}
\emph{(i) Coupled spatiotemporal prediction.}
PSWD combines three task axes that are typically handled separately.
Each target state is simultaneously a future prediction, a spatially refined field, and a temporally denser sub-step.
The model must therefore infer fine-resolution future trajectories from coarse past inputs, making PSWD a coupled spatiotemporal prediction problem rather than a single-axis task shift.
\emph{(ii) Limited adaptation capacity.}
While pretrained weather foundation models (WFMs) provide useful large-scale forecasting knowledge, PSWD requires adapting them to a richer output regime with finer spatial grids and denser future time steps.
Since modern WFMs are increasingly large, fully fine-tuning them for every target resolution, lead-time structure, or data regime is often impractical.
Recent work~\citep{cao2026taskadaptive} therefore emphasizes PEFT as an important paradigm for WFM adaptation.
PSWD tests whether limited trainable capacity can support a coupled shift in forecast horizon, spatial resolution, and temporal density.

\subsection{Scaffold-and-Correct Pipeline}
\label{sec:pipeline}

PSWD requires the model to forecast future atmospheric states while downscaling both spatially and temporally.
We therefore introduce a unified \emph{scaffold-and-correct} pipeline in Figure~\ref{fig:overview_scaffold} that organizes prediction into spatial and temporal stages.

\paragraph{Overview.}
We adopt a \emph{scaffold-and-correct} architecture: a coarse scaffold built primarily from pretrained or fixed components, followed by a learnable correction head that adds the missing detail.
First, a fixed spatial upsampler $U_s$ lifts coarse low-resolution inputs to the target HR grid, and a PEFT-adapted WFM backbone $\mathcal{B}$ predicts a spatial scaffold $\mathbf{z}$ at the fixed final lead time $h_{\mathrm{end}}=\max\mathcal{H}_f$.
Then, the spatial correction head $C_s$ adds a learned residual to produce the refined endpoint.
A fixed temporal interpolator $U_t$ bridges the upsampled present input and the predicted spatial endpoint to form temporal scaffolds $\tilde{\mathbf{y}}_{t+h}$ over the target states at $h\in\mathcal{H}_f$, and the temporal correction head $C_t$ adds another residual to produce the final outputs.
The entire architecture shares a single backbone and is trained end-to-end under a single loss.
The superscripts $S$ and $ST$ denote the spatial-stage and final spatiotemporal predictions, respectively.

\paragraph{Spatial scaffold and correction.}
Let $\phi_{\mathrm{eff}}=(\phi_{\mathcal{B}},\phi_{\mathrm{PEFT}})$ collect the frozen pretrained backbone parameters and the trainable PEFT parameters, and let $\phi_s,\phi_t$ denote the parameters of the two correction heads $C_s,C_t$.
For the common two-input case $\mathcal{H}_c=\{\delta_c,0\}$, a fixed spatial upsampler $U_s$ (e.g., bilinear interpolation) aligns the coarse inputs to the target HR grid:
\begin{equation}
\tilde{\mathbf{x}}_{t-\delta_c}
=
U_s(\mathbf{x}^\mathrm{LR}_{t-\delta_c}),
\qquad
\tilde{\mathbf{x}}_t
=
U_s(\mathbf{x}^\mathrm{LR}_t).
\end{equation}
The PEFT-adapted WFM backbone $\mathcal{B}$ produces a spatial scaffold $\mathbf{z}$ at the final lead time, and the spatial correction head $C_s$ adds a learned residual:
\begin{equation}
\mathbf{z}
=
\mathcal{B}(
\tilde{\mathbf{x}}_{t-\delta_c},
\tilde{\mathbf{x}}_t;
\phi_{\mathrm{eff}}
),
\qquad
\hat{\mathbf{y}}^{S}_{t+h_{\mathrm{end}}}
=
\mathbf{z}
+
C_s(\tilde{\mathbf{x}}_t,\mathbf{z};\phi_s).
\end{equation}
Here, $U_s$ transports large-scale synoptic structure to the target grid and $\mathcal{B}$ advances it to the final lead time using its pretrained forecasting dynamics.
Finally, $C_s$ supplies the sub-grid corrections that interpolation alone cannot recover.

\paragraph{Temporal scaffold and correction.}
The endpoint prediction $\hat{\mathbf{y}}^{S}_{t+h_{\mathrm{end}}}$ then anchors the temporal downscaling.
For each $h\in\mathcal{H}_f$, a fixed temporal interpolator $U_t$ (e.g., linear interpolation) between the upsampled present input $\tilde{\mathbf{x}}_t$ and the predicted endpoint $\hat{\mathbf{y}}^{S}_{t+h_{\mathrm{end}}}$ forms a temporal scaffold $\tilde{\mathbf{y}}_{t+h}$, and the lead-time-conditioned temporal correction head $C_t$ adds a residual:
\begin{equation}
\label{eq:temporal-stage}
\tilde{\mathbf{y}}_{t+h}
=
U_t(
\tilde{\mathbf{x}}_t,
\hat{\mathbf{y}}^{S}_{t+h_{\mathrm{end}}};
h
),
\qquad
\hat{\mathbf{y}}^{ST}_{t+h}
=
\tilde{\mathbf{y}}_{t+h}
+
C_t(
\tilde{\mathbf{y}}_{t+h},
\hat{\mathbf{y}}^{S}_{t+h_{\mathrm{end}}},
h;
\phi_t
).
\end{equation}
$U_t$ preserves the temporal continuity implied by the two endpoint states, while $C_t$ learns the sub-step deviations that linear interpolation cannot capture.

The full pipeline thus concentrates the adaptation burden in the WFM backbone $\mathcal{B}$ and the two lightweight correction heads $C_s$ and $C_t$, while $U_s$ and $U_t$ remain non-learnable scaffolds that preserve large-scale spatial and temporal structure.

\paragraph{Training objective.}
We train end-to-end with the variable-weighted state loss $\ell_{\mathrm{PT}}$ used to pretrain the backbone, applied uniformly to all trajectory outputs:
\begin{equation}
\label{eq:loss}
\mathcal{L}_{ST}
=
\frac{1}{|\mathcal{H}_f|}
\sum_{h\in\mathcal{H}_f}
\ell_{\mathrm{PT}}\!\left(
\hat{\mathbf{y}}^{ST}_{t+h},\,
\mathbf{y}^\mathrm{HR}_{t+h}
\right).
\end{equation}
No auxiliary loss is attached to the spatial endpoint $\hat{\mathbf{y}}^{S}_{t+h_{\mathrm{end}}}$: the spatial stage is supervised through gradients flowing back from the temporal stage, ensuring both stages co-adapt for the joint trajectory rather than optimizing toward stage-specific targets.
Section~\ref{sec:pipeline_ablation} evaluates these design choices.

\begin{figure*}[t]
    \centering

    \includegraphics[width=0.98\linewidth]{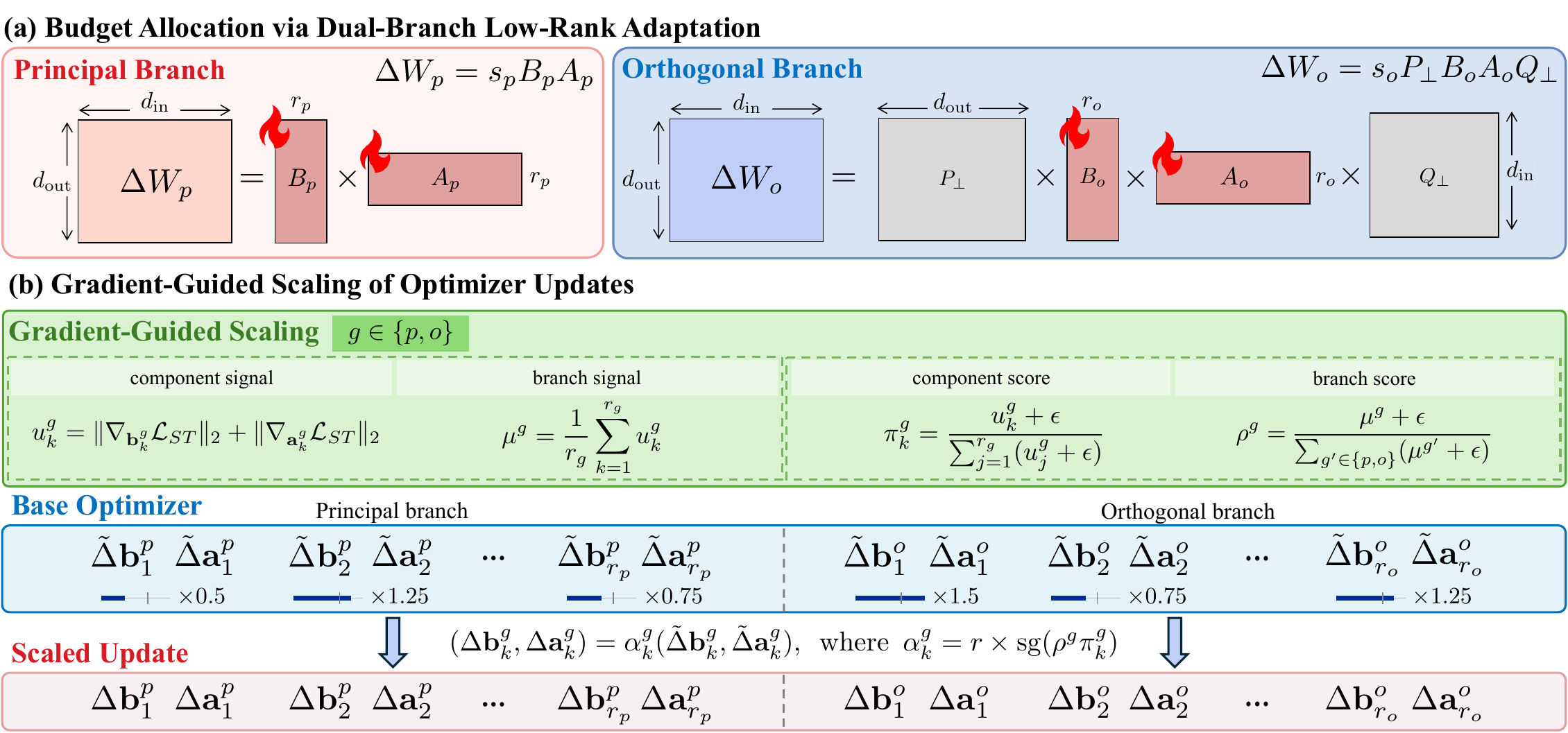}

    \caption{
    \textbf{Overview of PROLA. }
    \textbf{(a)} A fixed rank budget \(r=r_p+r_o\) is split into a principal branch
    \(\Delta W_p=s_pB_pA_p\), initialized from the top singular components of \(W_0\),
    and an orthogonal branch
    \(\Delta W_o=s_oP_\perp B_oA_oQ_\perp\), which captures corrections outside the selected principal subspaces.
    The effective weight combines the frozen residual weight and the two trainable branches as
    \(W^{\mathrm{eff}}=W^{\mathrm{res}}+\Delta W_p+\Delta W_o\).
    \textbf{(b)} Updates from the base optimizer are scaled using gradient-guided branch and component scores.
    For each factor pair, \(\alpha_k^g\) scales \((\tilde{\Delta}\mathbf b_k^g,\tilde{\Delta}\mathbf a_k^g)\), yielding \((\Delta\mathbf b_k^g,\Delta\mathbf a_k^g)\).
    }
    \label{fig:prola}

    \vspace{-2mm}
\end{figure*}

\section{PROLA: PRincipal-Orthogonal Low-rank Adaptation}

Low-rank PEFT reduces the number of trainable parameters, but under a fixed budget, performance depends on how the limited capacity is allocated in the pretrained weight space and rescaled during optimization.
PROLA addresses these two aspects through rank allocation and gradient-guided update scaling (Figure~\ref{fig:prola}).
In Section~\ref{sec:peft}, we allocate the rank budget between principal and orthogonal branches, with the former supporting pretrained-aligned adaptation and the latter capturing residual corrections outside the selected principal directions.
In Section~\ref{sec:step-allocation}, we further rescale updates from the base optimizer using rank-1 component scores computed from gradient magnitudes.
Together, these mechanisms determine where the limited rank budget is allocated and how strongly each component is updated, allowing PROLA to preserve pretrained structure while directing updates toward task-specific corrections.

\subsection{Budget Allocation via Dual-Branch Low-Rank Adaptation}
\label{sec:peft}

We first allocate the fixed low-rank budget in the pretrained weight space.
Given a pretrained weight matrix
\(W_0\in\mathbb R^{d_{\mathrm{out}}\times d_{\mathrm{in}}}\),
PROLA splits the total rank budget \(r=r_p+r_o\), with fixed hyperparameters \(r_p,r_o>0\),
between a principal branch for pretrained-aligned adaptation and an orthogonal branch
for corrections outside the selected principal directions~\cite{meng2024pissa,xiong2026oplora}.
The effective weight is
\begin{equation}
W^{\mathrm{eff}}
=
W^{\mathrm{res}}+\Delta W^{\mathrm{dual}},
\qquad
\Delta W^{\mathrm{dual}}
=
\underbrace{\Delta W_p}_{\text{principal update}}
+
\underbrace{\Delta W_o}_{\text{orthogonal update}},
\end{equation}
where \(W^{\mathrm{res}}\) is a frozen residual weight defined in
Eq.~\eqref{eq:residual_weight}.
The two branches contain \(r(d_{\mathrm{out}}+d_{\mathrm{in}})\)
trainable parameters, matching a rank-\(r\) adapter.

\textbf{Principal branch.}
We compute the SVD \(W_0=U\Sigma V^\top\) and select the top-\(r_p\)
singular directions \(U_p,V_p\) and singular values \(\Sigma_p\).
The principal update is
\begin{equation}
\Delta W_p=s_pB_pA_p,
\end{equation}
where \(B_p\in\mathbb R^{d_{\mathrm{out}}\times r_p}\),
\(A_p\in\mathbb R^{r_p\times d_{\mathrm{in}}}\), and \(s_p>0\) is a
scaling coefficient.
We initialize \(B_{p,0}=U_p\Sigma_p^{1/2}\) and
\(A_{p,0}=\Sigma_p^{1/2}V_p^\top\).
After initialization, both factors are optimized without additional constraints.

\textbf{Orthogonal branch.}
The remaining rank \(r_o\) is assigned to corrections outside the selected
principal directions.
With \(P_\perp=I_{d_{\mathrm{out}}}-U_pU_p^\top\) and
\(Q_\perp=I_{d_{\mathrm{in}}}-V_pV_p^\top\), we define
\begin{equation}
\Delta W_o=s_oP_\perp B_oA_oQ_\perp,
\end{equation}
where \(B_o\in\mathbb R^{d_{\mathrm{out}}\times r_o}\),
\(A_o\in\mathbb R^{r_o\times d_{\mathrm{in}}}\), and \(s_o>0\) is a
scaling coefficient.
By construction, \(U_p^\top\Delta W_o=0\) and \(\Delta W_oV_p=0\),
so this branch provides dedicated capacity outside both principal subspaces.

\textbf{Function-preserving initialization.}
Following PiSSA~\cite{meng2024pissa}, we residualize the frozen weight:
\begin{equation}
W^{\mathrm{res}}
=
W_0-\Delta W_{p,0},
\qquad
\Delta W_{p,0}
=
s_pB_{p,0}A_{p,0}.
\label{eq:residual_weight}
\end{equation}
We set \(B_{o,0}=0\) and sample the entries of \(A_{o,0}\) from
\(\mathcal N(0,\sigma_o^2)\), where \(\sigma_o\) controls the initialization scale.
This gives \(\Delta W_{o,0}=0\) and
\(W^{\mathrm{eff}}_0=W_0\), preserving the pretrained function.

\textbf{Role of the rank split.}
The principal factors remain freely trainable, so a principal-only adapter can also learn corrections outside its initial singular subspaces.
PROLA assigns \(r_p\) factors to the principal branch for pretrained-aligned adaptation and reserves \(r_o\) factors for a separate projected branch dedicated to orthogonal corrections.
The rank split fixes the capacity assigned to these two paths under the same total rank budget, while the learned factors determine their contributions to the effective update.
To examine how the resulting update is distributed across the pretrained subspaces, let
\(\Delta W^{\mathrm{eff}}:=W^{\mathrm{eff}}-W_0\),
\(P=U_pU_p^\top\), and \(Q=V_pV_p^\top\), with
\(P_\perp=I-P\) and \(Q_\perp=I-Q\).
Then
\begin{equation}
\Delta W^{\mathrm{eff}}
=
\underbrace{P\Delta W^{\mathrm{eff}}Q}_{\text{principal block}}
+
\underbrace{
P\Delta W^{\mathrm{eff}}Q_\perp
+
P_\perp\Delta W^{\mathrm{eff}}Q
}_{\text{cross blocks}}
+
\underbrace{
P_\perp\Delta W^{\mathrm{eff}}Q_\perp
}_{\text{orthogonal block}}.
\label{eq:blockwise_update_decomposition}
\end{equation}
Section~\ref{sec:update_diagnostics} further compares the update energy across these blocks.

\subsection{Gradient-Guided Scaling of Optimizer Updates}
\label{sec:step-allocation}

In Section~\ref{sec:peft}, we allocate the fixed rank budget across the principal and orthogonal branches.
We next rescale how strongly each rank-1 component is updated during optimization.
For \(g\in\{p,o\}\), where \(g\) indexes the principal branch \(p\) or the orthogonal branch \(o\), let \(r_g\) denote the corresponding branch rank.
Let \(\mathbf b_k^g\) be the \(k\)-th column of \(B_g\), and let \((\mathbf a_k^g)^\top\) be the \(k\)-th row of \(A_g\).
Then \((\mathbf b_k^g,\mathbf a_k^g)\) forms the \(k\)-th trainable rank-1 component of branch \(g\), with \(B_gA_g=\sum_{k=1}^{r_g}\mathbf b_k^g(\mathbf a_k^g)^\top
\).
For the orthogonal branch, this component is defined before applying the fixed projectors \(P_\perp\) and \(Q_\perp\).

\textbf{Scaling rank-1 updates.}
Given this component-wise view, we attach a rank-1 update scale \(\alpha_k^g\) to each factor pair.
Let \(\tilde{\Delta}\mathbf b_k^g\) and \(\tilde{\Delta}\mathbf a_k^g\) denote the raw updates proposed by the base optimizer.
The rescaled updates are
\begin{equation}
\Delta\mathbf b^g_k
=
\alpha^g_k
\tilde{\Delta}\mathbf b^g_k,
\qquad
\Delta\mathbf a^g_k
=
\alpha^g_k
\tilde{\Delta}\mathbf a^g_k .
\end{equation}
This post-optimizer scaling leaves the base optimizer's raw-update computation unchanged and introduces no additional trainable parameters.

\textbf{Gradient-guided scale computation.}
We compute \(\alpha_k^g\) from gradient-based signals at the component and branch levels.
The component signal \(u_k^g\) measures the gradient magnitude of the \(k\)-th rank-1 component, and the branch signal \(\mu^g\) averages these signals within branch \(g\):
\begin{equation}
u^g_k
=
\left\|
\nabla_{\mathbf b^g_k}\mathcal L_{ST}
\right\|_2
+
\left\|
\nabla_{\mathbf a^g_k}\mathcal L_{ST}
\right\|_2,
\qquad
\mu^g
=
\frac{1}{r_g}
\sum_{k=1}^{r_g}
u^g_k.
\end{equation}
These signals are normalized into branch and component scores and converted into the final rank-1 update scale:
\begin{equation}
\pi^g_k
=
\frac{u^g_k+\epsilon}
{\sum_{j=1}^{r_g}(u^g_j+\epsilon)},
\qquad
\rho^g
=
\frac{\mu^g+\epsilon}
{\sum_{g'\in\{p,o\}}(\mu^{g'}+\epsilon)},
\qquad
\alpha^g_k
=
r\times\operatorname{sg}\!\left(\rho^g\pi^g_k\right),
\end{equation}
where \(\pi_k^g\) is the component score, \(\rho^g\) is the branch score, \(r=r_p+r_o\), \(\epsilon>0\) is a numerical constant, and \(\operatorname{sg}(\cdot)\) denotes stop-gradient.

Since \(\rho^g\) and \(\pi_k^g\) are normalized across branches and within each branch, respectively, \(\alpha_k^g\) has unit average over all \(r\) low-rank components.
Thus, the scaling reallocates update magnitude toward high-signal components while preserving the average layer-level update scale.
Additional details and diagnostics are provided in Appendix~\ref{app:allocation_diagnostics}.

\section{Experiments}
\begin{table*}[t]
\centering
\small
\setlength{\tabcolsep}{3.5pt}
\renewcommand{\arraystretch}{1.50}
\setlength{\arrayrulewidth}{0.3pt}
\caption{\textbf{Quantitative results on WeatherBench~2~\cite{rasp2024weatherbench}.} Results are reported at 3h and 6h lead times.
We report latitude-weighted RMSE ($\downarrow$) and ACC ($\uparrow$). Each cell shows the mean above and $\pm$ standard deviation below, over three seeds.
Best displayed mean results among PEFT methods are boldfaced.}
\label{tab:quant_main_3h6h}
\resizebox{0.85\textwidth}{!}{%
\begin{tabularx}{\textwidth}{Xc|rr|rr|rr|rr|rr}
\toprule
\multirow{2}{*}{Method} & \multirow{2}{*}{\shortstack{Trainable\\Params}} & \multicolumn{2}{c|}{Z500} & \multicolumn{2}{c|}{T850} & \multicolumn{2}{c|}{T2M} & \multicolumn{2}{c|}{U10} & \multicolumn{2}{c}{V10} \\
& & RMSE & ACC & RMSE & ACC & RMSE & ACC & RMSE & ACC & RMSE & ACC \\
\midrule
\rowcolor{prolaSection}
\multicolumn{12}{l}{\textit{3h lead}} \\
\rowcolor{gray!6}
\ftcell{Full FT} & \ftcell{1.26B} & \ftcell{\prolaStackStat{60.13}{1.61}} & \ftcell{\prolaStackStat{0.998}{0.000}} & \ftcell{\prolaStackStat{0.813}{0.002}} & \ftcell{\prolaStackStat{0.987}{0.000}} & \ftcell{\prolaStackStat{1.064}{0.009}} & \ftcell{\prolaStackStat{0.979}{0.000}} & \ftcell{\prolaStackStat{0.902}{0.004}} & \ftcell{\prolaStackStat{0.975}{0.000}} & \ftcell{\prolaStackStat{0.925}{0.009}} & \ftcell{\prolaStackStat{0.974}{0.001}} \\
Head only & 1.07M & \prolaStackStat{94.93}{0.49} & \prolaStackStat{0.996}{0.000} & \prolaStackStat{1.049}{0.006} & \prolaStackStat{0.978}{0.000} & \prolaStackStat{2.083}{0.009} & \prolaStackStat{0.915}{0.001} & \prolaStackStat{1.648}{0.018} & \prolaStackStat{0.914}{0.001} & \prolaStackStat{1.563}{0.018} & \prolaStackStat{0.924}{0.002} \\
LoRA & 3.92M & \prolaStackStat{54.72}{0.36} & \prolaStackBest{0.999}{0.000} & \prolaStackStat{0.898}{0.002} & \prolaStackStat{0.984}{0.000} & \prolaStackStat{1.199}{0.007} & \prolaStackStat{0.973}{0.000} & \prolaStackStat{1.219}{0.008} & \prolaStackStat{0.954}{0.000} & \prolaStackStat{1.181}{0.002} & \prolaStackStat{0.957}{0.000} \\
DoRA & 4.16M & \prolaStackStat{54.32}{0.35} & \prolaStackBest{0.999}{0.000} & \prolaStackStat{0.893}{0.002} & \prolaStackStat{0.984}{0.000} & \prolaStackStat{1.191}{0.006} & \prolaStackStat{0.973}{0.000} & \prolaStackStat{1.209}{0.008} & \prolaStackStat{0.955}{0.000} & \prolaStackStat{1.173}{0.001} & \prolaStackStat{0.958}{0.000} \\
AdaLoRA & 5.35M & \prolaStackStat{53.40}{0.27} & \prolaStackBest{0.999}{0.000} & \prolaStackStat{0.884}{0.006} & \prolaStackStat{0.984}{0.000} & \prolaStackStat{1.176}{0.005} & \prolaStackStat{0.974}{0.000} & \prolaStackStat{1.175}{0.005} & \prolaStackStat{0.957}{0.000} & \prolaStackStat{1.158}{0.002} & \prolaStackStat{0.959}{0.000} \\
PiSSA & 3.92M & \prolaStackStat{54.55}{0.89} & \prolaStackBest{0.999}{0.000} & \prolaStackStat{0.893}{0.002} & \prolaStackStat{0.984}{0.000} & \prolaStackStat{1.198}{0.005} & \prolaStackStat{0.973}{0.000} & \prolaStackStat{1.215}{0.005} & \prolaStackStat{0.954}{0.000} & \prolaStackStat{1.175}{0.003} & \prolaStackStat{0.958}{0.000} \\
OPLoRA & 3.92M & \prolaStackStat{54.94}{0.39} & \prolaStackBest{0.999}{0.000} & \prolaStackStat{0.899}{0.003} & \prolaStackStat{0.984}{0.000} & \prolaStackStat{1.204}{0.007} & \prolaStackStat{0.973}{0.000} & \prolaStackStat{1.223}{0.007} & \prolaStackStat{0.954}{0.000} & \prolaStackStat{1.185}{0.003} & \prolaStackStat{0.957}{0.000} \\
\rowcolor{prolaSky}
PROLA & 3.92M & \prolaStackBest{50.58}{0.42} & \prolaStackBest{0.999}{0.000} & \prolaStackBest{0.864}{0.003} & \prolaStackBest{0.985}{0.000} & \prolaStackBest{1.129}{0.002} & \prolaStackBest{0.976}{0.000} & \prolaStackBest{1.143}{0.003} & \prolaStackBest{0.960}{0.000} & \prolaStackBest{1.120}{0.001} & \prolaStackBest{0.962}{0.000} \\
\midrule
\rowcolor{prolaSection}
\multicolumn{12}{l}{\textit{6h lead}} \\
\rowcolor{gray!6}
\ftcell{Full FT} & \ftcell{1.26B} & \ftcell{\prolaStackStat{106.68}{2.63}} & \ftcell{\prolaStackStat{0.995}{0.000}} & \ftcell{\prolaStackStat{1.114}{0.008}} & \ftcell{\prolaStackStat{0.976}{0.000}} & \ftcell{\prolaStackStat{1.323}{0.022}} & \ftcell{\prolaStackStat{0.970}{0.001}} & \ftcell{\prolaStackStat{1.135}{0.012}} & \ftcell{\prolaStackStat{0.961}{0.001}} & \ftcell{\prolaStackStat{1.209}{0.010}} & \ftcell{\prolaStackStat{0.956}{0.001}} \\
Head only & 1.07M & \prolaStackStat{144.03}{1.47} & \prolaStackStat{0.990}{0.000} & \prolaStackStat{1.315}{0.008} & \prolaStackStat{0.965}{0.001} & \prolaStackStat{2.656}{0.021} & \prolaStackStat{0.856}{0.001} & \prolaStackStat{1.947}{0.024} & \prolaStackStat{0.877}{0.002} & \prolaStackStat{1.921}{0.018} & \prolaStackStat{0.883}{0.003} \\
LoRA & 3.92M & \prolaStackStat{79.17}{0.33} & \prolaStackBest{0.997}{0.000} & \prolaStackStat{0.994}{0.003} & \prolaStackStat{0.980}{0.000} & \prolaStackStat{1.280}{0.012} & \prolaStackStat{0.969}{0.001} & \prolaStackStat{1.314}{0.005} & \prolaStackStat{0.946}{0.000} & \prolaStackStat{1.306}{0.002} & \prolaStackStat{0.948}{0.000} \\
DoRA & 4.16M & \prolaStackStat{78.40}{0.30} & \prolaStackBest{0.997}{0.000} & \prolaStackStat{0.989}{0.003} & \prolaStackStat{0.980}{0.000} & \prolaStackStat{1.272}{0.013} & \prolaStackStat{0.970}{0.001} & \prolaStackStat{1.306}{0.006} & \prolaStackStat{0.947}{0.000} & \prolaStackStat{1.298}{0.001} & \prolaStackStat{0.948}{0.000} \\
AdaLoRA & 5.35M & \prolaStackStat{77.51}{0.94} & \prolaStackBest{0.997}{0.000} & \prolaStackStat{0.983}{0.006} & \prolaStackStat{0.981}{0.000} & \prolaStackStat{1.251}{0.002} & \prolaStackStat{0.971}{0.001} & \prolaStackStat{1.282}{0.004} & \prolaStackStat{0.949}{0.000} & \prolaStackStat{1.279}{0.002} & \prolaStackStat{0.950}{0.000} \\
PiSSA & 3.92M & \prolaStackStat{78.43}{0.93} & \prolaStackBest{0.997}{0.000} & \prolaStackStat{0.989}{0.004} & \prolaStackStat{0.980}{0.000} & \prolaStackStat{1.251}{0.009} & \prolaStackStat{0.971}{0.001} & \prolaStackStat{1.298}{0.004} & \prolaStackStat{0.948}{0.000} & \prolaStackStat{1.289}{0.003} & \prolaStackStat{0.949}{0.000} \\
OPLoRA & 3.92M & \prolaStackStat{79.60}{0.38} & \prolaStackBest{0.997}{0.000} & \prolaStackStat{0.996}{0.003} & \prolaStackStat{0.980}{0.000} & \prolaStackStat{1.281}{0.012} & \prolaStackStat{0.969}{0.000} & \prolaStackStat{1.319}{0.005} & \prolaStackStat{0.946}{0.000} & \prolaStackStat{1.311}{0.002} & \prolaStackStat{0.947}{0.000} \\
\rowcolor{prolaSky}
PROLA & 3.92M & \prolaStackBest{72.35}{0.64} & \prolaStackBest{0.997}{0.000} & \prolaStackBest{0.956}{0.002} & \prolaStackBest{0.982}{0.000} & \prolaStackBest{1.212}{0.006} & \prolaStackBest{0.973}{0.001} & \prolaStackBest{1.240}{0.005} & \prolaStackBest{0.952}{0.000} & \prolaStackBest{1.234}{0.002} & \prolaStackBest{0.953}{0.000} \\
\bottomrule
\end{tabularx}%
}

\end{table*}

\subsection{Experimental Settings}

For global PSWD, we adapt Aurora~\cite{bodnar2025foundation} on WeatherBench~2~\cite{rasp2024weatherbench}.
Given \(\mathcal{X}^{\mathrm{LR}}_t=(\mathbf{x}^{\mathrm{LR}}_{t-\delta_c},\mathbf{x}^{\mathrm{LR}}_t)\) on the \(5.625^\circ\) grid with \(\delta_c=6\mathrm{h}\), we predict \(\mathcal{Y}^{\mathrm{HR}}_t\) on the \(1.5^\circ\) grid at leads \(\mathcal H_f=\{1\mathrm{h},\ldots,6\mathrm{h}\}\), \textit{i.e.,} \(\delta_f=1\mathrm{h}\).
The 68 target channels comprise three surface variables and five upper-air variables at 13 pressure levels.
We report results for five representative variables: 500 hPa geopotential (Z500), 850 hPa temperature (T850), 2m temperature (T2M), and 10m wind components (U10 and V10).
Table~\ref{tab:quant_main_3h6h} reports latitude-weighted RMSE and ACC at 3h (an intermediate lead) and 6h (the endpoint) as mean \(\pm\) standard deviation over three seeds.
All methods train both correction heads within the same scaffold-and-correct pipeline; Full FT also updates the backbone, while Head only keeps it frozen.
The low-rank methods adapt the same attention projections.
PROLA, LoRA~\cite{hu2022lora}, DoRA~\cite{liu2024dora}, PiSSA~\cite{meng2024pissa}, and OPLoRA~\cite{xiong2026oplora} use total rank \(r=8\), and AdaLoRA~\cite{zhang2023adalora} targets rank 8.

We extend the evaluation to ERA5-to-CERRA downscaling~\cite{hersbach2020era5,ridal2024cerra}, using coarse ERA5 inputs at \(t-6\mathrm{h}\) and \(t\) to predict high-resolution CERRA fields over Central Europe at leads \(\mathcal H_f=\{3\mathrm{h},6\mathrm{h}\}\).
Table~\ref{tab:regional_generalization} reports latitude-weighted RMSE for the same five variables at 3h.
Regional precipitation forecasting follows the protocol in~\cite{cao2026taskadaptive}, using ERA5 inputs at \(t-12\mathrm{h}\) and \(t\) to predict 6h precipitation totals ending at each forecast valid time.
Table~\ref{tab:precip_generalization} reports SEEPS, RMSE, and ACC at 12h.
Both tasks compare PROLA with LoRA, PiSSA, and OPLoRA.
See Appendices~\ref{app:experimental_setup}--\ref{app:peft_config} for experimental details and metric definitions.

\subsection{Main Results}
\begin{wrapfigure}{r}{0.38\textwidth}
\centering
\includegraphics[width=\linewidth]{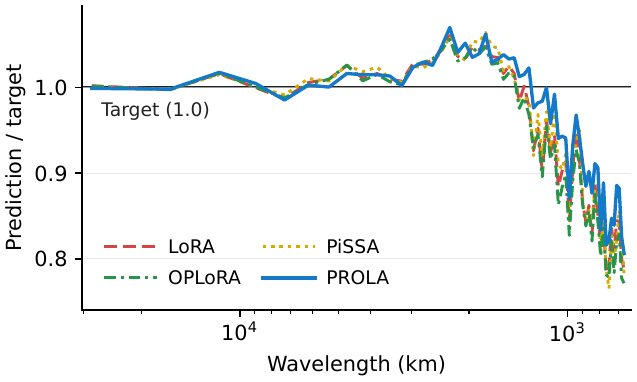}
\caption{\textbf{Spatial spectral fidelity.}
Prediction-to-target ratios of mean spherical harmonic power for T2M
at 6h over test cases. A ratio of one matches target power.}
\label{fig:spectral_t2m}
\vspace{-0.5em}
\end{wrapfigure}

Table~\ref{tab:quant_main_3h6h} shows that PROLA achieves the lowest RMSE and highest ACC among PEFT methods across all reported variables and leads.
Relative to AdaLoRA (36\% more trainable parameters), PROLA reduces Z500 RMSE by 5.3\% at 3h (50.58 vs.\ 53.40) and 6.7\% at 6h (72.35 vs.\ 77.51), with reductions of 2.3--4.0\% and 2.7--3.5\% on the remaining variables.
Notably, despite adapting only 3.92M parameters (under 0.4\% of Full FT), PROLA outperforms Full FT on Z500 at both leads and on T850 and T2M at 6h.
These results suggest that structured low-rank adaptation provides useful implicit regularization in PSWD by limiting backbone update flexibility.

Figure~\ref{fig:spectral_t2m} compares prediction-to-target ratios of mean spherical harmonic power for T2M at 6h.
The ratios fall below one at short wavelengths, indicating underestimated fine-scale temperature variability.
PROLA remains closer to one at shorter wavelengths, supporting improved spatial spectral fidelity.
Full per-variable and per-lead results, further analysis, and qualitative forecasts are provided in Appendices~\ref{app:full_results} and~\ref{app:qualitative}.

\subsection{Generalization to Regional Weather Tasks}
\label{sec:generalization}

\begin{table*}[t]
\centering
\begin{minipage}{0.92\linewidth}
\centering

\footnotesize
\setlength{\tabcolsep}{1.5pt}
\captionsetup{
    position=top,
    font=normalsize,
    skip=3pt,
    margin=0pt,
    justification=raggedright,
    singlelinecheck=false
}

\noindent
\begin{minipage}[t]{0.52\linewidth}
\centering
\caption{\textbf{ERA5-to-CERRA.}\newline
Latitude-weighted RMSE ($\downarrow$) at 3h.}
\label{tab:regional_generalization}
\begin{tabularx}{\linewidth}{l*{5}{>{\centering\arraybackslash}X}}
\toprule
Method & Z500 & T850 & T2M & U10 & V10 \\
\midrule
LoRA & 72.14 & 1.076 & 2.150 & 1.370 & 1.327 \\
PiSSA & 71.80 & 1.072 & 2.126 & 1.354 & 1.311 \\
OPLoRA & 72.56 & 1.081 & 2.132 & 1.371 & 1.332 \\
\rowcolor{prolaSky}
PROLA & \textbf{68.44} & \textbf{1.028} & \textbf{2.021} & \textbf{1.316} & \textbf{1.274} \\
\bottomrule
\end{tabularx}
\end{minipage}\hfill
\begin{minipage}[t]{0.42\linewidth}
\centering
\caption{\textbf{Precipitation.}\newline
6h totals at 12h lead (RMSE in cm).}
\label{tab:precip_generalization}
\begin{tabularx}{\linewidth}{l*{3}{>{\centering\arraybackslash}X}}
\toprule
Method & SEEPS $\downarrow$ & RMSE $\downarrow$ & ACC $\uparrow$ \\
\midrule
LoRA & 0.608 & 0.212 & 0.660 \\
PiSSA & 0.612 & 0.213 & 0.656 \\
OPLoRA & 0.605 & 0.212 & 0.661 \\
\rowcolor{prolaSky}
PROLA & \textbf{0.590} & \textbf{0.208} & \textbf{0.676} \\
\bottomrule
\end{tabularx}
\end{minipage}
\end{minipage}
\end{table*}

\begin{figure*}[t]
\centering
\includegraphics[width=0.94\textwidth]{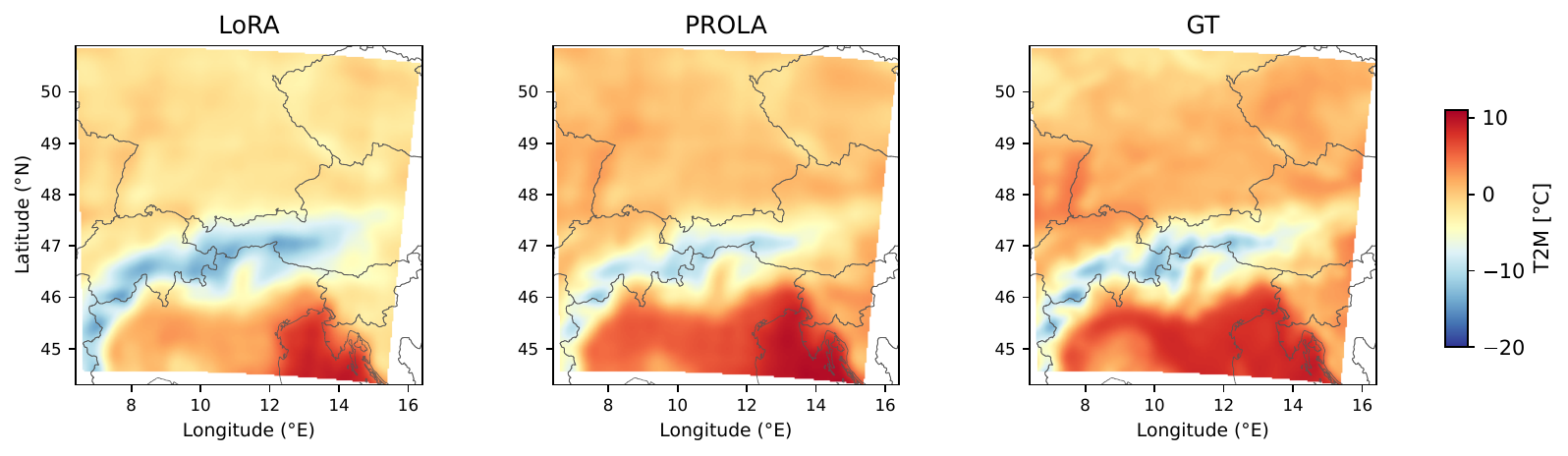}
\caption{\textbf{Qualitative ERA5-to-CERRA prediction at 6h.}
T2M predictions from LoRA and PROLA, and the CERRA target,
valid at 12:00 UTC on 4 February 2018.
}
\label{fig:cerra_qualitative}
\end{figure*}

\textbf{ERA5-to-CERRA downscaling.}
Table~\ref{tab:regional_generalization} shows that PROLA achieves the lowest latitude-weighted RMSE on all five reported variables at 3h.
Relative to PiSSA, the strongest baseline in this comparison, PROLA reduces Z500 RMSE by 4.7\% (68.44 vs.\ 71.80) and T2M RMSE by 4.9\% (2.021 vs.\ 2.126).
Figure~\ref{fig:cerra_qualitative} compares the 6h T2M forecasts with the CERRA target.
In this example, PROLA reproduces the contrast between the cooler Alpine region and the warmer areas to its south more closely than LoRA.
PROLA also better matches the warmer temperatures north of the Alps, where LoRA predicts a cooler field.
See Appendix~\ref{app:regional_generalization} for the full per-lead results.

\textbf{Regional precipitation forecasting.}
Table~\ref{tab:precip_generalization} shows that PROLA achieves the lowest SEEPS and RMSE and the highest ACC among the compared PEFT methods at the 12h lead.
Compared with OPLoRA, PROLA reduces SEEPS from 0.605 to 0.590 and RMSE from 0.212 to 0.208\,cm, improving the prediction of both precipitation categories and accumulated amounts.
The higher ACC (0.676 vs.\ 0.661) also indicates closer agreement with the spatial pattern of precipitation anomalies.
Together with the ERA5-to-CERRA results, these gains show that PROLA outperforms the low-rank baselines across both regional downscaling and precipitation forecasting.
See Appendix~\ref{app:regional_generalization} for results at additional lead times.

\section{Analysis}

\begin{table*}[t]
\centering
\caption{\textbf{Direct prediction and pipeline ablation.} Latitude-weighted RMSE ($\downarrow$) at 3h and 6h. Best results in the pipeline ablation are boldfaced.}
\label{tab:ablation_head}
\resizebox{0.86\textwidth}{!}{%
\small
\setlength{\tabcolsep}{3pt}
\renewcommand{\arraystretch}{1.1}
\setlength{\arrayrulewidth}{0.3pt}
\begin{tabularx}{\textwidth}{X|*{5}{r}|*{5}{r}}
\toprule
 & \multicolumn{5}{c}{\textbf{3h lead}} & \multicolumn{5}{c}{\textbf{6h lead}} \\
\cmidrule(lr){2-6}\cmidrule(lr){7-11}
\textbf{Method / Variant} & Z500 & T850 & T2M & U10 & V10 & Z500 & T850 & T2M & U10 & V10 \\
\midrule
\rowcolor{prolaSection}
\multicolumn{11}{l}{\textit{Direct prediction}} \\
LoRA & 58.91 & 0.979 & 1.256 & 1.481 & 1.484 & 129.60 & 1.214 & 1.394 & 1.755 & 1.781 \\
PiSSA & 56.47 & 0.942 & 1.191 & 1.272 & 1.239 & 128.00 & 1.152 & 1.292 & 1.392 & 1.405 \\
OPLoRA & 58.15 & 0.983 & 1.254 & 1.463 & 1.457 & 131.16 & 1.219 & 1.388 & 1.741 & 1.735 \\
\rowcolor{prolaSky}
PROLA & 53.13 & 0.898 & 1.165 & 1.211 & 1.177 & 125.07 & 1.095 & 1.289 & 1.358 & 1.344 \\
\midrule
\rowcolor{prolaSection}
\multicolumn{11}{l}{\textit{Scaffold-and-correct ablation (PROLA)}} \\
w/o $C_s,C_t$ & 87.73 & 0.959 & 1.307 & 1.277 & 1.270 & 111.87 & 1.037 & 1.289 & 1.339 & 1.371 \\
w/o $C_t$ & 84.53 & 0.938 & 1.290 & 1.259 & 1.254 & 108.80 & 1.016 & 1.281 & 1.332 & 1.365 \\
w/o $C_s$ & 53.35 & 0.880 & 1.138 & 1.155 & 1.133 & 78.25 & 0.974 & \textbf{1.209} & \textbf{1.233} & 1.237 \\
\rowcolor{prolaSky}
Full pipeline & \textbf{50.30} & \textbf{0.866} & \textbf{1.128} & \textbf{1.140} & \textbf{1.119} & \textbf{73.07} & \textbf{0.958} & 1.211 & 1.236 & \textbf{1.234} \\
\bottomrule
\end{tabularx}%
}
\end{table*}

\begin{table*}[t]
\centering
\caption{\textbf{Ablation studies on PROLA design.} Latitude-weighted RMSE ($\downarrow$) at 3h and 6h. The design ablations use $r_p+r_o=8$. \abcheck marks indicate update scaling; $\rho$ and $\pi$ denote branch score and component score. Best results are boldfaced.}
\label{tab:ablation_prola}
\resizebox{0.86\textwidth}{!}{%
\small
\setlength{\tabcolsep}{3pt}
\renewcommand{\arraystretch}{1.1}
\setlength{\arrayrulewidth}{0.3pt}
\begin{tabularx}{\textwidth}{Xccc|*{5}{r}|*{5}{r}}
\toprule
 &  &  &  & \multicolumn{5}{c}{\textbf{3h lead}} & \multicolumn{5}{c}{\textbf{6h lead}} \\
\cmidrule(lr){5-9}\cmidrule(lr){10-14}
\textbf{Configuration} & $r_p$ & $r_o$ & Scaling & Z500 & T850 & T2M & U10 & V10 & Z500 & T850 & T2M & U10 & V10 \\
\midrule
\rowcolor{prolaSection}
\multicolumn{14}{l}{\textbf{(i)}\enspace\textit{Branch composition and update scaling}} \\
Principal & 8 & 0 & -- & 53.66 & 0.894 & 1.193 & 1.209 & 1.178 & 78.55 & 0.987 & 1.248 & 1.294 & 1.290 \\
Principal & 8 & 0 & \abcheck & 53.06 & 0.887 & 1.184 & 1.198 & 1.166 & 77.71 & 0.979 & 1.241 & 1.289 & 1.285 \\
Orthogonal & 0 & 8 & -- & 54.54 & 0.899 & 1.197 & 1.216 & 1.188 & 79.83 & 0.994 & 1.272 & 1.314 & 1.314 \\
Orthogonal & 0 & 8 & \abcheck & 52.81 & 0.886 & 1.175 & 1.189 & 1.157 & 76.67 & 0.977 & 1.260 & 1.284 & 1.278 \\
Both branches & 4 & 4 & -- & 51.36 & 0.875 & 1.143 & 1.165 & 1.133 & 74.38 & 0.967 & 1.221 & 1.256 & 1.250 \\
\midrule
\rowcolor{prolaSection}
\multicolumn{14}{l}{\textbf{(ii)}\enspace\textit{Rank allocation}} \\
Principal-heavy & 6 & 2 & \abcheck & 51.46 & 0.876 & 1.153 & 1.161 & 1.136 & 74.19 & 0.970 & 1.225 & 1.249 & 1.244 \\
Orthogonal-heavy & 2 & 6 & \abcheck & 52.52 & 0.890 & 1.176 & 1.200 & 1.160 & 76.35 & 0.983 & 1.263 & 1.285 & 1.282 \\
\midrule
\rowcolor{prolaSection}
\multicolumn{14}{l}{\textbf{(iii)}\enspace\textit{Scaling components}} \\
$\rho$ only & 4 & 4 & \abcheck & 52.16 & 0.885 & 1.168 & 1.188 & 1.151 & 76.06 & 0.978 & 1.240 & 1.273 & 1.264 \\
$\pi$ only & 4 & 4 & \abcheck & 51.94 & 0.882 & 1.155 & 1.176 & 1.144 & 75.45 & 0.976 & 1.235 & 1.267 & 1.261 \\
\midrule
\rowcolor{prolaSky}
PROLA & 4 & 4 & \abcheck & \textbf{50.30} & \textbf{0.866} & \textbf{1.128} & \textbf{1.140} & \textbf{1.119} & \textbf{73.07} & \textbf{0.958} & \textbf{1.211} & \textbf{1.236} & \textbf{1.234} \\
\bottomrule
\end{tabularx}
}
\end{table*}

\subsection{Ablation of Scaffold-and-Correct Pipeline}
\label{sec:pipeline_ablation}

Table~\ref{tab:ablation_head} compares direct prediction across PEFT methods and ablates the scaffold-and-correct pipeline with PROLA.
Direct prediction jointly outputs all target lead times without the spatial-to-temporal decomposition.
All pipeline ablations use PROLA and remove the spatial correction head \(C_s\), the temporal correction head \(C_t\), or both, leaving the corresponding scaffolds uncorrected.
The full pipeline achieves the lowest RMSE across all five variables at 3h, reducing Z500 RMSE from 53.13 with direct prediction to 50.30.
Removing both correction heads raises 3h Z500 RMSE to 87.73, indicating that the scaffolds alone are insufficient for PSWD.
Removing \(C_t\) alone yields a much higher 3h Z500 RMSE than removing \(C_s\) alone (84.53 vs.\ 53.35), highlighting the role of temporal correction at intermediate leads.
Although \(w/o~C_s\) remains competitive on T2M and U10 at 6h, its higher Z500 and T850 errors support spatial endpoint correction as a temporal anchor.

\subsection{PROLA Branch and Scaling Ablation}

Table~\ref{tab:ablation_prola} compares branch configurations and update scaling at a fixed total rank of \(r_p+r_o=8\).
The results in~\textbf{(i)} show that even without update scaling, 3h Z500 RMSE is lower with both branches (51.36) than with the principal branch alone (53.66) or the orthogonal branch alone (54.54).
Update scaling improves both single-branch variants and further lowers 3h Z500 RMSE to 50.30 when both branches are used.
With scaling enabled, the balanced \(4+4\) split outperforms both uneven allocations in~\textbf{(ii)} on all five variables at both leads.
At 6h, Z500 RMSE is 73.07 for \(4+4\), compared with 74.19 for \(6+2\) and 76.35 for \(2+6\).
The comparisons in~\textbf{(iii)} keep the \(4+4\) split fixed. Using the branch score \(\rho\) alone raises 6h Z500 RMSE to 76.06, while using the component score \(\pi\) alone raises it to 75.45.
Full PROLA (bottom row) combines both scores and achieves the lowest RMSE across all five variables at 3h and 6h.
See Appendix~\ref{app:additional_ablations} for the full per-lead comparisons.

\subsection{Update Geometry under a Fixed Rank Budget}
\label{sec:update_diagnostics}

We apply Eq.~\eqref{eq:blockwise_update_decomposition} to compare how PROLA and PiSSA use the same rank budget.
Let \(\mathcal A\) index the adapted attention matrices.
For each \(\ell\in\mathcal A\), we consider the effective update
\(\Delta W_\ell^{\mathrm{eff}}=W_\ell^{\mathrm{eff}}-W_{\ell,0}\),
which includes the learned changes from both branches in PROLA.
We decompose this update using the top \(r\) left and right singular vectors of the pretrained weight \(W_{\ell,0}\).
All methods are evaluated using these same reference subspaces.
Let \(\Delta W_\ell^b\) denote the component within both reference subspaces
(\(b=\mathrm{prin}\)), outside both (\(\mathrm{orth}\)), or within one and outside the other
(\(\mathrm{cross}\), summing both cross blocks).
The energy share of block \(b\) is obtained by summing its squared Frobenius norm across adapted matrices and normalizing by the total update energy:
\(\bar e_b=
\sum_{\ell\in\mathcal A}\|\Delta W_\ell^b\|_F^2/\sum_{\ell\in\mathcal A}\|\Delta W_\ell^{\mathrm{eff}}\|_F^2\).

\begin{wrapfigure}{r}{0.42\textwidth}
\centering
\includegraphics[width=\linewidth]{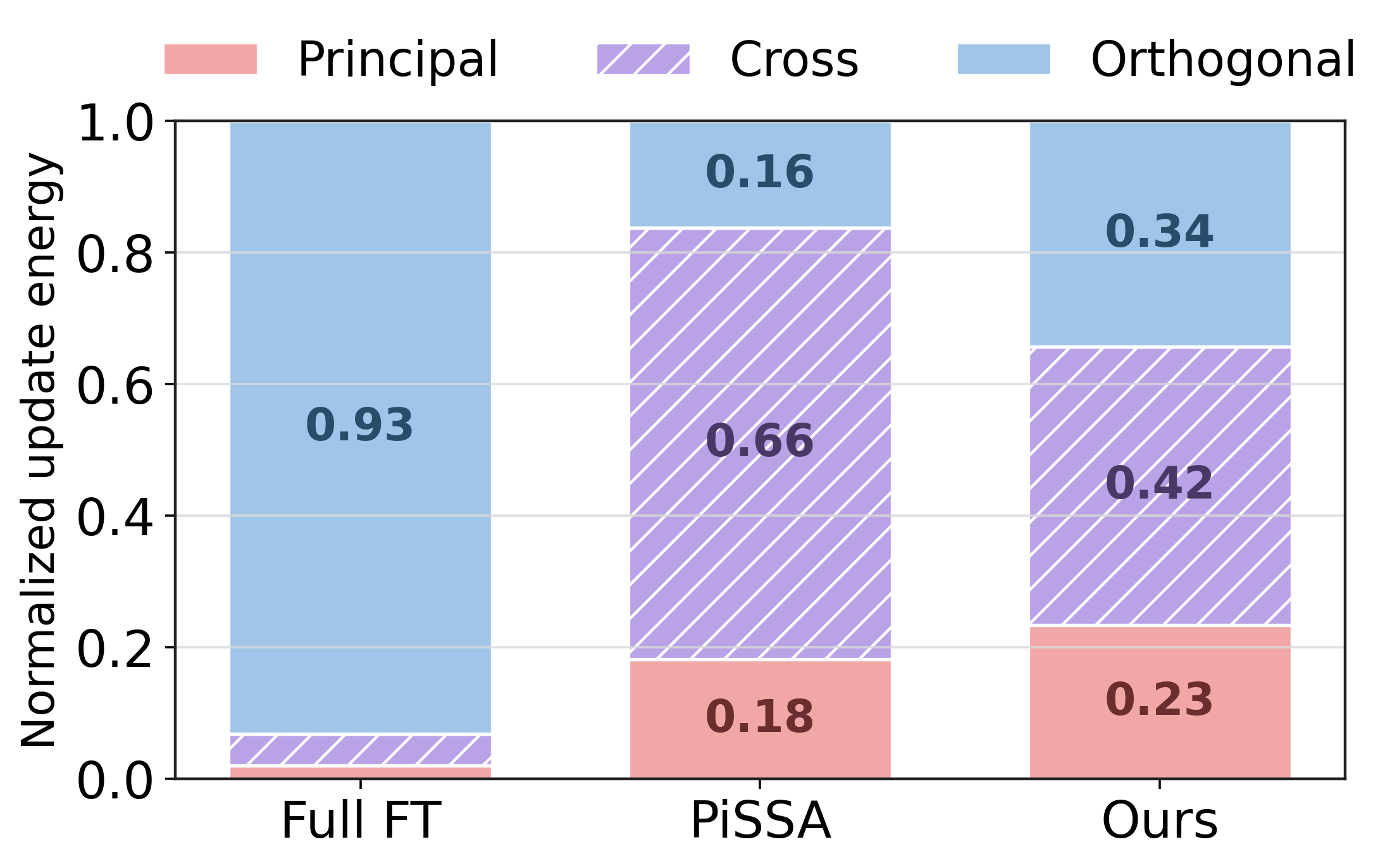}
\caption{\textbf{Blockwise update energy.}
Fractions of total update energy in the principal, cross, and orthogonal blocks.}
\label{fig:update_diagnostics}
\end{wrapfigure}

Figure~\ref{fig:update_diagnostics} shows that PROLA more than doubles the orthogonal share of update energy relative to PiSSA (34\% vs.\ 16\%).
PROLA retains 23\% of its update energy in the principal block and 42\% in the cross blocks, whereas PiSSA places 18\% in the principal block and concentrates 66\% in the cross blocks.
The principal factors in PiSSA remain freely trainable after initialization, allowing it to learn corrections outside its initial principal subspaces.
Under the same rank budget, PROLA achieves a larger orthogonal share, accompanied by a lower cross-block share and a higher principal share.
Thus, its increased share of orthogonal corrections is maintained alongside updates within the pretrained principal subspaces.
For reference, the orthogonal share is 93\% for unconstrained Full FT under the same partition.
See Appendix~\ref{app:cross_subspace} for the projector definitions and energy aggregation used in this comparison.

\section{Related Work}

\textbf{Data-driven weather modeling.}
Data-driven weather forecasting has progressed from early global models~\citep{pathak2022fourcastnet, keisler2022forecasting} to systems achieving competitive or state-of-the-art performance on global benchmarks~\citep{GraphCast:lam2023learning, bi2023accurate, lang2024aifs, rasp2024weatherbench, chen2023fuxi, chen2025fengwu, Stormer:nguyen2023scaling, bonev2023spherical}, alongside hybrid physics-ML and probabilistic systems~\citep{NeuralGCM:kochkov2024neural, GenCast:price2025probabilistic}.
Weather downscaling reconstructs high-resolution fields from low-resolution inputs through spatial statistical or generative refinement~\citep{vandal2017deepsd, mardani2025residual, gonzalez2023dl4ds, wang2021deep, rampal2022high, harris2022generative, price2022increasing}, with fewer studies addressing time-evolving or joint spatiotemporal refinement~\citep{glawion2025global, leinonen2020stochastic, glawion2023spategan}.
Pretrained weather foundation models (WFMs) support diverse atmospheric tasks~\citep{bodnar2025foundation, nguyen2023climax, schmude2024prithvi, chen2023foundation}, with their growing scale motivating PEFT~\citep{cao2026taskadaptive}.
Existing work often treats forecasting, spatial downscaling, temporal refinement, and WFM adaptation separately, whereas PSWD couples forecasting with spatiotemporal refinement in one objective and provides a comprehensive PEFT benchmark for WFMs.

\textbf{Low-rank parameter-efficient fine-tuning.}
PEFT avoids costly full fine-tuning by adapting large pretrained backbones with few trainable parameters~\citep{han2024parameter}.
LoRA trains low-rank factors with a frozen backbone~\citep{hu2022lora}; variants add quantization, dynamic rank selection, parameter sharing, adaptive rank allocation, factor-wise learning rates, and magnitude-direction decomposition~\citep{liu2024dora, zhang2023adalora, dettmers2023qlora, valipour2023dylora, kopiczko2023vera, hayou2024lora+}.
Other methods use SVD/decomposition-based initialization~\citep{meng2024pissa, li2024loftq, yang2024corda} to exploit pretrained weight structure, or orthogonality constraints to reduce interference with pretrained or previously learned subspaces~\citep{xiong2026oplora, wang2023orthogonal, wang2026complementary}.
Related approaches distribute principal components across devices~\citep{wang2025hdpissa}, restrict orthogonal transformations to selected subspaces~\citep{wu2026efficient, zhao2026loft}, or combine coefficient updates within a fixed principal basis with orthogonal-complement adaptation~\citep{chen2026hiplora}.
PROLA splits a fixed rank budget between principal and orthogonal branches, trains its principal factors, and scales optimizer updates using gradient-based branch and rank-1 component scores.

\section{Conclusion}

We formulated Predictive Spatiotemporal Weather Downscaling (PSWD) as predicting future high-resolution global weather trajectories from recent coarse-resolution global states.
We addressed PSWD with a spatial-to-temporal scaffold-and-correct pipeline that reuses a shared pretrained backbone and reconstructs target trajectories with fixed numerical scaffolds and lightweight residual heads.
For parameter-efficient backbone adaptation, we proposed PROLA (Principal-Orthogonal Low-rank Adaptation), which splits a fixed rank budget across principal and orthogonal branches and rescales rank-1 optimizer updates using gradient signals.
Experiments show that PROLA outperforms low-rank PEFT baselines under matched low-rank budgets, indicating that preserving pretrained structure while reserving capacity for orthogonal corrections is effective for PSWD.

\begin{ack}
This research was partly supported by the ASTRA Project through the National Research Foundation (NRF) funded by the Ministry of Science and ICT (No. RS-2024-00439619, 90\%) and by the National Research Foundation of Korea (RS-2022-NR068758, 10\%). The authors declare no competing interests.
\end{ack}

\bibliographystyle{unsrtnat}
\bibliography{references}

@inproceedings{hu2022lora,
  title={{LoRA}: Low-rank adaptation of large language models},
  author={Hu, Edward J. and Shen, Yelong and Wallis, Phillip and Allen-Zhu, Zeyuan and Li, Yuanzhi and Wang, Shean and Wang, Lu and Chen, Weizhu},
  booktitle={ICLR},
  year={2022}
}

@inproceedings{dettmers2023qlora,
  title={{QLoRA}: Efficient finetuning of quantized {LLMs}},
  author={Dettmers, Tim and Pagnoni, Artidoro and Holtzman, Ari and Zettlemoyer, Luke},
  booktitle={NeurIPS},
  year={2023}
}

@inproceedings{valipour2023dylora,
  title={{DyLoRA}: Parameter-efficient tuning of pre-trained models using dynamic search-free low-rank adaptation},
  author={Valipour, Mojtaba and Rezagholizadeh, Mehdi and Kobyzev, Ivan and Ghodsi, Ali},
  booktitle={EACL},
  year={2023}
}

@inproceedings{kopiczko2023vera,
  title={{VeRA}: Vector-based random matrix adaptation},
  author={Kopiczko, Dawid J. and Blankevoort, Tijmen and Asano, Yuki M.},
  booktitle={ICLR},
  year={2024}
}

@inproceedings{zhang2023adalora,
  title={Adaptive budget allocation for parameter-efficient fine-tuning},
  author={Zhang, Qingru and Chen, Minshuo and Bukharin, Alexander and He, Pengcheng and Cheng, Yu and Chen, Weizhu and Zhao, Tuo},
  booktitle={ICLR},
  year={2023}
}

@inproceedings{hayou2024lora+,
  title={{LoRA+}: Efficient low rank adaptation of large models},
  author={Hayou, Soufiane and Ghosh, Nikhil and Yu, Bin},
  booktitle={ICML},
  year={2024}
}

@inproceedings{liu2024dora,
  title={{DoRA}: Weight-decomposed low-rank adaptation},
  author={Liu, Shih-Yang and Wang, Chien-Yi and Yin, Hongxu and Molchanov, Pavlo and Wang, Yu-Chiang Frank and Cheng, Kwang-Ting and Chen, Min-Hung},
  booktitle={ICML},
  year={2024}
}

@inproceedings{meng2024pissa,
  title={{PiSSA}: Principal singular values and singular vectors adaptation of large language models},
  author={Meng, Fanxu and Wang, Zhaohui and Zhang, Muhan},
  booktitle={NeurIPS},
  year={2024}
}

@inproceedings{xiong2026oplora,
  title={{OPLoRA}: Orthogonal projection {LoRA} prevents catastrophic forgetting during parameter-efficient fine-tuning},
  author={Xiong, Yifeng and Xie, Xiaohui},
  booktitle={AAAI},
  year={2026}
}

@article{pathak2022fourcastnet,
  title={{FourCastNet}: A global data-driven high-resolution weather model using adaptive {Fourier} neural operators},
  author={Pathak, Jaideep and Subramanian, Shashank and Harrington, Peter and Raja, Sanjeev and Chattopadhyay, Ashesh and Mardani, Morteza and Kurth, Thorsten and Hall, David and Li, Zongyi and Azizzadenesheli, Kamyar and others},
  journal={arXiv preprint arXiv:2202.11214},
  year={2022}
}

@article{keisler2022forecasting,
  title={Forecasting global weather with graph neural networks},
  author={Keisler, Ryan},
  journal={arXiv preprint arXiv:2202.07575},
  year={2022}
}

@article{GraphCast:lam2023learning,
  title={Learning skillful medium-range global weather forecasting},
  author={Lam, Remi and Sanchez-Gonzalez, Alvaro and Willson, Matthew and Wirnsberger, Peter and Fortunato, Meire and Alet, Ferran and Ravuri, Suman and Ewalds, Timo and Eaton-Rosen, Zach and Hu, Weihua and others},
  journal={Science},
  year={2023}
}

@article{bi2023accurate,
  title={Accurate medium-range global weather forecasting with {3D} neural networks},
  author={Bi, Kaifeng and Xie, Lingxi and Zhang, Hengheng and Chen, Xin and Gu, Xiaotao and Tian, Qi},
  journal={Nature},
  year={2023}
}

@article{chen2023fuxi,
  title={{FuXi}: A cascade machine learning forecasting system for 15-day global weather forecast},
  author={Chen, Lei and Zhong, Xiaohui and Zhang, Feng and Cheng, Yuan and Xu, Yinghui and Qi, Yuan and Li, Hao},
  journal={npj Clim. Atmos. Sci.},
  year={2023}
}

@article{chen2025fengwu,
  title={The operational medium-range deterministic weather forecasting can be extended beyond a 10-day lead time},
  author={Chen, Kang and Han, Tao and Ling, Fenghua and Gong, Junchao and Bai, Lei and Wang, Xinyu and Luo, Jing-Jia and Fei, Ben and Zhang, Wenlong and Chen, Xi and others},
  journal={Commun. Earth Environ.},
  year={2025}
}

@article{lang2024aifs,
  title={{AIFS}---{ECMWF}'s data-driven forecasting system},
  author={Lang, Simon and Alexe, Mihai and Chantry, Matthew and Dramsch, Jesper and Pinault, Florian and Raoult, Baudouin and Clare, Mariana C. A. and Lessig, Christian and Maier-Gerber, Michael and Magnusson, Linus and others},
  journal={arXiv preprint arXiv:2406.01465},
  year={2024}
}

@inproceedings{Stormer:nguyen2023scaling,
  title={Scaling transformer neural networks for skillful and reliable medium-range weather forecasting},
  author={Nguyen, Tung and Shah, Rohan and Bansal, Hritik and Arcomano, Troy and Maulik, Romit and Kotamarthi, Veerabhadra and Foster, Ian and Madireddy, Sandeep and Grover, Aditya},
  booktitle={NeurIPS},
  year={2024}
}

@article{NeuralGCM:kochkov2024neural,
  title={Neural general circulation models for weather and climate},
  author={Kochkov, Dmitrii and Yuval, Janni and Langmore, Ian and Norgaard, Peter and Smith, Jamie and Mooers, Griffin and Kl{\"{o}}wer, Milan and Lottes, James and Rasp, Stephan and D{\"{u}}ben, Peter and others},
  journal={Nature},
  year={2024}
}

@article{GenCast:price2025probabilistic,
  title={Probabilistic weather forecasting with machine learning},
  author={Price, Ilan and Sanchez-Gonzalez, Alvaro and Alet, Ferran and Andersson, Tom R. and El-Kadi, Andrew and Masters, Dominic and Ewalds, Timo and Stott, Jacklynn and Mohamed, Shakir and Battaglia, Peter and others},
  journal={Nature},
  year={2025}
}

@inproceedings{nguyen2023climax,
  title={{ClimaX}: A foundation model for weather and climate},
  author={Nguyen, Tung and Brandstetter, Johannes and Kapoor, Ashish and Gupta, Jayesh K. and Grover, Aditya},
  booktitle={ICML},
  year={2023}
}

@article{bodnar2025foundation,
  title={A foundation model for the {Earth} system},
  author={Bodnar, Cristian and Bruinsma, Wessel P. and Lucic, Ana and Stanley, Megan and Allen, Anna and Brandstetter, Johannes and Garvan, Patrick and Riechert, Maik and Weyn, Jonathan A. and Dong, Haiyu and others},
  journal={Nature},
  year={2025}
}

@article{gonzalez2023dl4ds,
  title={{DL4DS}---Deep learning for empirical downscaling},
  author={Gomez Gonzalez, Carlos Alberto},
  journal={Environmental Data Sci.},
  year={2023}
}

@inproceedings{vandal2017deepsd,
  title={{DeepSD}: Generating high resolution climate change projections through single image super-resolution},
  author={Vandal, Thomas and Kodra, Evan and Ganguly, Sangram and Michaelis, Andrew and Nemani, Ramakrishna and Ganguly, Auroop R.},
  booktitle={KDD},
  year={2017}
}

@article{wang2021deep,
  title={Deep learning for daily precipitation and temperature downscaling},
  author={Wang, Fang and Tian, Di and Lowe, Lisa and Kalin, Latif and Lehrter, John},
  journal={Water Resour. Res.},
  year={2021}
}

@article{rampal2022high,
  title={High-resolution downscaling with interpretable deep learning: Rainfall extremes over {New Zealand}},
  author={Rampal, Neelesh and Gibson, Peter B. and Sood, Abha and Stuart, Stephen and Fauchereau, Nicolas C. and Brandolino, Chris and Noll, Ben and Meyers, Tristan},
  journal={Weather Clim. Extremes},
  year={2022}
}

@article{leinonen2020stochastic,
  title={Stochastic super-resolution for downscaling time-evolving atmospheric fields with a generative adversarial network},
  author={Leinonen, Jussi and Nerini, Daniele and Berne, Alexis},
  journal={IEEE Trans. Geosci. Remote Sens.},
  year={2021}
}

@article{harris2022generative,
  title={A generative deep learning approach to stochastic downscaling of precipitation forecasts},
  author={Harris, Lucy and McRae, Andrew T. T. and Chantry, Matthew and Dueben, Peter D. and Palmer, Tim N.},
  journal={J. Adv. Model. Earth Syst.},
  year={2022}
}

@inproceedings{price2022increasing,
  title={Increasing the accuracy and resolution of precipitation forecasts using deep generative models},
  author={Price, Ilan and Rasp, Stephan},
  booktitle={AISTATS},
  year={2022}
}

@article{mardani2025residual,
  title={Residual corrective diffusion modeling for km-scale atmospheric downscaling},
  author={Mardani, Morteza and Brenowitz, Noah and Cohen, Yair and Pathak, Jaideep and Chen, Chieh-Yu and Liu, Cheng-Chin and Vahdat, Arash and Nabian, Mohammad Amin and Ge, Tao and Subramaniam, Akshay and others},
  journal={Commun. Earth Environ.},
  year={2025}
}

@article{rasp2024weatherbench,
  title={{WeatherBench 2}: A benchmark for the next generation of data-driven global weather models},
  author={Rasp, Stephan and Hoyer, Stephan and Merose, Alexander and Langmore, Ian and Battaglia, Peter and Russell, Tyler and Sanchez-Gonzalez, Alvaro and Yang, Vivian and Carver, Rob and Agrawal, Shreya and others},
  journal={J. Adv. Model. Earth Syst.},
  year={2024}
}

@inproceedings{bonev2023spherical,
  title={Spherical {Fourier} neural operators: Learning stable dynamics on the sphere},
  author={Bonev, Boris and Kurth, Thorsten and Hundt, Christian and Pathak, Jaideep and Baust, Maximilian and Kashinath, Karthik and Anandkumar, Anima},
  booktitle={ICML},
  year={2023}
}

@article{glawion2023spategan,
  title={{spateGAN}: Spatio-temporal downscaling of rainfall fields using a {cGAN} approach},
  author={Glawion, Luca and Polz, Julius and Kunstmann, Harald and Fersch, Benjamin and Chwala, Christian},
  journal={Earth Space Sci.},
  year={2023}
}

@article{glawion2025global,
  title={Global spatio-temporal {ERA5} precipitation downscaling to km and sub-hourly scale using generative {AI}},
  author={Glawion, Luca and Polz, Julius and Kunstmann, Harald and Fersch, Benjamin and Chwala, Christian},
  journal={npj Clim. Atmos. Sci.},
  year={2025}
}

@article{schmude2024prithvi,
  title={{Prithvi WxC}: Foundation model for weather and climate},
  author={Schmude, Johannes and Roy, Sujit and Trojak, Will and Jakubik, Johannes and Civitarese, Daniel Salles and Singh, Shraddha and Kuehnert, Julian and Ankur, Kumar and Gupta, Aman and Phillips, Christopher E. and others},
  journal={arXiv preprint arXiv:2409.13598},
  year={2024}
}

@inproceedings{cao2026taskadaptive,
  title={Task-adaptive parameter-efficient fine-tuning for weather foundation models},
  author={Cao, Shilei and Lin, Hehai and Cheng, Jiashun and Liu, Yang and Li, Guowen and Wang, Xuehe and Zheng, Juepeng and Liang, Haoyuan and Jin, Meng and Qin, Chengwei and others},
  booktitle={ICLR},
  year={2026}
}

@article{chen2023foundation,
  title={Foundation models for weather and climate data understanding: A comprehensive survey},
  author={Chen, Shengchao and Long, Guodong and Jiang, Jing and Liu, Dikai and Zhang, Chengqi},
  journal={arXiv preprint arXiv:2312.03014},
  year={2023}
}

@article{han2024parameter,
  title={Parameter-efficient fine-tuning for large models: A comprehensive survey},
  author={Han, Zeyu and Gao, Chao and Liu, Jinyang and Zhang, Jeff and Zhang, Sai Qian},
  journal={Transactions on Machine Learning Research},
  year={2024}
}

@inproceedings{li2024loftq,
  title={{LoftQ}: {LoRA}-fine-tuning-aware quantization for large language models},
  author={Li, Yixiao and Yu, Yifan and Liang, Chen and He, Pengcheng and Karampatziakis, Nikos and Chen, Weizhu and Zhao, Tuo},
  booktitle={ICLR},
  year={2024}
}

@inproceedings{yang2024corda,
  title={{CorDA}: Context-oriented decomposition adaptation of large language models for task-aware parameter-efficient fine-tuning},
  author={Yang, Yibo and Li, Xiaojie and Zhou, Zhongzhu and Song, Shuaiwen L. and Wu, Jianlong and Nie, Liqiang and Ghanem, Bernard},
  booktitle={NeurIPS},
  year={2024}
}

@inproceedings{wang2023orthogonal,
  title={Orthogonal subspace learning for language model continual learning},
  author={Wang, Xiao and Chen, Tianze and Ge, Qiming and Xia, Han and Bao, Rong and Zheng, Rui and Zhang, Qi and Gui, Tao and Huang, Xuanjing},
  booktitle={EMNLP Findings},
  year={2023}
}

@article{oyama2023deep,
  title={Deep generative model super-resolves spatially correlated multiregional climate data},
  author={Oyama, Norihiro and Ishizaki, Noriko N. and Koide, Satoshi and Yoshida, Hiroaki},
  journal={Sci. Rep.},
  year={2023}
}

@article{wurth2019minute,
  title={Minute-scale forecasting of wind power---Results from the collaborative workshop of {IEA Wind Task} 32 and 36},
  author={W{\"{u}}rth, Ines and Valldecabres, Laura and Simon, Elliot and M{\"{o}}hrlen, Corinna and Uzuno{\u{g}}lu, Bahri and Gilbert, Ciaran and Giebel, Gregor and Schlipf, David and Kaifel, Anton},
  journal={Energies},
  year={2019}
}

@article{ravuri2021skillful,
  title={Skilful precipitation nowcasting using deep generative models of radar},
  author={Ravuri, Suman and Lenc, Karel and Willson, Matthew and Kangin, Dmitry and Lam, Remi and Mirowski, Piotr and Fitzsimons, Megan and Athanassiadou, Maria and Kashem, Sheleem and Madge, Sam and others},
  journal={Nature},
  year={2021}
}

@article{sharman2014description,
  title={Description and derived climatologies of automated in situ eddy-dissipation-rate reports of atmospheric turbulence},
  author={Sharman, R. D. and Cornman, L. B. and Meymaris, G. and Pearson, J. and Farrar, T.},
  journal={J. Appl. Meteorol. Climatol.},
  year={2014}
}

@inproceedings{loshchilov2017decoupled,
  title={Decoupled weight decay regularization},
  author={Loshchilov, Ilya and Hutter, Frank},
  booktitle={ICLR},
  year={2019}
}

@inproceedings{wang2026complementary,
  title={Complementary subspace low-rank adaptation of vision-language models for few-shot classification},
  author={Wang, Zhongqi and Dai, Jia and Li, Kai and Li, Xu and Guo, Yanmeng},
  booktitle={ICASSP},
  year={2026}
}

@inproceedings{wang2025hdpissa,
  title={{HD-PiSSA}: High-rank distributed orthogonal adaptation},
  author={Wang, Yiding and Meng, Fanxu and Zhang, Xuefeng and Jiang, Fan and Tang, Pingzhi and Zhang, Muhan},
  booktitle={EMNLP},
  year={2025}
}

@inproceedings{wu2026efficient,
  title={Efficient orthogonal fine-tuning with principal subspace adaptation},
  author={Wu, Fei and Hu, Jia and Min, Geyong and Wang, Shiqiang},
  booktitle={ICLR},
  year={2026}
}

@article{zhao2026loft,
  title={{LOFT}: Low-rank orthogonal fine-tuning via task-aware support selection},
  author={Zhao, Lanxin and Mishra, Bamdev and Jawanpuria, Pratik and Lin, Lequan and Shi, Dai and Gao, Junbin and Han, Andi},
  journal={arXiv preprint arXiv:2605.11872},
  year={2026}
}

@article{chen2026hiplora,
  title={{HiP-LoRA}: Budgeted spectral plasticity for robust low-rank adaptation},
  author={Chen, Lixian and Tan, Jianhong},
  journal={arXiv preprint arXiv:2604.17751},
  year={2026}
}

@article{couairon2026archesweathergen,
  title={{ArchesWeatherGen}: Skillful and compute-efficient probabilistic weather forecasting with machine learning},
  author={Couairon, Guillaume and Singh, Renu and Charantonis, Anastase and Lessig, Christian and Monteleoni, Claire},
  journal={Sci. Adv.},
  year={2026}
}

@misc{qwen2026qwen35,
    title = {Qwen3.5: Towards Native Multimodal Agents},
    url = {https://qwen.ai/blog?id=qwen3.5},
    author = {Qwen Team},
    month = {February},
    year = {2026}
}

@inproceedings{li2023bird,
  title={Can {LLM} already serve as a database interface? {A} {BIg} bench for large-scale database grounded text-to-{SQLs}},
  author={Li, Jinyang and Hui, Binyuan and Qu, Ge and Yang, Jiaxi and Li, Binhua and Li, Bowen and Wang, Bailin and Qin, Bowen and Geng, Ruiying and Huo, Nan and others},
  booktitle={NeurIPS},
  year={2023}
}

@article{ridal2024cerra,
  title={{CERRA}, the {Copernicus European Regional Reanalysis} system},
  author={Ridal, Martin and Bazile, Eric and Le Moigne, Patrick and Randriamampianina, Roger and Schimanke, Semjon and Andrae, Ulf and Berggren, Lars and Brousseau, Pierre and Dahlgren, Per and Edvinsson, Lisette and others},
  journal={Q. J. R. Meteorol. Soc.},
  year={2024}
}

@article{hersbach2020era5,
  title={The {ERA5} global reanalysis},
  author={Hersbach, Hans and Bell, Bill and Berrisford, Paul and Hirahara, Shoji and Hor{\'{a}}nyi, Andr{\'{a}}s and Mu{\~{n}}oz-Sabater, Joaqu{\'{i}}n and Nicolas, Julien and Peubey, Carole and Radu, Raluca and Schepers, Dinand and others},
  journal={Q. J. R. Meteorol. Soc.},
  year={2020}
}

@article{rodwell2010equitable,
  title={A new equitable score suitable for verifying precipitation in numerical weather prediction},
  author={Rodwell, Mark J. and Richardson, David S. and Hewson, Tim D. and Haiden, Thomas},
  journal={Q. J. R. Meteorol. Soc.},
  year={2010}
}

@article{shazeer2020glu,
  title={{GLU} variants improve transformer},
  author={Shazeer, Noam},
  journal={arXiv preprint arXiv:2002.05202},
  year={2020}
}

\newpage
\appendix
The appendix is organized into the following sections.
\begin{outline}
    \1 Appendix \ref{app:cross_subspace}: Additional Analysis of Update Geometry
    \1 Appendix \ref{app:allocation_diagnostics}: Details of Gradient-Guided Scaling of Optimizer Updates
    \1 Appendix \ref{app:experimental_setup}: Detailed Experimental Setup
    \1 Appendix \ref{app:metrics}: Evaluation Metrics
    \1 Appendix \ref{app:implementation}: Implementation Details
    \1 Appendix \ref{app:peft_config}: PROLA and Baseline Configurations
    \1 Appendix \ref{app:full_results}: Additional Quantitative Results
    \1 Appendix \ref{app:qualitative}: Qualitative Results
    \1 Appendix \ref{app:additional_ablations}: Additional Ablations
    \1 Appendix \ref{app:limitation}: Limitations and Future Work
    \1 Appendix \ref{app:impact}: Broader Impacts

\end{outline}

\section{Additional Analysis of Update Geometry}
\label{app:cross_subspace}

This appendix provides the reference subspaces, block definitions, and energy aggregation used in Section~\ref{sec:update_diagnostics}.

\textbf{Blockwise diagnostic.}
We compare the final checkpoints of Full FT, PiSSA, and PROLA on the same set \(\mathcal A\) of 96 attention weight matrices (QKV and output projections).
For each \(\ell\in\mathcal A\), we analyze the effective update
\begin{equation}
\Delta W_\ell^{\mathrm{eff}}
=
W_\ell^{\mathrm{eff}}-W_{\ell,0}.
\end{equation}
For this diagnostic, we evaluate the decomposition in Eq.~\eqref{eq:blockwise_update_decomposition} using the same top-\(r\) reference subspaces for all methods, with \(r=8\).
Let \(U_{r,\ell}\) and \(V_{r,\ell}\) contain the top-\(r\) left and right singular vectors of \(W_{\ell,0}\), respectively.
The projectors are
\begin{equation}
P_\ell=U_{r,\ell}U_{r,\ell}^{\top},
\qquad
Q_\ell=V_{r,\ell}V_{r,\ell}^{\top},
\qquad
P_{\perp,\ell}=I-P_\ell,
\qquad
Q_{\perp,\ell}=I-Q_\ell.
\end{equation}
The principal, cross, and orthogonal blocks are
\begin{align}
\Delta W_\ell^{\mathrm{prin}}
&=
P_\ell\Delta W_\ell^{\mathrm{eff}}Q_\ell,
\\
\Delta W_\ell^{\mathrm{cross}}
&=
P_\ell\Delta W_\ell^{\mathrm{eff}}Q_{\perp,\ell}
+
P_{\perp,\ell}\Delta W_\ell^{\mathrm{eff}}Q_\ell,
\\
\Delta W_\ell^{\mathrm{orth}}
&=
P_{\perp,\ell}\Delta W_\ell^{\mathrm{eff}}Q_{\perp,\ell}.
\end{align}

PROLA constructs its orthogonal branch using the top-\(r_p\) directions \(U_{p,\ell},V_{p,\ell}\), with \(r_p=4\):
\begin{equation}
\Delta W_{o,\ell}
=
s_{o,\ell}
(I-U_{p,\ell}U_{p,\ell}^{\top})
B_{o,\ell}A_{o,\ell}
(I-V_{p,\ell}V_{p,\ell}^{\top}).
\end{equation}
The branch therefore satisfies
\begin{equation}
U_{p,\ell}^{\top}\Delta W_{o,\ell}=0,
\qquad
\Delta W_{o,\ell}V_{p,\ell}=0.
\end{equation}
These identities concern the top-\(r_p\) construction subspaces.
Under the top-\(r\) diagnostic partition, this branch may also contribute to the principal and cross blocks.
The freely trainable principal branch may contribute to the orthogonal block.
Figure~\ref{fig:update_diagnostics} measures blocks of the combined effective update.

\textbf{Normalized block energy.}
For each block \(b\in\{\mathrm{prin},\mathrm{cross},\mathrm{orth}\}\), we sum its squared Frobenius norm across adapted matrices and normalize by the total update energy:
\begin{equation}
\bar e_b =
\frac{\sum_{\ell\in\mathcal A}\|\Delta W_\ell^b\|_F^2}
{\sum_{\ell\in\mathcal A}\|\Delta W_\ell^{\mathrm{eff}}\|_F^2}.
\end{equation}

\section{Details of Gradient-Guided Scaling of Optimizer Updates}
\label{app:allocation_diagnostics}

This appendix provides additional details on the gradient-guided
scaling of optimizer updates introduced in
Section~\ref{sec:step-allocation}.
The scaling is applied to the trainable low-rank factors of each
adapted layer \(\ell\in\mathcal A\).
Let \(B^g_\ell\) and \(A^g_\ell\) denote the low-rank factors of
branch \(g\in\{p,o\}\) at layer \(\ell\), where \(p\) and \(o\)
denote the principal and orthogonal branches, respectively.
The \(k\)-th rank-1 component is represented by the factor pair
\((\mathbf b^g_{\ell,k},\mathbf a^g_{\ell,k})\), where
\(\mathbf b^g_{\ell,k}\) is the \(k\)-th column of \(B^g_\ell\)
and \((\mathbf a^g_{\ell,k})^\top\) is the \(k\)-th row of
\(A^g_\ell\).
For the orthogonal branch, this component is defined before applying its fixed projectors, as specified in Appendix~\ref{app:peft_config}.

The component signal and branch signal are
\begin{equation}
u^g_{\ell,k}
=
\left\|
\nabla_{\mathbf b^g_{\ell,k}}\mathcal L_{ST}
\right\|_2
+
\left\|
\nabla_{\mathbf a^g_{\ell,k}}\mathcal L_{ST}
\right\|_2,
\qquad
\mu^g_\ell
=
\frac{1}{r_g}
\sum_{k=1}^{r_g}
u^g_{\ell,k}.
\end{equation}
They are normalized into the component score and branch score:
\begin{equation}
\pi^g_{\ell,k}
=
\frac{u^g_{\ell,k}+\epsilon}
{\sum_{j=1}^{r_g}(u^g_{\ell,j}+\epsilon)},
\qquad
\rho^g_\ell
=
\frac{\mu^g_\ell+\epsilon}
{\sum_{g'\in\{p,o\}}(\mu^{g'}_\ell+\epsilon)} .
\end{equation}
The rank-1 update scale is
\begin{equation}
\alpha^g_{\ell,k}
=
r \times
\operatorname{sg}\!\left(
\rho^g_\ell \pi^g_{\ell,k}
\right),
\qquad
r=r_p+r_o,
\end{equation}
where \(\epsilon>0\) is a numerical constant and
\(\operatorname{sg}(\cdot)\) denotes stop-gradient.
The stop-gradient operation treats the scores as measurements
of the current training signal and prevents gradients from being
propagated through the score computation.
For single-branch variants, the branch sums and normalizations
are restricted to the existing branch, for which
\(\rho^g_\ell=1\).

We apply \(\alpha^g_{\ell,k}\) after the base optimizer has
formed its update.
Let
\(\widetilde{\Delta}\mathbf b^g_{\ell,k}\) and
\(\widetilde{\Delta}\mathbf a^g_{\ell,k}\)
denote the optimizer updates before scaling.
The applied updates are
\begin{equation}
\Delta\mathbf b^g_{\ell,k}
=
\alpha^g_{\ell,k}
\widetilde{\Delta}\mathbf b^g_{\ell,k},
\qquad
\Delta\mathbf a^g_{\ell,k}
=
\alpha^g_{\ell,k}
\widetilde{\Delta}\mathbf a^g_{\ell,k}.
\end{equation}
Thus, gradient-guided scaling rescales the applied parameter
updates after the base optimizer computes them, without changing
the base optimizer's raw-update computation.

\textbf{Mean-preserving property.}
Since \(\sum_{k=1}^{r_g}\pi^g_{\ell,k}=1\) and
\(\sum_{g\in\{p,o\}}\rho^g_\ell=1\), we have
\begin{equation}
\frac{1}{r}
\sum_{g\in\{p,o\}}
\sum_{k=1}^{r_g}
\alpha^g_{\ell,k}
=
\frac{1}{r}
\sum_{g\in\{p,o\}}
\sum_{k=1}^{r_g}
r\rho^g_\ell\pi^g_{\ell,k}
=
1.
\end{equation}
Therefore, the multiplicative scales have unit mean across the
\(r\) rank-1 components, while their relative values redistribute
update strength.

For the scaling variants in Table~\ref{tab:ablation_prola},
\(\rho\)-only scaling uses uniform component allocation within
each branch, giving
\(\alpha^g_{\ell,k}=(r/r_g)\operatorname{sg}(\rho^g_\ell)\).
The \(\pi\)-only variant fixes branch allocation at \(r_g/r\),
giving
\(\alpha^g_{\ell,k}=r_g\operatorname{sg}(\pi^g_{\ell,k})\).
Both variants retain the two branches with \(r_p=r_o=4\);
disabling scaling sets all \(\alpha^g_{\ell,k}\) to one.

\textbf{Component concentration diagnostic.}
We examine how gradient-guided scaling distributes optimizer updates across rank-1 components within each layer.
Using the mean-preserving property, we define the normalized
update-scale distribution as
\begin{equation}
\omega^g_{\ell,k}
=
\frac{\alpha^g_{\ell,k}}{r}.
\end{equation}
The normalized entropy is
\begin{equation}
H_\ell =
-\frac{1}{\log r}
\sum_{g\in\{p,o\}}\sum_{k=1}^{r_g}
\omega_{\ell,k}^g \log \omega_{\ell,k}^g.
\end{equation}

At a fixed trained checkpoint, we compute \(H_\ell\) separately for each validation batch and average over batches.
Within each attention block, we then average the entropies of the QKV and output projections.

\begin{figure}[!t]
\centering
\includegraphics[width=0.71\linewidth]{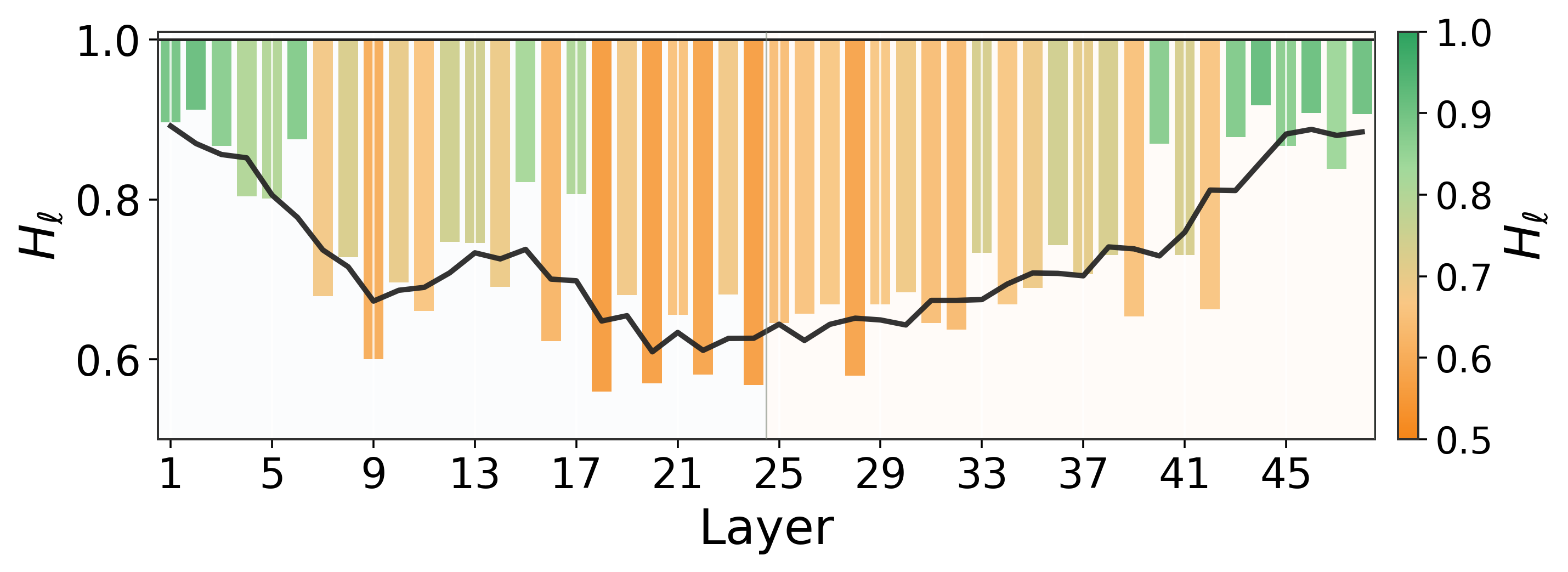}
\caption{\textbf{Layer-wise concentration of rank-1 update scales.}
Each bar extends from uniform scaling (\(H_\ell=1\)) to the mean normalized entropy of one attention block. Blocks are ordered from encoder to decoder.}
\label{fig:allocation_concentration}
\end{figure}

Figure~\ref{fig:allocation_concentration} shows that mean entropy falls below the uniform value of \(1\) in every block, with the largest departures in the middle layers.
Even the lowest block average, approximately \(0.56\), remains well above the single-component limit of \(0\).
These averages indicate that scaling departs from uniform allocation while remaining distributed across multiple rank-1 components, including in the middle layers.

\section{Detailed Experimental Setup}
\label{app:experimental_setup}

For the main global PSWD experiments, we use
Aurora~\cite{bodnar2025foundation} as the pretrained weather
backbone, starting from the same pretrained checkpoint%
\footnote{Copyright (c) 2024 Microsoft Corporation.
Licensed under MIT license} for all PEFT variants.
The experiments use the WeatherBench~2 dataset%
\footnote{Copyright (c) 2023 Google LLC.
Licensed under Apache-2.0 license}~\cite{rasp2024weatherbench}
at two resolutions: \(5.625^\circ\) data available at 6-hour
intervals for low-resolution inputs, and \(1.5^\circ\) data
available at 1-hour intervals for high-resolution targets.
The input lags are $\mathcal{H}_c=\{0,\delta_c\}$ with $\delta_c=6\mathrm{h}$.
The target leads are $\mathcal{H}_f=\{1\mathrm{h},\ldots,6\mathrm{h}\}$,
with $\delta_f=1\mathrm{h}$ and $h_{\mathrm{end}}=6\mathrm{h}$.

The \(5.625^\circ\) inputs use a \(32\times64\) grid.
We crop the raw \(121\times240\) target grid to \(120\times240\) to accommodate the backbone's patch size of 4.
We train on 2007-2016, validate on 2017, and test on 2018.
We use 68 weather variables in total.
The surface variables are \(T2M\), \(U10\), and \(V10\), and
the atmospheric variables are geopotential \(Z\), temperature
\(T\), specific humidity \(Q\), zonal wind \(U\), and meridional
wind \(V\) at 13 pressure levels:
\(50,100,150,200,250,300,400,500,600,700,850,925,\)
and \(1000\) hPa.
This gives \(5\times13=65\) atmospheric channels and
3 surface channels.

\textbf{ERA5-to-CERRA downscaling.}
We additionally use \(5.625^\circ\) ERA5 inputs to predict
CERRA fields over a \(128\times128\) crop of Central Europe
at approximately \(5.5\) km spacing~\cite{hersbach2020era5,ridal2024cerra}.
All months of 2012--2016 are used for training, 2017 for
validation, and 2018 for testing.
Inputs at $t-\delta_c$ and $t$, with $\delta_c=6\mathrm{h}$, are paired
with targets at $t+h$ for $h\in\mathcal{H}_f=\{3\mathrm{h},6\mathrm{h}\}$
($\delta_f=3\mathrm{h}$). All required times lie within the corresponding split.
ERA5 inputs are bilinearly interpolated to the geographic
coordinates of the CERRA crop.
The 13 channels comprise \(T2M\), \(U10\), and \(V10\),
together with \(Z,T,Q,U,V\) at 500 and 850 hPa.
Wind fields are converted to geographic components where
needed, and relative humidity is converted to specific humidity.

\textbf{Regional precipitation forecasting.}
We evaluate regional precipitation forecasting using a
\(1.5^\circ\), \(40\times40\) ERA5 grid covering
\(0^\circ\)--\(58.5^\circ\)N and
\(75^\circ\)--\(133.5^\circ\)E.
We train on 2007--2016, validate on 2017, and test on 2018,
keeping input and target frames within each split.
Inputs at \(t-12\mathrm{h}\) and \(t\) are used to predict
6-hour accumulated precipitation at 12h, 24h, and 36h lead times.
Each target represents the accumulation over the six hours
ending at its valid time.
Following the regional precipitation
protocol~\cite{cao2026taskadaptive}, the inputs include surface
temperature, horizontal winds, mean sea-level pressure, and
precipitation, together with upper-air temperature, horizontal
winds, relative humidity, and geopotential at the 13 pressure
levels listed above.

\textbf{ArchesWeather backbone.}
We also evaluate the scaffold-and-correct pipeline with
pretrained ArchesWeather-M~\cite{couairon2026archesweathergen} at the same \(5.625^\circ\) input
and \(1.5^\circ\) target resolutions as the main benchmark.
To match the backbone's forecasting interval, the inputs are at
$t-\delta_c$ and $t$ with $\delta_c=24\mathrm{h}$.
The target leads are $\mathcal{H}_f=\{6\mathrm{h},12\mathrm{h},18\mathrm{h},24\mathrm{h}\}$,
with $\delta_f=6\mathrm{h}$ and $h_{\mathrm{end}}=24\mathrm{h}$.
We train on 2007--2016 and evaluate on 2018.

\textbf{Additional resolution gap.}
We evaluate an additional \(5.625^\circ\)-to-\(2.8125^\circ\)
setting, with a \(64\times128\) target grid obtained by
conservative remapping of the fine-resolution reference fields.
Inputs are at $t-\delta_c$ and $t$ with $\delta_c=6\mathrm{h}$.
The target leads are $\mathcal{H}_f=\{3\mathrm{h},6\mathrm{h}\}$ with $\delta_f=3\mathrm{h}$.

\textbf{Text-to-SQL adaptation.}
\phantomsection\label{app:text_to_sql}
We evaluate Qwen3.5-9B~\cite{qwen2026qwen35} on BIRD~\cite{li2023bird} to assess transfer to schema-conditioned SQL generation. We use database-disjoint training, validation, and evaluation splits, with the corrected development set used for evaluation.

\section{Evaluation Metrics}
\label{app:metrics}

For the main global PSWD benchmark, we report latitude-weighted root mean squared error (RMSE) and anomaly correlation coefficient (ACC).
Let $v$ denote a reported variable, $h\in\mathcal{H}_f$ a target lead time,
and $t_n$ the anchor time of sample $n$.
At latitude index $i$ and longitude index $j$, $\hat y^{(v,h)}_{n,i,j}$
and $y^{(v,h)}_{n,i,j}$ denote the denormalized values of variable $v$ in
$\hat{\mathbf{y}}^{ST}_{t_n+h}$ and $\mathbf{y}^{\mathrm{HR}}_{t_n+h}$,
respectively. Here $(H,W)=(H_{\mathrm{HR}},W_{\mathrm{HR}})$.
We use latitude weights
\begin{equation}
w_i
=
\frac{\cos(\varphi_i)}
{\frac{1}{H}\sum_{i'=1}^{H}\cos(\varphi_{i'})},
\end{equation}
where \(\varphi_i\) is the latitude at row \(i\).
Following \citep{cao2026taskadaptive}, we report the mean of sample-wise latitude-weighted RMSE values:
\begin{equation}
\mathrm{RMSE}_{v,h}
=
\frac{1}{N}
\sum_{n=1}^{N}
\sqrt{
\frac{1}{HW}
\sum_{i=1}^{H}
\sum_{j=1}^{W}
w_i
\left(
\hat y^{(v,h)}_{n,i,j}
-
y^{(v,h)}_{n,i,j}
\right)^2
}.
\end{equation}

For ACC, we subtract a temporal mean field from both predictions and targets.
Let \(c^{(v)}_{i,j}\) denote the temporal mean of the denormalized high-resolution ground-truth fields for variable \(v\):
\begin{equation}
c^{(v)}_{i,j}
=
\frac{1}{|\mathcal T_{\mathrm{clim}}|}
\sum_{\tau\in\mathcal T_{\mathrm{clim}}}
y^{(v)}_{\tau,i,j},
\end{equation}
where \(\mathcal T_{\mathrm{clim}}\) contains the training-period target valid times from 2007--2016.
We compute sample-wise anomaly correlations and then average over samples:
\begin{equation}
\mathrm{ACC}_{v,h}
=
\frac{1}{N}
\sum_{n=1}^{N}
\frac{
\sum_{i=1}^{H}
\sum_{j=1}^{W}
w_i
\left(\hat y^{(v,h)}_{n,i,j}-c^{(v)}_{i,j}\right)
\left(y^{(v,h)}_{n,i,j}-c^{(v)}_{i,j}\right)
}{
\max\!\left\{
\sqrt{
\sum_{i=1}^{H}
\sum_{j=1}^{W}
w_i
\left(\hat y^{(v,h)}_{n,i,j}-c^{(v)}_{i,j}\right)^2
}
\sqrt{
\sum_{i=1}^{H}
\sum_{j=1}^{W}
w_i
\left(y^{(v,h)}_{n,i,j}-c^{(v)}_{i,j}\right)^2
}
,\epsilon
\right\}
}.
\end{equation}
Here, \(\epsilon=10^{-8}\).
Weather fields are evaluated after denormalization: RMSE is in the corresponding physical units, and ACC is dimensionless.
For CERRA, we use the same sample-wise RMSE aggregation on the regional target grid.

\textbf{Seasonal and spatial-tail evaluation.}
\phantomsection\label{app:tail_metrics}
Seasonal results group test cases into DJF, MAM, JJA, and SON according to the target valid time.
For spatial-tail evaluation, we identify the upper and lower 1\% of grid points within each target field using its 99th and 1st percentiles.
Tail RMSE uses latitude weighting over the target-selected grid points.
For the Brier skill score (BSS), forecast and observed event indicators use the same target-derived threshold, and the reference assigns event probability \(0.01\) to every grid point.
We report \(1-\mathrm{BS}/\mathrm{BS}_{\mathrm{ref}}\), with latitude weighting for both scores.

\textbf{Regional precipitation metrics.}
We report RMSE, ACC, and the Stable Equitable Error in Probability Space (SEEPS)~\cite{rodwell2010equitable} for 6-hour accumulated precipitation at each evaluated lead time.
Precipitation RMSE is the square root of the latitude-weighted mean squared error over all evaluated times and grid points, and is reported in cm.
ACC is averaged over samples using a training-period climatology indexed by day of year and hour of day.
SEEPS divides precipitation into dry, light, and heavy categories using a dry threshold of \(0.1\) mm and the local two-thirds quantile of wet precipitation.
We report latitude-weighted mean SEEPS over locations with climatological dry fractions between \(0.1\) and \(0.85\).
Lower SEEPS indicates better performance.

\textbf{Spherical harmonic spectra.}
Following spectral evaluation in global weather forecasting~\cite{GenCast:price2025probabilistic,couairon2026archesweathergen}, we decompose each predicted and target field into orthonormal spherical harmonics.
For fixed $v$ and $h$, write $\hat y_n$ and $y_n$ for the predicted and target
fields of sample $n$ at valid time $t_n+h$. For a field $y_n$ with coefficients
$a_{n,\ell,m}$, the power at degree $\ell$ is
\begin{equation}
\mathcal P_\ell(y_n)
=\frac{1}{4\pi}\sum_{m=-\ell}^{\ell}|a_{n,\ell,m}|^2.
\end{equation}
We compute power separately for each case and plot the ratio of mean prediction power to mean target power,
\begin{equation}
\frac{N^{-1}\sum_{n=1}^{N}\mathcal P_\ell(\hat y_n)}
{N^{-1}\sum_{n=1}^{N}\mathcal P_\ell(y_n)}.
\end{equation}
A ratio of one denotes matching power at that spatial scale.
Figure~\ref{fig:spectral_t2m} uses T2M at 6h from the seed-0 models over all 1,458 test cases in 2018.
We evaluate degrees \(\ell=1,\ldots,59\) on the native \(120\times240\) grid and convert degree to wavelength as \(2\pi R/\sqrt{\ell(\ell+1)}\), with Earth's radius \(R=6371\) km.

\textbf{Text-to-SQL evaluation.}
For BIRD~\cite{li2023bird}, we report execution accuracy (EX), the percentage of predicted SQL queries whose execution results match those of the reference queries on the corresponding databases. We compute EX using a custom SQLite evaluator with unordered result-set comparison.

\textbf{Aggregation across seeds.}
For the main comparisons, we compute each test metric separately for seeds 0, 1, and 2, and report the mean and sample standard deviation across these three values.
The standard deviation uses denominator \(n-1\).

\section{Implementation Details}
\label{app:implementation}

The main Aurora training configuration uses two NVIDIA RTX 3090 GPUs.
For PEFT and Head only experiments, we use per-GPU batch size \(4\) with gradient accumulation over \(4\) steps, giving an effective global batch size of \(32\).
For Full FT, we use per-GPU batch size \(2\) with gradient accumulation over \(8\) steps, giving the same effective global batch size of \(32\).
Models in this benchmark are trained for \(10\) epochs.
We use AdamW~\cite{loshchilov2017decoupled} with \((\beta_1,\beta_2)=(0.9,0.999)\), \(\epsilon=10^{-8}\), and weight decay \(0.05\).
The learning rate follows a step-level cosine schedule with linear warmup.
For PEFT and Head only experiments, the base learning rate is \(5\times10^{-4}\), the warmup learning rate is \(10^{-6}\), the minimum learning rate is \(10^{-6}\), and warmup lasts for one epoch.
For Full FT, the backbone learning rate is \(5\times10^{-5}\), with warmup and minimum learning rates \(10^{-7}\).
The correction heads in Full FT use the learning rate \(5\times10^{-4}\).
Gradient norm clipping is applied with a maximum norm of \(1.0\).
The main comparisons use seeds 0, 1, and 2, with the same seed set for every method.
We evaluate the final checkpoint after the prescribed training budget.

\textbf{Training objective.}
For the main Aurora benchmark, we use a weighted L1 loss on fields normalized with Aurora's variable-specific statistics~\cite{bodnar2025foundation}.
At each lead time, the surface term sums the variable-weighted spatial mean absolute errors and is multiplied by \(1/4\).
The upper-air term averages spatial mean absolute errors over pressure levels and then takes a weighted sum across variables.
The combined loss is divided by the number of variable types, \(3+5=8\).
The surface weights for \((T2M,U10,V10)\) are \((3.0,0.77,0.66)\), and the upper-air weights for \((Z,T,Q,U,V)\) are \((3.0,1.5,0.6,0.77,0.54)\), shared across pressure levels.
We average $\mathcal{L}_{ST}$ in Eq.~\eqref{eq:loss} over training examples.
Its six lead terms supervise the final predictions $\hat{\mathbf{y}}^{ST}_{t+h}$
for $h\in\mathcal{H}_f$. The spatial endpoint $\hat{\mathbf{y}}^{S}_{t+h_{\mathrm{end}}}$
receives supervision through the temporal stage.

\textbf{Regional weather experiments.}
ERA5-to-CERRA adaptation uses \(10\) epochs, an effective global batch size of \(8\), and a learning rate of \(5\times10^{-4}\).
Regional precipitation forecasting uses \(10\) epochs, an effective global batch size of \(32\), and a learning rate of \(3.75\times10^{-4}\).
Both settings use AdamW with \((\beta_1,\beta_2)=(0.9,0.999)\), weight decay \(0.05\), and a cosine learning-rate schedule with one epoch of linear warmup.
These comparisons use seed 0 and the final checkpoint after 10 epochs.
Within each task, LoRA, PiSSA, OPLoRA, and PROLA adapt the same attention projections and train the same task-specific heads.

\textbf{ArchesWeather experiments.}
We fine-tune the pretrained ArchesWeather-M backbone~\cite{couairon2026archesweathergen} for one epoch, corresponding to 7,300 optimizer steps, with an effective global batch size of \(2\) and a learning rate of \(3\times10^{-4}\).
We use AdamW with \((\beta_1,\beta_2)=(0.9,0.98)\), weight decay \(0.05\), and 250 warmup steps followed by cosine decay.
All compared methods adapt the attention projections and train the correction heads, using seed 0 and the final checkpoint.

\textbf{Text-to-SQL adaptation.}
We compare the unadapted model with LoRA, PiSSA, OPLoRA, and PROLA. All adapters target attention and MLP projections with rank 16 and 40.1M trainable parameters; PROLA uses \(r_p=r_o=8\). Each adapter is trained for one epoch with a learning rate of \(2\times10^{-5}\) and seed 42.

\section{PROLA and Baseline Configurations}
\label{app:peft_config}

For the main Aurora comparison, PROLA and all LoRA-style PEFT baselines fine-tune the linear projections in the backbone's self-attention operations.
All other pretrained backbone parameters remain frozen unless the method is Full FT.
The spatial correction head \(C_s\) and temporal correction head \(C_t\) are trained in all methods in this comparison except where explicitly ablated.
Thus, PEFT baseline comparisons use the same prediction pipeline and differ only in the backbone adaptation strategy.

The default main comparison uses rank \(r=8\) for all fixed-rank PEFT baselines.
LoRA~\cite{hu2022lora} uses the standard zero-output initialization.
Unless otherwise stated, all adapters use no adapter dropout and do not train bias terms.
DoRA~\cite{liu2024dora} uses the same rank, target modules, dropout, and bias setting as LoRA, with the additional magnitude parameter trained following the original parameterization.
PiSSA~\cite{meng2024pissa} initializes the low-rank factors from the top-\(r\) singular components of each pretrained weight and residualizes the frozen base weight to preserve the pretrained function at initialization.
OPLoRA~\cite{xiong2026oplora} uses fixed projectors onto the complements of the top-8 left and right singular subspaces of each pretrained weight, with the same trainable rank, target modules, dropout, and bias setting as LoRA.

For PROLA, we use the same total low-rank budget \(r=8\), split evenly into \(r_p=4\) for the principal branch and \(r_o=4\) for the orthogonal branch.
The reference singular directions and orthogonal projectors are computed from the pretrained weights and kept fixed during adaptation, while the principal factors \(A_p,B_p\) remain trainable.
We residualize the frozen base weight as in Eq.~\eqref{eq:residual_weight},
set $B_{o,0}=0$, and sample the entries of $A_{o,0}$ from
$\mathcal{N}(0,\sigma_o^2)$ with $\sigma_o=0.02$,
preserving the pretrained function.
The complete PROLA configuration uses gradient-guided update scaling
(Appendix~\ref{app:allocation_diagnostics}).
With the principal branch removed ($r_p=0$, $r_o=8$),
the orthogonal-only ablation retains fixed projectors onto the
complements of each pretrained weight's top-8 left and right
singular subspaces.
We set \(s_g=\alpha/r_g\) for each PROLA branch \(g\in\{p,o\}\) and \(s=\alpha/r\) for LoRA, DoRA, PiSSA, and OPLoRA, with \(\alpha=1\) for all five methods.

AdaLoRA~\cite{zhang2023adalora} uses $\alpha=16$,
$r_{\mathrm{init}}=12$, $r_{\mathrm{target}}=8$,
mask updates every 10 optimizer steps,
importance EMA coefficients $\beta_1=\beta_2=0.85$,
and orthogonal regularization weight $10^{-3}$.
Initial and final warmups last one and two epochs of optimizer
steps, respectively.
Its target modules, dropout, and bias settings match the
fixed-rank baselines.

Head only trains $C_s$ and $C_t$ with the backbone frozen;
Full FT also trains the backbone, using a smaller learning rate
for its pretrained weights.
Table~\ref{tab:aurora_trainable_params} reports the trainable-parameter
counts for the main Aurora configuration.
For PROLA, these comprise $2{,}850{,}816$ backbone PEFT parameters
($\phi_{\mathrm{PEFT}}$), $831{,}856$ temporal correction-head
parameters ($\phi_t$), and $236{,}828$ spatial correction-head
parameters ($\phi_s$).

\begin{table}[!t]
    \centering
    \caption{Trainable parameter counts for the main Aurora configuration.
    Counts include backbone adaptation and correction-head parameters.}
    \label{tab:aurora_trainable_params}
    \begin{tabular}{lr}
        \toprule
        Method & Trainable parameters \\
        \midrule
        Head only & 1,068,684 \\
        LoRA      & 3,919,500 \\
        DoRA      & 4,157,068 \\
        AdaLoRA   & 5,346,060 \\
        PiSSA     & 3,919,500 \\
        OPLoRA    & 3,919,500 \\
        PROLA & 3,919,500 \\
        Full FT   & 1,257,336,076 \\
        \bottomrule
    \end{tabular}
\end{table}

\textbf{Correction heads and numerical scaffolds.}
The correction heads \(C_s\) and \(C_t\) use separate surface
and upper-air branches with hidden widths 32 and 512, respectively.
Both use a single depthwise \(3\times3\) convolution followed by
pointwise SwiGLU~\cite{shazeer2020glu} projection, zero-initialized output projections,
and no dropout.
The spatial head \(C_s\) predicts a terminal residual from the
backbone forecast and endpoint conditioning fields.
The temporal head \(C_t\) combines endpoint-conditioned residual
basis maps using lead-dependent softmax gates.
It uses 4 basis maps for surface variables and 12 for upper-air
variables, with gate hidden widths 64 and 256 and a
16-dimensional lead-time embedding.

The fixed spatial operator \(U_s\) is bilinear interpolation.
It upsamples both surface fields and upper-air fields frame-wise from the \(32\times64\) low-resolution grid to the \(120\times240\) target grid.

The temporal scaffold in Eq.~\eqref{eq:temporal-stage} uses linear interpolation:
\begin{equation}
\tilde{\mathbf y}_{t+h}
= U_t\!\left(\tilde{\mathbf{x}}_t,
             \hat{\mathbf{y}}^{S}_{t+h_{\mathrm{end}}};h\right)
= \tilde{\mathbf{x}}_t
  + \frac{h}{h_{\mathrm{end}}}
    \left(\hat{\mathbf{y}}^{S}_{t+h_{\mathrm{end}}} - \tilde{\mathbf{x}}_t\right),
\qquad h\in\mathcal{H}_f.
\end{equation}
For the main Aurora benchmark, $h_{\mathrm{end}}=6\mathrm{h}$ and $\delta_f=1\mathrm{h}$.
This operator provides a non-learnable temporal scaffold, while \(C_t\) learns the residual sub-step deviations that the linear scaffold cannot capture.

\textbf{Baseline fairness.}
The PEFT methods in Table~\ref{tab:quant_main_3h6h} share the scaffold-and-correct pipeline, correction heads, and adapted attention projections.
The adapted modules are the QKV and output projections in
48 Aurora attention blocks
(encoder depths \((6,10,8)\) and decoder depths \((8,10,6)\)).
All non-adapted pretrained backbone parameters are frozen for PEFT methods.

\section{Additional Quantitative Results}
\label{app:full_results}

\begin{table*}[p]
\centering
\footnotesize
\setlength{\tabcolsep}{2pt}
\renewcommand{\arraystretch}{1.02}
\caption{\textbf{Full RMSE ($\downarrow$) results on WeatherBench~2~\cite{rasp2024weatherbench}.} We report latitude-weighted RMSE at all lead times from 1h to 6h as mean $\pm$ sample standard deviation over three seeds (0, 1, and 2). Best displayed mean results among PEFT methods are boldfaced, including ties after rounding.}
\label{tab:quant_main_full_leads}
\begin{tabular}{@{}l|c|c|cccccc@{}}
\toprule
Method & \shortstack{Trainable\\Params} & Variable & 1h & 2h & 3h & 4h & 5h & 6h \\
\midrule
\rowcolor{gray!6}
 &  & \ftcell{Z500} & \ftcell{\prolaMeanStd{48.22}{0.67}} & \ftcell{\prolaMeanStd{49.44}{1.39}} & \ftcell{\prolaMeanStd{60.13}{1.61}} & \ftcell{\prolaMeanStd{72.88}{1.82}} & \ftcell{\prolaMeanStd{85.63}{2.27}} & \ftcell{\prolaMeanStd{106.68}{2.63}} \\
\rowcolor{gray!6}
 &  & \ftcell{T850} & \ftcell{\prolaMeanStd{0.903}{0.011}} & \ftcell{\prolaMeanStd{0.834}{0.004}} & \ftcell{\prolaMeanStd{0.813}{0.002}} & \ftcell{\prolaMeanStd{0.874}{0.002}} & \ftcell{\prolaMeanStd{0.966}{0.003}} & \ftcell{\prolaMeanStd{1.114}{0.008}} \\
\rowcolor{gray!6}
 &  & \ftcell{T2M} & \ftcell{\prolaMeanStd{1.472}{0.003}} & \ftcell{\prolaMeanStd{1.225}{0.004}} & \ftcell{\prolaMeanStd{1.064}{0.009}} & \ftcell{\prolaMeanStd{0.955}{0.015}} & \ftcell{\prolaMeanStd{1.044}{0.019}} & \ftcell{\prolaMeanStd{1.323}{0.022}} \\
\rowcolor{gray!6}
 &  & \ftcell{U10} & \ftcell{\prolaMeanStd{1.239}{0.013}} & \ftcell{\prolaMeanStd{1.052}{0.008}} & \ftcell{\prolaMeanStd{0.902}{0.004}} & \ftcell{\prolaMeanStd{0.873}{0.003}} & \ftcell{\prolaMeanStd{0.941}{0.007}} & \ftcell{\prolaMeanStd{1.135}{0.012}} \\
\rowcolor{gray!6}
\multirow{-5}{*}{\ftcell{Full FT}} & \multirow{-5}{*}{\ftcell{1.26B}} & \ftcell{V10} & \ftcell{\prolaMeanStd{1.310}{0.024}} & \ftcell{\prolaMeanStd{1.094}{0.017}} & \ftcell{\prolaMeanStd{0.925}{0.009}} & \ftcell{\prolaMeanStd{0.894}{0.003}} & \ftcell{\prolaMeanStd{0.984}{0.003}} & \ftcell{\prolaMeanStd{1.209}{0.010}} \\
\midrule
\multirow{5}{*}{Head only} & \multirow{5}{*}{1.07M} & Z500 & \prolaMeanStd{68.16}{0.72} & \prolaMeanStd{78.45}{0.40} & \prolaMeanStd{94.93}{0.49} & \prolaMeanStd{112.98}{0.68} & \prolaMeanStd{128.19}{0.97} & \prolaMeanStd{144.03}{1.47} \\
 &  & T850 & \prolaMeanStd{1.016}{0.004} & \prolaMeanStd{1.013}{0.006} & \prolaMeanStd{1.049}{0.006} & \prolaMeanStd{1.135}{0.005} & \prolaMeanStd{1.218}{0.005} & \prolaMeanStd{1.315}{0.008} \\
 &  & T2M & \prolaMeanStd{1.704}{0.006} & \prolaMeanStd{1.931}{0.007} & \prolaMeanStd{2.083}{0.009} & \prolaMeanStd{2.297}{0.012} & \prolaMeanStd{2.553}{0.017} & \prolaMeanStd{2.656}{0.021} \\
 &  & U10 & \prolaMeanStd{1.540}{0.014} & \prolaMeanStd{1.578}{0.016} & \prolaMeanStd{1.648}{0.018} & \prolaMeanStd{1.752}{0.020} & \prolaMeanStd{1.847}{0.022} & \prolaMeanStd{1.947}{0.024} \\
 &  & V10 & \prolaMeanStd{1.423}{0.022} & \prolaMeanStd{1.476}{0.019} & \prolaMeanStd{1.563}{0.018} & \prolaMeanStd{1.685}{0.017} & \prolaMeanStd{1.799}{0.018} & \prolaMeanStd{1.921}{0.018} \\
\midrule
\multirow{5}{*}{LoRA} & \multirow{5}{*}{3.92M} & Z500 & \prolaMeanStd{55.67}{0.42} & \prolaMeanStd{51.19}{0.44} & \prolaMeanStd{54.72}{0.36} & \prolaMeanStd{60.22}{0.29} & \prolaMeanStd{64.76}{0.19} & \prolaMeanStd{79.17}{0.33} \\
 &  & T850 & \prolaMeanStd{0.943}{0.004} & \prolaMeanStd{0.913}{0.003} & \prolaMeanStd{0.898}{0.002} & \prolaMeanStd{0.923}{0.001} & \prolaMeanStd{0.944}{0.001} & \prolaMeanStd{0.994}{0.003} \\
 &  & T2M & \prolaMeanStd{1.399}{0.022} & \prolaMeanStd{1.266}{0.009} & \prolaMeanStd{1.199}{0.007} & \prolaMeanStd{1.140}{0.008} & \prolaMeanStd{1.182}{0.010} & \prolaMeanStd{1.280}{0.012} \\
 &  & U10 & \prolaMeanStd{1.294}{0.010} & \prolaMeanStd{1.248}{0.009} & \prolaMeanStd{1.219}{0.008} & \prolaMeanStd{1.236}{0.005} & \prolaMeanStd{1.255}{0.005} & \prolaMeanStd{1.314}{0.005} \\
 &  & V10 & \prolaMeanStd{1.221}{0.006} & \prolaMeanStd{1.191}{0.005} & \prolaMeanStd{1.181}{0.002} & \prolaMeanStd{1.214}{0.001} & \prolaMeanStd{1.242}{0.001} & \prolaMeanStd{1.306}{0.002} \\
\midrule
\multirow{5}{*}{DoRA} & \multirow{5}{*}{4.16M} & Z500 & \prolaMeanStd{55.45}{0.46} & \prolaMeanStd{50.83}{0.43} & \prolaMeanStd{54.32}{0.35} & \prolaMeanStd{59.73}{0.31} & \prolaMeanStd{64.10}{0.20} & \prolaMeanStd{78.40}{0.30} \\
 &  & T850 & \prolaMeanStd{0.941}{0.003} & \prolaMeanStd{0.909}{0.003} & \prolaMeanStd{0.893}{0.002} & \prolaMeanStd{0.918}{0.001} & \prolaMeanStd{0.939}{0.001} & \prolaMeanStd{0.989}{0.003} \\
 &  & T2M & \prolaMeanStd{1.392}{0.022} & \prolaMeanStd{1.258}{0.009} & \prolaMeanStd{1.191}{0.006} & \prolaMeanStd{1.131}{0.007} & \prolaMeanStd{1.174}{0.010} & \prolaMeanStd{1.272}{0.013} \\
 &  & U10 & \prolaMeanStd{1.286}{0.010} & \prolaMeanStd{1.239}{0.009} & \prolaMeanStd{1.209}{0.008} & \prolaMeanStd{1.226}{0.005} & \prolaMeanStd{1.245}{0.005} & \prolaMeanStd{1.306}{0.006} \\
 &  & V10 & \prolaMeanStd{1.215}{0.005} & \prolaMeanStd{1.184}{0.003} & \prolaMeanStd{1.173}{0.001} & \prolaMeanStd{1.205}{0.002} & \prolaMeanStd{1.233}{0.001} & \prolaMeanStd{1.298}{0.001} \\
\midrule
\multirow{5}{*}{AdaLoRA} & \multirow{5}{*}{5.35M} & Z500 & \prolaMeanStd{54.37}{0.47} & \prolaMeanStd{49.74}{0.21} & \prolaMeanStd{53.40}{0.27} & \prolaMeanStd{58.93}{0.32} & \prolaMeanStd{63.20}{0.11} & \prolaMeanStd{77.51}{0.94} \\
 &  & T850 & \prolaMeanStd{0.934}{0.011} & \prolaMeanStd{0.901}{0.008} & \prolaMeanStd{0.884}{0.006} & \prolaMeanStd{0.910}{0.004} & \prolaMeanStd{0.931}{0.005} & \prolaMeanStd{0.983}{0.006} \\
 &  & T2M & \prolaMeanStd{1.373}{0.010} & \prolaMeanStd{1.241}{0.002} & \prolaMeanStd{1.176}{0.005} & \prolaMeanStd{1.115}{0.005} & \prolaMeanStd{1.152}{0.002} & \prolaMeanStd{1.251}{0.002} \\
 &  & U10 & \prolaMeanStd{1.256}{0.009} & \prolaMeanStd{1.206}{0.006} & \prolaMeanStd{1.175}{0.005} & \prolaMeanStd{1.193}{0.005} & \prolaMeanStd{1.216}{0.005} & \prolaMeanStd{1.282}{0.004} \\
 &  & V10 & \prolaMeanStd{1.213}{0.009} & \prolaMeanStd{1.175}{0.006} & \prolaMeanStd{1.158}{0.002} & \prolaMeanStd{1.187}{0.001} & \prolaMeanStd{1.214}{0.001} & \prolaMeanStd{1.279}{0.002} \\
\midrule
\multirow{5}{*}{PiSSA} & \multirow{5}{*}{3.92M} & Z500 & \prolaMeanStd{55.65}{0.28} & \prolaMeanStd{51.28}{0.95} & \prolaMeanStd{54.55}{0.89} & \prolaMeanStd{59.60}{0.57} & \prolaMeanStd{63.79}{0.44} & \prolaMeanStd{78.43}{0.93} \\
 &  & T850 & \prolaMeanStd{0.942}{0.003} & \prolaMeanStd{0.910}{0.002} & \prolaMeanStd{0.893}{0.002} & \prolaMeanStd{0.918}{0.000} & \prolaMeanStd{0.938}{0.002} & \prolaMeanStd{0.989}{0.004} \\
 &  & T2M & \prolaMeanStd{1.382}{0.020} & \prolaMeanStd{1.261}{0.008} & \prolaMeanStd{1.198}{0.005} & \prolaMeanStd{1.131}{0.005} & \prolaMeanStd{1.158}{0.007} & \prolaMeanStd{1.251}{0.009} \\
 &  & U10 & \prolaMeanStd{1.291}{0.006} & \prolaMeanStd{1.245}{0.005} & \prolaMeanStd{1.215}{0.005} & \prolaMeanStd{1.228}{0.003} & \prolaMeanStd{1.243}{0.003} & \prolaMeanStd{1.298}{0.004} \\
 &  & V10 & \prolaMeanStd{1.218}{0.009} & \prolaMeanStd{1.187}{0.007} & \prolaMeanStd{1.175}{0.003} & \prolaMeanStd{1.204}{0.001} & \prolaMeanStd{1.228}{0.001} & \prolaMeanStd{1.289}{0.003} \\
\midrule
\multirow{5}{*}{OPLoRA} & \multirow{5}{*}{3.92M} & Z500 & \prolaMeanStd{55.77}{0.42} & \prolaMeanStd{51.37}{0.43} & \prolaMeanStd{54.94}{0.39} & \prolaMeanStd{60.48}{0.33} & \prolaMeanStd{65.11}{0.23} & \prolaMeanStd{79.60}{0.38} \\
 &  & T850 & \prolaMeanStd{0.944}{0.004} & \prolaMeanStd{0.914}{0.003} & \prolaMeanStd{0.899}{0.003} & \prolaMeanStd{0.925}{0.001} & \prolaMeanStd{0.946}{0.001} & \prolaMeanStd{0.996}{0.003} \\
 &  & T2M & \prolaMeanStd{1.400}{0.026} & \prolaMeanStd{1.270}{0.011} & \prolaMeanStd{1.204}{0.007} & \prolaMeanStd{1.145}{0.006} & \prolaMeanStd{1.184}{0.010} & \prolaMeanStd{1.281}{0.012} \\
 &  & U10 & \prolaMeanStd{1.297}{0.010} & \prolaMeanStd{1.251}{0.008} & \prolaMeanStd{1.223}{0.007} & \prolaMeanStd{1.240}{0.005} & \prolaMeanStd{1.259}{0.004} & \prolaMeanStd{1.319}{0.005} \\
 &  & V10 & \prolaMeanStd{1.224}{0.008} & \prolaMeanStd{1.194}{0.006} & \prolaMeanStd{1.185}{0.003} & \prolaMeanStd{1.217}{0.001} & \prolaMeanStd{1.246}{0.001} & \prolaMeanStd{1.311}{0.002} \\
\midrule
\rowcolor{cyan!10}
 &  & Z500 & \prolaBestMeanStd{52.98}{0.22} & \prolaBestMeanStd{47.49}{0.49} & \prolaBestMeanStd{50.58}{0.42} & \prolaBestMeanStd{55.44}{0.22} & \prolaBestMeanStd{58.62}{0.15} & \prolaBestMeanStd{72.35}{0.64} \\
\rowcolor{cyan!10}
 &  & T850 & \prolaBestMeanStd{0.927}{0.006} & \prolaBestMeanStd{0.887}{0.004} & \prolaBestMeanStd{0.864}{0.003} & \prolaBestMeanStd{0.885}{0.002} & \prolaBestMeanStd{0.904}{0.002} & \prolaBestMeanStd{0.956}{0.002} \\
\rowcolor{cyan!10}
 &  & T2M & \prolaBestMeanStd{1.372}{0.022} & \prolaBestMeanStd{1.205}{0.004} & \prolaBestMeanStd{1.129}{0.002} & \prolaBestMeanStd{1.062}{0.002} & \prolaBestMeanStd{1.109}{0.006} & \prolaBestMeanStd{1.212}{0.006} \\
\rowcolor{cyan!10}
 &  & U10 & \prolaBestMeanStd{1.236}{0.009} & \prolaBestMeanStd{1.181}{0.006} & \prolaBestMeanStd{1.143}{0.003} & \prolaBestMeanStd{1.156}{0.002} & \prolaBestMeanStd{1.175}{0.002} & \prolaBestMeanStd{1.240}{0.005} \\
\rowcolor{cyan!10}
\multirow{-5}{*}{PROLA (Ours)} & \multirow{-5}{*}{3.92M} & V10 & \prolaBestMeanStd{1.178}{0.004} & \prolaBestMeanStd{1.137}{0.002} & \prolaBestMeanStd{1.120}{0.001} & \prolaBestMeanStd{1.148}{0.001} & \prolaBestMeanStd{1.171}{0.001} & \prolaBestMeanStd{1.234}{0.002} \\
\bottomrule
\end{tabular}
\end{table*}

\begin{table*}[p]
\centering
\footnotesize
\setlength{\tabcolsep}{2pt}
\renewcommand{\arraystretch}{1.02}
\caption{\textbf{Full ACC ($\uparrow$) results on WeatherBench~2~\cite{rasp2024weatherbench}.} We report latitude-weighted ACC at all lead times from 1h to 6h as mean $\pm$ sample standard deviation over three seeds (0, 1, and 2). Best displayed mean results among PEFT methods are boldfaced, including ties after rounding.}
\label{tab:quant_main_full_leads_acc}
\begin{tabular}{@{}l|c|c|cccccc@{}}
\toprule
Method & \shortstack{Trainable\\Params} & Variable & 1h & 2h & 3h & 4h & 5h & 6h \\
\midrule
\rowcolor{gray!6}
 &  & \ftcell{Z500} & \ftcell{\prolaMeanStd{0.999}{0.000}} & \ftcell{\prolaMeanStd{0.999}{0.000}} & \ftcell{\prolaMeanStd{0.998}{0.000}} & \ftcell{\prolaMeanStd{0.997}{0.000}} & \ftcell{\prolaMeanStd{0.996}{0.000}} & \ftcell{\prolaMeanStd{0.995}{0.000}} \\
\rowcolor{gray!6}
 &  & \ftcell{T850} & \ftcell{\prolaMeanStd{0.984}{0.000}} & \ftcell{\prolaMeanStd{0.986}{0.000}} & \ftcell{\prolaMeanStd{0.987}{0.000}} & \ftcell{\prolaMeanStd{0.985}{0.000}} & \ftcell{\prolaMeanStd{0.981}{0.000}} & \ftcell{\prolaMeanStd{0.976}{0.000}} \\
\rowcolor{gray!6}
 &  & \ftcell{T2M} & \ftcell{\prolaMeanStd{0.958}{0.001}} & \ftcell{\prolaMeanStd{0.972}{0.000}} & \ftcell{\prolaMeanStd{0.979}{0.000}} & \ftcell{\prolaMeanStd{0.983}{0.000}} & \ftcell{\prolaMeanStd{0.980}{0.001}} & \ftcell{\prolaMeanStd{0.970}{0.001}} \\
\rowcolor{gray!6}
 &  & \ftcell{U10} & \ftcell{\prolaMeanStd{0.952}{0.001}} & \ftcell{\prolaMeanStd{0.966}{0.000}} & \ftcell{\prolaMeanStd{0.975}{0.000}} & \ftcell{\prolaMeanStd{0.976}{0.000}} & \ftcell{\prolaMeanStd{0.973}{0.000}} & \ftcell{\prolaMeanStd{0.961}{0.001}} \\
\rowcolor{gray!6}
\multirow{-5}{*}{\ftcell{Full FT}} & \multirow{-5}{*}{\ftcell{1.26B}} & \ftcell{V10} & \ftcell{\prolaMeanStd{0.947}{0.002}} & \ftcell{\prolaMeanStd{0.964}{0.001}} & \ftcell{\prolaMeanStd{0.974}{0.001}} & \ftcell{\prolaMeanStd{0.976}{0.000}} & \ftcell{\prolaMeanStd{0.971}{0.000}} & \ftcell{\prolaMeanStd{0.956}{0.001}} \\
\midrule
\multirow{5}{*}{Head only} & \multirow{5}{*}{1.07M} & Z500 & \prolaMeanStd{0.998}{0.000} & \prolaMeanStd{0.997}{0.000} & \prolaMeanStd{0.996}{0.000} & \prolaMeanStd{0.994}{0.000} & \prolaMeanStd{0.992}{0.000} & \prolaMeanStd{0.990}{0.000} \\
 &  & T850 & \prolaMeanStd{0.979}{0.000} & \prolaMeanStd{0.979}{0.000} & \prolaMeanStd{0.978}{0.000} & \prolaMeanStd{0.974}{0.000} & \prolaMeanStd{0.970}{0.000} & \prolaMeanStd{0.965}{0.001} \\
 &  & T2M & \prolaMeanStd{0.944}{0.001} & \prolaMeanStd{0.927}{0.001} & \prolaMeanStd{0.915}{0.001} & \prolaMeanStd{0.894}{0.002} & \prolaMeanStd{0.868}{0.001} & \prolaMeanStd{0.856}{0.001} \\
 &  & U10 & \prolaMeanStd{0.925}{0.001} & \prolaMeanStd{0.922}{0.001} & \prolaMeanStd{0.914}{0.001} & \prolaMeanStd{0.902}{0.002} & \prolaMeanStd{0.890}{0.002} & \prolaMeanStd{0.877}{0.002} \\
 &  & V10 & \prolaMeanStd{0.938}{0.002} & \prolaMeanStd{0.933}{0.002} & \prolaMeanStd{0.924}{0.002} & \prolaMeanStd{0.910}{0.002} & \prolaMeanStd{0.898}{0.003} & \prolaMeanStd{0.883}{0.003} \\
\midrule
\multirow{5}{*}{LoRA} & \multirow{5}{*}{3.92M} & Z500 & \prolaMeanStd{0.998}{0.000} & \prolaBestMeanStd{0.999}{0.000} & \prolaBestMeanStd{0.999}{0.000} & \prolaMeanStd{0.998}{0.000} & \prolaBestMeanStd{0.998}{0.000} & \prolaBestMeanStd{0.997}{0.000} \\
 &  & T850 & \prolaMeanStd{0.982}{0.000} & \prolaMeanStd{0.983}{0.000} & \prolaMeanStd{0.984}{0.000} & \prolaMeanStd{0.983}{0.000} & \prolaMeanStd{0.982}{0.000} & \prolaMeanStd{0.980}{0.000} \\
 &  & T2M & \prolaMeanStd{0.963}{0.001} & \prolaMeanStd{0.970}{0.000} & \prolaMeanStd{0.973}{0.000} & \prolaMeanStd{0.976}{0.000} & \prolaMeanStd{0.974}{0.000} & \prolaMeanStd{0.969}{0.001} \\
 &  & U10 & \prolaMeanStd{0.948}{0.001} & \prolaMeanStd{0.952}{0.000} & \prolaMeanStd{0.954}{0.000} & \prolaMeanStd{0.952}{0.000} & \prolaMeanStd{0.951}{0.000} & \prolaMeanStd{0.946}{0.000} \\
 &  & V10 & \prolaMeanStd{0.954}{0.000} & \prolaMeanStd{0.957}{0.000} & \prolaMeanStd{0.957}{0.000} & \prolaMeanStd{0.955}{0.000} & \prolaMeanStd{0.953}{0.000} & \prolaMeanStd{0.948}{0.000} \\
\midrule
\multirow{5}{*}{DoRA} & \multirow{5}{*}{4.16M} & Z500 & \prolaBestMeanStd{0.999}{0.000} & \prolaBestMeanStd{0.999}{0.000} & \prolaBestMeanStd{0.999}{0.000} & \prolaMeanStd{0.998}{0.000} & \prolaBestMeanStd{0.998}{0.000} & \prolaBestMeanStd{0.997}{0.000} \\
 &  & T850 & \prolaMeanStd{0.982}{0.000} & \prolaMeanStd{0.983}{0.000} & \prolaMeanStd{0.984}{0.000} & \prolaMeanStd{0.983}{0.000} & \prolaMeanStd{0.982}{0.000} & \prolaMeanStd{0.980}{0.000} \\
 &  & T2M & \prolaMeanStd{0.963}{0.001} & \prolaMeanStd{0.970}{0.000} & \prolaMeanStd{0.973}{0.000} & \prolaMeanStd{0.976}{0.000} & \prolaMeanStd{0.974}{0.000} & \prolaMeanStd{0.970}{0.001} \\
 &  & U10 & \prolaMeanStd{0.948}{0.001} & \prolaMeanStd{0.952}{0.000} & \prolaMeanStd{0.955}{0.000} & \prolaMeanStd{0.953}{0.000} & \prolaMeanStd{0.952}{0.000} & \prolaMeanStd{0.947}{0.000} \\
 &  & V10 & \prolaMeanStd{0.955}{0.000} & \prolaMeanStd{0.957}{0.000} & \prolaMeanStd{0.958}{0.000} & \prolaMeanStd{0.955}{0.000} & \prolaMeanStd{0.953}{0.000} & \prolaMeanStd{0.948}{0.000} \\
\midrule
\multirow{5}{*}{AdaLoRA} & \multirow{5}{*}{5.35M} & Z500 & \prolaBestMeanStd{0.999}{0.000} & \prolaBestMeanStd{0.999}{0.000} & \prolaBestMeanStd{0.999}{0.000} & \prolaMeanStd{0.998}{0.000} & \prolaBestMeanStd{0.998}{0.000} & \prolaBestMeanStd{0.997}{0.000} \\
 &  & T850 & \prolaMeanStd{0.982}{0.000} & \prolaBestMeanStd{0.984}{0.000} & \prolaMeanStd{0.984}{0.000} & \prolaMeanStd{0.983}{0.000} & \prolaMeanStd{0.983}{0.000} & \prolaMeanStd{0.981}{0.000} \\
 &  & T2M & \prolaBestMeanStd{0.964}{0.001} & \prolaMeanStd{0.971}{0.000} & \prolaMeanStd{0.974}{0.000} & \prolaMeanStd{0.977}{0.000} & \prolaMeanStd{0.975}{0.000} & \prolaMeanStd{0.971}{0.001} \\
 &  & U10 & \prolaMeanStd{0.951}{0.000} & \prolaMeanStd{0.955}{0.000} & \prolaMeanStd{0.957}{0.000} & \prolaMeanStd{0.956}{0.000} & \prolaMeanStd{0.954}{0.000} & \prolaMeanStd{0.949}{0.000} \\
 &  & V10 & \prolaMeanStd{0.955}{0.000} & \prolaMeanStd{0.958}{0.000} & \prolaMeanStd{0.959}{0.000} & \prolaMeanStd{0.957}{0.000} & \prolaMeanStd{0.955}{0.000} & \prolaMeanStd{0.950}{0.000} \\
\midrule
\multirow{5}{*}{PiSSA} & \multirow{5}{*}{3.92M} & Z500 & \prolaBestMeanStd{0.999}{0.000} & \prolaBestMeanStd{0.999}{0.000} & \prolaBestMeanStd{0.999}{0.000} & \prolaMeanStd{0.998}{0.000} & \prolaBestMeanStd{0.998}{0.000} & \prolaBestMeanStd{0.997}{0.000} \\
 &  & T850 & \prolaMeanStd{0.982}{0.000} & \prolaMeanStd{0.983}{0.000} & \prolaMeanStd{0.984}{0.000} & \prolaMeanStd{0.983}{0.000} & \prolaMeanStd{0.982}{0.000} & \prolaMeanStd{0.980}{0.000} \\
 &  & T2M & \prolaMeanStd{0.963}{0.001} & \prolaMeanStd{0.970}{0.001} & \prolaMeanStd{0.973}{0.000} & \prolaMeanStd{0.976}{0.000} & \prolaMeanStd{0.975}{0.000} & \prolaMeanStd{0.971}{0.001} \\
 &  & U10 & \prolaMeanStd{0.948}{0.000} & \prolaMeanStd{0.952}{0.000} & \prolaMeanStd{0.954}{0.000} & \prolaMeanStd{0.953}{0.000} & \prolaMeanStd{0.952}{0.000} & \prolaMeanStd{0.948}{0.000} \\
 &  & V10 & \prolaMeanStd{0.955}{0.000} & \prolaMeanStd{0.957}{0.000} & \prolaMeanStd{0.958}{0.000} & \prolaMeanStd{0.955}{0.000} & \prolaMeanStd{0.954}{0.000} & \prolaMeanStd{0.949}{0.000} \\
\midrule
\multirow{5}{*}{OPLoRA} & \multirow{5}{*}{3.92M} & Z500 & \prolaMeanStd{0.998}{0.000} & \prolaBestMeanStd{0.999}{0.000} & \prolaBestMeanStd{0.999}{0.000} & \prolaMeanStd{0.998}{0.000} & \prolaBestMeanStd{0.998}{0.000} & \prolaBestMeanStd{0.997}{0.000} \\
 &  & T850 & \prolaMeanStd{0.982}{0.000} & \prolaMeanStd{0.983}{0.000} & \prolaMeanStd{0.984}{0.000} & \prolaMeanStd{0.983}{0.000} & \prolaMeanStd{0.982}{0.000} & \prolaMeanStd{0.980}{0.000} \\
 &  & T2M & \prolaMeanStd{0.962}{0.001} & \prolaMeanStd{0.970}{0.000} & \prolaMeanStd{0.973}{0.000} & \prolaMeanStd{0.975}{0.000} & \prolaMeanStd{0.974}{0.000} & \prolaMeanStd{0.969}{0.000} \\
 &  & U10 & \prolaMeanStd{0.948}{0.001} & \prolaMeanStd{0.951}{0.000} & \prolaMeanStd{0.954}{0.000} & \prolaMeanStd{0.952}{0.000} & \prolaMeanStd{0.951}{0.000} & \prolaMeanStd{0.946}{0.000} \\
 &  & V10 & \prolaMeanStd{0.954}{0.000} & \prolaMeanStd{0.957}{0.000} & \prolaMeanStd{0.957}{0.000} & \prolaMeanStd{0.954}{0.000} & \prolaMeanStd{0.952}{0.000} & \prolaMeanStd{0.947}{0.000} \\
\midrule
\rowcolor{cyan!10}
 &  & Z500 & \prolaBestMeanStd{0.999}{0.000} & \prolaBestMeanStd{0.999}{0.000} & \prolaBestMeanStd{0.999}{0.000} & \prolaBestMeanStd{0.999}{0.000} & \prolaBestMeanStd{0.998}{0.000} & \prolaBestMeanStd{0.997}{0.000} \\
\rowcolor{cyan!10}
 &  & T850 & \prolaBestMeanStd{0.983}{0.000} & \prolaBestMeanStd{0.984}{0.000} & \prolaBestMeanStd{0.985}{0.000} & \prolaBestMeanStd{0.984}{0.000} & \prolaBestMeanStd{0.984}{0.000} & \prolaBestMeanStd{0.982}{0.000} \\
\rowcolor{cyan!10}
 &  & T2M & \prolaBestMeanStd{0.964}{0.001} & \prolaBestMeanStd{0.973}{0.001} & \prolaBestMeanStd{0.976}{0.000} & \prolaBestMeanStd{0.979}{0.000} & \prolaBestMeanStd{0.977}{0.001} & \prolaBestMeanStd{0.973}{0.001} \\
\rowcolor{cyan!10}
 &  & U10 & \prolaBestMeanStd{0.953}{0.000} & \prolaBestMeanStd{0.957}{0.000} & \prolaBestMeanStd{0.960}{0.000} & \prolaBestMeanStd{0.958}{0.000} & \prolaBestMeanStd{0.957}{0.000} & \prolaBestMeanStd{0.952}{0.000} \\
\rowcolor{cyan!10}
\multirow{-5}{*}{PROLA (Ours)} & \multirow{-5}{*}{3.92M} & V10 & \prolaBestMeanStd{0.958}{0.000} & \prolaBestMeanStd{0.961}{0.000} & \prolaBestMeanStd{0.962}{0.000} & \prolaBestMeanStd{0.959}{0.000} & \prolaBestMeanStd{0.958}{0.000} & \prolaBestMeanStd{0.953}{0.000} \\
\bottomrule
\end{tabular}
\end{table*}

\textbf{Full main results.}
Tables~\ref{tab:quant_main_full_leads} and~\ref{tab:quant_main_full_leads_acc} report complete per-lead RMSE and ACC results over the target lead times \(1\mathrm{h}\)--\(6\mathrm{h}\), respectively.
Each entry reports the mean and sample standard deviation over seeds 0, 1, and 2, following the latitude-weighted evaluation protocol in Appendix~\ref{app:metrics}.
These results extend the representative \(3\mathrm{h}\) and \(6\mathrm{h}\) comparisons to the full forecast trajectory.

\textbf{Regional weather prediction.}
\phantomsection\label{app:regional_generalization}
\begin{table*}[t]
\centering

\small
\setlength{\tabcolsep}{3pt}
\renewcommand{\arraystretch}{1.18}
\setlength{\arrayrulewidth}{0.3pt}

\caption{
\textbf{Generalization to regional weather prediction.}
ERA5-to-CERRA results report latitude-weighted RMSE ($\downarrow$) for five variables at 3h and 6h.
Regional precipitation results report SEEPS ($\downarrow$), RMSE in cm ($\downarrow$), and ACC ($\uparrow$) at 12h, 24h, and 36h.
All methods are trained for 10 epochs and evaluated on the 2018 test split.
Best displayed results within each block are boldfaced.
}
\label{tab:regional_generalization_full}

\resizebox{0.95\textwidth}{!}{%
\begin{tabularx}{\textwidth}{X|rrrrr|rrrrr}
\toprule
\rowcolor{prolaSection}
\multicolumn{11}{l}{\textit{ERA5 $\rightarrow$ CERRA: regional spatiotemporal downscaling}} \\
& \multicolumn{5}{c|}{\textit{3h lead}} & \multicolumn{5}{c}{\textit{6h lead}} \\
Method & Z500 & T850 & T2M & U10 & V10 & Z500 & T850 & T2M & U10 & V10 \\
\midrule
LoRA & 72.14 & 1.076 & 2.150 & 1.370 & 1.327 & 86.33 & 1.092 & 2.147 & 1.373 & 1.340 \\
PiSSA & 71.80 & 1.072 & 2.126 & 1.354 & 1.311 & 85.72 & 1.084 & 2.114 & 1.352 & 1.323 \\
OPLoRA & 72.56 & 1.081 & 2.132 & 1.371 & 1.332 & 86.13 & 1.099 & 2.146 & 1.375 & 1.348 \\
\rowcolor{prolaSky}
PROLA & \textbf{68.44} & \textbf{1.028} & \textbf{2.021} & \textbf{1.316} & \textbf{1.274} & \textbf{81.41} & \textbf{1.036} & \textbf{2.019} & \textbf{1.319} & \textbf{1.283} \\
\bottomrule
\end{tabularx}%
}

\par\vspace{0.7em}

\resizebox{0.95\textwidth}{!}{%
\begin{tabularx}{\textwidth}{X|rrr|rrr|rrr}
\toprule
\rowcolor{prolaSection}
\multicolumn{10}{l}{\textit{Regional precipitation forecasting: future 6h accumulated precipitation}} \\
& \multicolumn{3}{c|}{\textit{12h lead}} & \multicolumn{3}{c|}{\textit{24h lead}} & \multicolumn{3}{c}{\textit{36h lead}} \\
Method & SEEPS $\downarrow$ & RMSE $\downarrow$ & ACC $\uparrow$ & SEEPS $\downarrow$ & RMSE $\downarrow$ & ACC $\uparrow$ & SEEPS $\downarrow$ & RMSE $\downarrow$ & ACC $\uparrow$ \\
\midrule
LoRA & 0.608 & 0.212 & 0.660 & 0.706 & 0.228 & 0.586 & 0.784 & 0.238 & 0.537 \\
PiSSA & 0.612 & 0.213 & 0.656 & 0.711 & 0.229 & 0.583 & 0.788 & 0.238 & 0.537 \\
OPLoRA & 0.605 & 0.212 & 0.661 & 0.704 & 0.228 & 0.587 & 0.780 & 0.238 & 0.538 \\
\rowcolor{prolaSky}
PROLA & \textbf{0.590} & \textbf{0.208} & \textbf{0.676} & \textbf{0.692} & \textbf{0.225} & \textbf{0.600} & \textbf{0.751} & \textbf{0.233} & \textbf{0.562} \\
\bottomrule
\end{tabularx}%
}

\end{table*}

Table~\ref{tab:regional_generalization_full} provides the full results for the regional tasks in Section~\ref{sec:generalization}.
For ERA5-to-CERRA downscaling, PROLA achieves the lowest RMSE across all five reported variables at both 3h and 6h.
For regional precipitation forecasting, it achieves the lowest SEEPS and RMSE and the highest ACC at 12h, 24h, and 36h.
These comparisons use seed 0 and the 10-epoch training settings described in Appendices~\ref{app:experimental_setup} and~\ref{app:implementation}.

\textbf{Transfer of the adaptation design.}\space\space
We adapt Qwen3.5-9B~\cite{qwen2026qwen35} to BIRD~\cite{li2023bird} to test whether reserving capacity for orthogonal corrections alongside the principal branch also benefits schema-conditioned query generation.
Table~\ref{tab:bird_generalization} shows that at rank 16, PROLA achieves 52.87\% execution accuracy, exceeding PiSSA and LoRA by 0.91 and 3.33 percentage points, respectively.
See Appendices~\ref{app:text_to_sql},~\ref{app:metrics}, and~\ref{app:implementation} for the data split, evaluation metric, and training settings.

\begin{table}[!t]
\centering
\caption{\textbf{Text-to-SQL transfer.} Execution accuracy (EX) on BIRD for Qwen3.5-9B.}
\label{tab:bird_generalization}
\small
\setlength{\tabcolsep}{7pt}
\begin{tabular}{lrrrrr}
\toprule
 & Pretrained & LoRA & PiSSA & OPLoRA & \cellcolor{prolaSky}PROLA \\
\midrule
EX (\%) $\uparrow$ & 45.50 & 49.54 & 51.96 & 49.41 & \cellcolor{prolaSky}\textbf{52.87} \\
\bottomrule
\end{tabular}\par
\end{table}

\begin{table*}[t]
\centering
\small
\setlength{\tabcolsep}{5pt}
\renewcommand{\arraystretch}{1.15}
\caption{\textbf{Backbone transfer to ArchesWeather.} Latitude-weighted RMSE ($\downarrow$) on the 2018 test split after one training epoch and total trainable adapter rank 8. PROLA uses \(r_p=r_o=4\). Best displayed results within each lead time are boldfaced.}
\label{tab:app_arches_full_leads}
\begin{tabular}{l|rrrrr|rrrrr}
\toprule
\rowcolor{gray!10}
\cellcolor{white} & \multicolumn{5}{c|}{\textit{6h lead}} & \multicolumn{5}{c}{\textit{12h lead}} \\
Method & Z500 & T850 & T2M & U10 & V10 & Z500 & T850 & T2M & U10 & V10 \\
\midrule
LoRA & 138.22 & 1.241 & 1.967 & 2.055 & 2.079 & 123.23 & 1.291 & 1.757 & \textbf{2.368} & 3.175 \\
PiSSA & 134.41 & 1.205 & 1.904 & 2.556 & 2.112 & 121.61 & 1.247 & 2.062 & 2.457 & \textbf{2.272} \\
OPLoRA & 139.20 & 1.237 & 1.839 & 2.093 & 2.097 & 124.38 & 1.296 & 1.760 & 2.519 & 3.102 \\
\rowcolor{cyan!10}
PROLA & \textbf{133.03} & \textbf{1.192} & \textbf{1.531} & \textbf{2.031} & \textbf{2.077} & \textbf{119.97} & \textbf{1.246} & \textbf{1.614} & 2.749 & 2.838 \\

\midrule
\rowcolor{gray!10}
\cellcolor{white} & \multicolumn{5}{c|}{\textit{18h lead}} & \multicolumn{5}{c}{\textit{24h lead}} \\
Method & Z500 & T850 & T2M & U10 & V10 & Z500 & T850 & T2M & U10 & V10 \\
\midrule
LoRA & 144.02 & 1.261 & 2.058 & \textbf{2.344} & 2.967 & 148.49 & 1.347 & 1.755 & \textbf{2.729} & 3.768 \\
PiSSA & 137.46 & 1.237 & 2.036 & 2.679 & \textbf{2.141} & 148.09 & 1.334 & 1.910 & 3.296 & \textbf{2.415} \\
OPLoRA & 145.96 & 1.271 & 2.052 & 2.400 & 2.979 & 150.05 & 1.357 & 1.733 & 2.875 & 3.708 \\
\rowcolor{cyan!10}
PROLA & \textbf{132.14} & \textbf{1.228} & \textbf{1.862} & 2.610 & 2.928 & \textbf{142.66} & \textbf{1.316} & \textbf{1.604} & 3.191 & 3.459 \\

\bottomrule
\end{tabular}
\end{table*}

\textbf{Transfer to ArchesWeather.}
Table~\ref{tab:app_arches_full_leads} evaluates the same adaptation methods with the ArchesWeather-M backbone at 6h, 12h, 18h, and 24h.
PROLA uses \(r_p=r_o=4\), and all methods use the same attention projections and correction heads.
PROLA achieves the lowest Z500 and T2M RMSE at all four lead times, extending the comparison beyond the Aurora backbone.
The one-epoch training configuration is given in Appendix~\ref{app:implementation}.

\begin{table}[tbp]
\centering
\small
\setlength{\tabcolsep}{10pt}
\renewcommand{\arraystretch}{1.20}
\caption{\textbf{Additional resolution gap.} Latitude-weighted RMSE ($\downarrow$) on the 2018 test split for $5.625^\circ$-to-$2.8125^\circ$ downscaling. Best displayed values are boldfaced.}
\label{tab:additional_resolution}
\begin{tabular}{l|cc|cc}
\toprule
\rowcolor{gray!10}
\cellcolor{white} & \multicolumn{2}{c|}{\textit{3h lead}} & \multicolumn{2}{c}{\textit{6h lead}} \\
Method & Z500 & T2M & Z500 & T2M \\
\midrule
LoRA & 52.27 & 1.148 & 76.14 & 1.141 \\
PiSSA & 54.62 & 1.171 & 82.54 & 1.159 \\
OPLoRA & 52.08 & 1.145 & 76.25 & 1.134 \\
\rowcolor{cyan!10}
PROLA & \textbf{48.01} & \textbf{1.054} & \textbf{69.58} & \textbf{0.998} \\
\bottomrule
\end{tabular}
\end{table}

\textbf{Additional resolution gap.}
Table~\ref{tab:additional_resolution} compares the four PEFT methods for \(5.625^\circ\)-to-\(2.8125^\circ\) downscaling with targets at 3h and 6h.
PROLA achieves the lowest Z500 and T2M RMSE at both lead times.

\begin{table}[!tbp]
\centering
\small
\setlength{\tabcolsep}{6pt}
\renewcommand{\arraystretch}{1.12}
\caption{\textbf{Seasonal robustness.} Latitude-weighted Z500 RMSE ($\downarrow$) at 6h on the 2018 test split. Best displayed values are boldfaced.}
\label{tab:seasonal_robustness}
\begin{tabular}{lrrrr}
\toprule
Method & DJF & MAM & JJA & SON \\
\midrule
LoRA & 82.78 & 79.28 & 75.29 & 79.70 \\
PiSSA & 81.70 & 78.54 & 74.37 & 79.25 \\
OPLoRA & 83.28 & 79.88 & 75.70 & 79.99 \\
\rowcolor{cyan!10}
PROLA & \textbf{75.99} & \textbf{73.22} & \textbf{69.73} & \textbf{73.65} \\
\bottomrule
\end{tabular}
\end{table}

\begin{table}[tbp]
\centering
\small
\setlength{\tabcolsep}{6pt}
\renewcommand{\arraystretch}{1.12}
\caption{\textbf{Spatial-tail performance.} Z500 at 6h on the 2018 test split. Upper and lower tails are defined within each target field as in Appendix~\ref{app:tail_metrics}. Best displayed values are boldfaced.}
\label{tab:tail_robustness}
\begin{tabular}{lrrrr}
\toprule
Method & Upper RMSE $\downarrow$ & Upper BSS $\uparrow$ & Lower RMSE $\downarrow$ & Lower BSS $\uparrow$ \\
\midrule
LoRA & 56.87 & 0.514 & 129.19 & 0.780 \\
PiSSA & 55.50 & 0.513 & 126.28 & 0.781 \\
OPLoRA & 56.63 & 0.512 & 127.55 & 0.776 \\
\rowcolor{cyan!10}
PROLA & \textbf{50.80} & \textbf{0.545} & \textbf{116.36} & \textbf{0.795} \\
\bottomrule
\end{tabular}
\end{table}

\textbf{Seasonal and spatial-tail robustness.}
Tables~\ref{tab:seasonal_robustness} and~\ref{tab:tail_robustness} evaluate Z500 forecasts at 6h across seasons and within the upper and lower spatial tails of the target fields.
PROLA has the lowest RMSE in all four seasons and both tails, and the highest BSS for both tail events.
The evaluation protocol is defined in Appendix~\ref{app:tail_metrics}.

\begin{table*}[t]
\centering
\small
\setlength{\tabcolsep}{5pt}
\renewcommand{\arraystretch}{1.15}
\caption{\textbf{Spherical spectral distance.} Mean absolute log-ratio of prediction to target power ($\downarrow$), evaluated over degrees 1--59 using all 1,458 test cases in 2018. Best displayed values are boldfaced, including rounded ties.}
\label{tab:app_spectral_distance}
\begin{tabular}{l|rrrrr|rrrrr}
\toprule
& \multicolumn{5}{c|}{\textit{3h lead}} & \multicolumn{5}{c}{\textit{6h lead}} \\
Method & Z500 & T850 & T2M & U10 & V10 & Z500 & T850 & T2M & U10 & V10 \\
\midrule
LoRA & 0.301 & 0.627 & 0.272 & 0.396 & 0.372 & 0.077 & 0.349 & 0.081 & 0.286 & 0.260 \\
PiSSA & 0.304 & 0.619 & 0.277 & 0.392 & 0.369 & 0.082 & 0.331 & 0.079 & 0.291 & 0.263 \\
OPLoRA & 0.302 & 0.628 & 0.277 & 0.402 & 0.378 & 0.077 & 0.350 & 0.087 & 0.290 & 0.264 \\
\rowcolor{cyan!10}
PROLA & \textbf{0.294} & \textbf{0.589} & \textbf{0.229} & \textbf{0.329} & \textbf{0.312} & \textbf{0.072} & \textbf{0.293} & \textbf{0.062} & \textbf{0.248} & \textbf{0.204} \\
\bottomrule
\end{tabular}
\end{table*}

\begin{figure*}[t]
\centering
\includegraphics[width=0.95\textwidth]{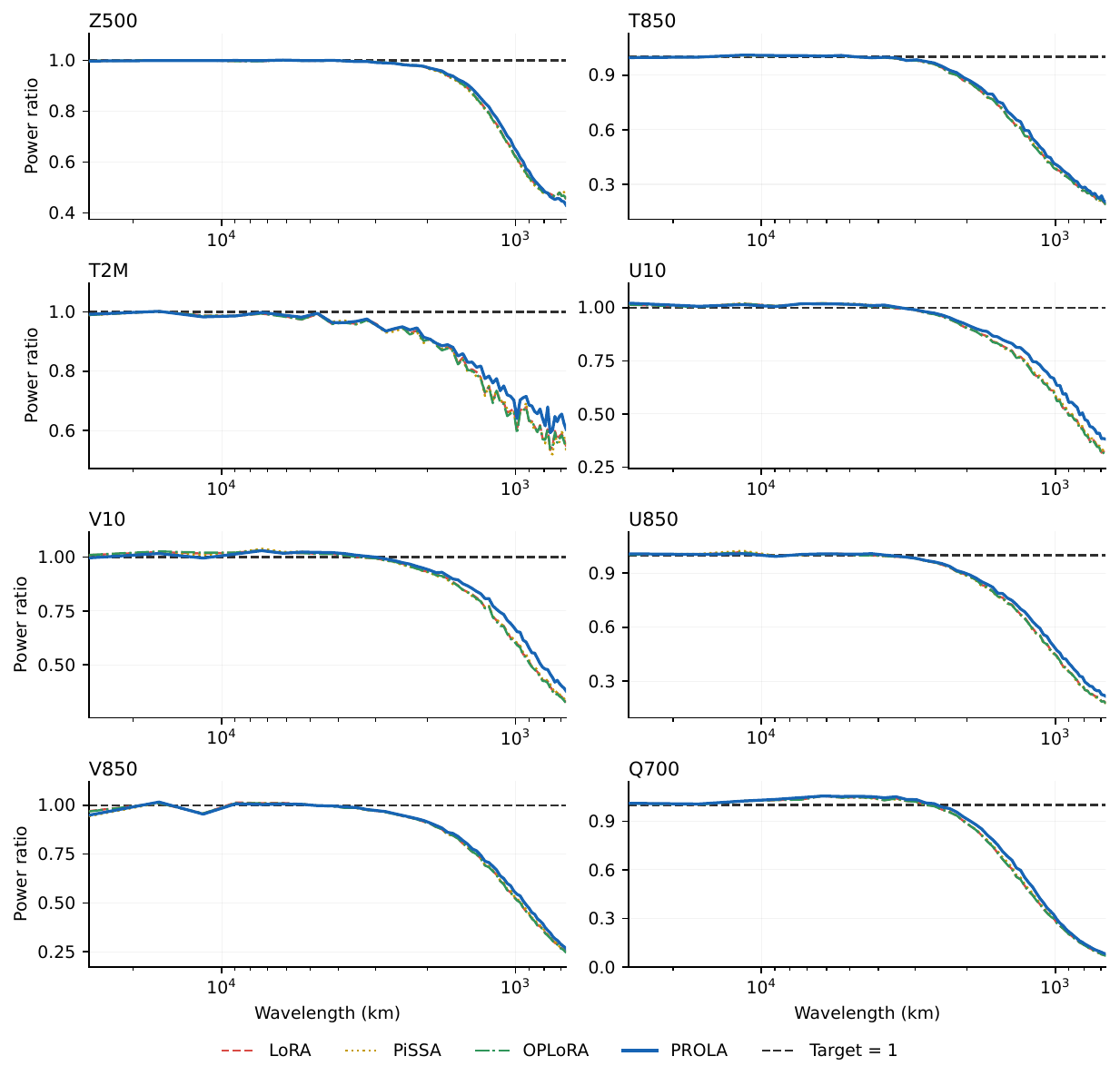}
\caption{\textbf{Spherical power ratios at 3h.} Ratios of mean prediction power to mean target power for eight variables, evaluated on all 1,458 test cases in 2018. The dashed horizontal line marks the target ratio of one.}
\label{fig:app_spectrum_3h}
\end{figure*}

\begin{figure*}[t]
\centering
\includegraphics[width=0.95\textwidth]{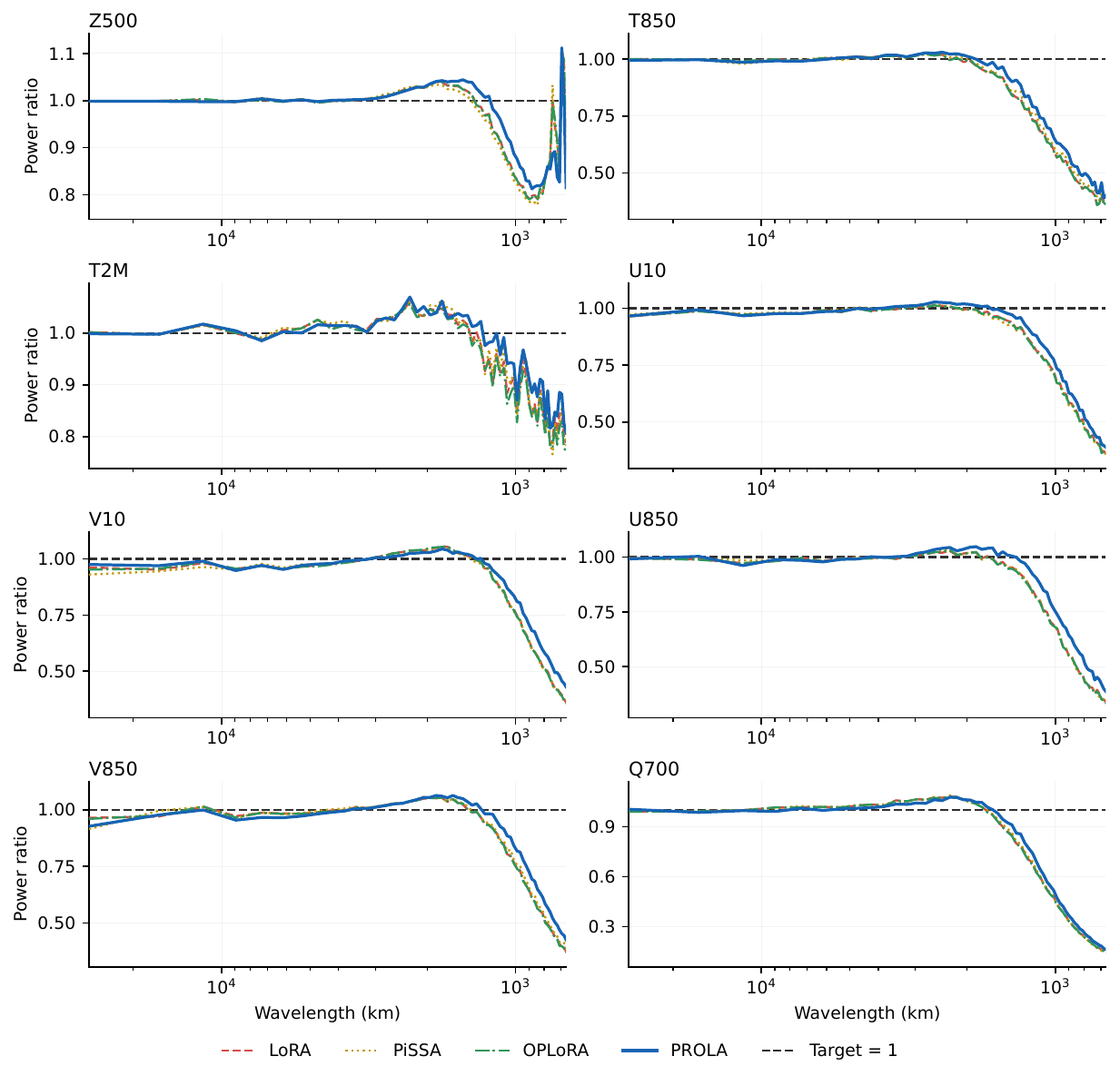}
\caption{\textbf{Spherical power ratios at 6h.} Ratios of mean prediction power to mean target power for eight variables, evaluated on all 1,458 test cases in 2018. The dashed horizontal line marks the target ratio of one.}
\label{fig:app_spectrum_6h}
\end{figure*}

\textbf{Spatial spectral fidelity.}
Table~\ref{tab:app_spectral_distance} summarizes spherical spectral fidelity for the five representative variables at 3h and 6h, using all 1,458 test cases in 2018 and the seed-0 models.
For each variable and lead time, let \(q_\ell\) be the ratio of mean prediction power to mean target power defined in Appendix~\ref{app:metrics}.
We report the mean absolute log-power ratio, where lower values indicate closer spectral agreement.
PROLA has the lowest distance in all ten reported variable--lead combinations.
For T2M at 6h, its distance is 0.062, compared with 0.079 for PiSSA, a reduction of 21.5\%.
Figures~\ref{fig:app_spectrum_3h} and~\ref{fig:app_spectrum_6h} show the corresponding power-ratio curves for eight variables.

\begin{table*}[t]
\centering
\small
\setlength{\tabcolsep}{2.5pt}
\renewcommand{\arraystretch}{1.10}
\caption{\textbf{Reconstruction and physical diagnostics at 6h.} Results use all 1,458 test cases in 2018. SNR denotes variance-normalized reconstruction SNR (dB); tendency RMSE is in \(\mathrm{m}^2\,\mathrm{s}^{-2}\) for Z500 and K for T2M; wind-derivative RMSE is in \(10^{-6}\,\mathrm{s}^{-1}\). Best displayed values are boldfaced, including rounded ties.}
\label{tab:app_physical_diagnostics}
\begin{tabularx}{\linewidth}{l|*{2}{>{\centering\arraybackslash}X}|*{2}{>{\centering\arraybackslash}X}|*{2}{>{\centering\arraybackslash}X}}
\toprule
\multirow{2}{*}{Method} & \multicolumn{2}{c|}{SNR $\uparrow$} & \multicolumn{2}{c|}{Tendency RMSE $\downarrow$} & \multicolumn{2}{c}{850-hPa wind RMSE $\downarrow$} \\
& Z500 & T2M & Z500 & T2M & Divergence & Vorticity \\
\midrule
LoRA & 30.78 & 21.46 & 40.48 & 0.418 & 8.70 & 15.53 \\
PiSSA & 30.87 & 21.62 & 40.26 & 0.417 & 8.61 & 15.45 \\
OPLoRA & 30.73 & 21.45 & 40.53 & 0.418 & 8.71 & 15.55 \\
\rowcolor{cyan!10}
PROLA & \textbf{31.48} & \textbf{21.87} & \textbf{39.89} & \textbf{0.415} & \textbf{8.46} & \textbf{15.18} \\
\bottomrule
\end{tabularx}
\end{table*}

\textbf{Reconstruction and physical diagnostics.}
Table~\ref{tab:app_physical_diagnostics} reports additional diagnostics at 6h on the same 1,458 cases.
Variance-normalized reconstruction SNR is computed for each case as $10\log_{10}\!\bigl(\operatorname{Var}_w(y)/ \operatorname{MSE}_w(\hat y,y)\bigr)$ and then averaged, where $w$ denotes latitude weighting.
For each variable, tendency error is the latitude-weighted RMSE between the
predicted and observed changes from $t+h_{\mathrm{end}}-\delta_f$ to
$t+h_{\mathrm{end}}$, corresponding to the 5h and 6h leads on this benchmark.
We also report RMSE of horizontal divergence and relative vorticity computed from the 850-hPa wind components using spherical finite differences, with periodic longitude and evaluation restricted to \(|\varphi|<89^\circ\).
For tendency and wind-derivative errors, squared errors are averaged over space and cases before taking the square root.
PROLA improves all six reported diagnostics, including Z500 and T2M reconstruction, their hourly evolution, and wind-field derivatives.

\begin{figure}[t]
\centering
\includegraphics[width=\linewidth]{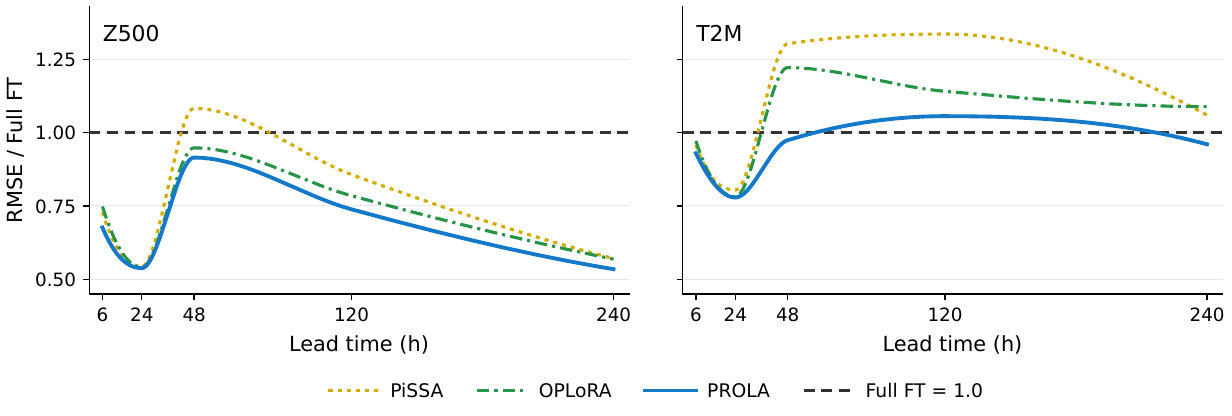}
\caption{\textbf{Relative RMSE over long rollouts.}
Z500 (left) and T2M (right) RMSE normalized by Full FT at each lead time.
The dashed horizontal line denotes Full FT at 1.0; lower values indicate lower RMSE.}
\label{fig:app_rollout_relative}
\end{figure}

\textbf{Long-horizon rollouts.}
Figure~\ref{fig:app_rollout_relative} compares Z500 and T2M RMSE at lead times from 6h to 240h, normalized by Full FT at each horizon.
PROLA has the lowest relative RMSE among the compared PEFT methods at all reported lead times for both variables.

\textbf{Full fine-tuning learning-rate sweep.}
\begin{table*}[t]
\centering
\small
\setlength{\tabcolsep}{5pt}
\renewcommand{\arraystretch}{1.15}
\caption{\textbf{Full fine-tuning learning-rate sweep.} Latitude-weighted RMSE ($\downarrow$) at 6h on the 2018 test split. The main Full FT configuration uses a backbone learning rate of \(5\times10^{-5}\). Best displayed values are boldfaced, including rounded ties.}
\label{tab:app_fullft_lr_sweep}
\begin{tabular}{l|rrrrr}
\toprule
Method / backbone LR & Z500 & T850 & T2M & U10 & V10 \\
\midrule
Full FT, \(2.5\times10^{-5}\) & 116.94 & 1.116 & 1.306 & 1.178 & 1.235 \\
Full FT, \(5\times10^{-5}\) & 108.90 & 1.105 & 1.304 & 1.122 & 1.203 \\
Full FT, \(10^{-4}\) & 106.06 & 1.096 & 1.324 & \textbf{1.093} & \textbf{1.165} \\
\midrule
\rowcolor{cyan!10}
PROLA & \textbf{73.07} & \textbf{0.958} & \textbf{1.211} & 1.236 & 1.234 \\

\bottomrule
\end{tabular}
\end{table*}

Table~\ref{tab:app_fullft_lr_sweep} compares three Full FT learning
rates under the same training budget.
PROLA outperforms every tested Full FT setting on Z500, T850,
and T2M, while Full FT achieves lower U10 and V10 errors.

\textbf{Training cost measurement.}
\begin{table}[t]
\centering
\small
\setlength{\tabcolsep}{6pt}
\renewcommand{\arraystretch}{1.08}
    \caption{
    \textbf{Training cost of the PEFT baselines and PROLA.}
    We report the average training time per epoch and the peak per-GPU memory recorded during training.
    }
\vspace{1pt}
\label{tab:app_training_cost}
\begin{tabular}{lcc}
\toprule
\textbf{Method} & \textbf{Training Time / Epoch} & \textbf{Peak GPU Memory} \\
\midrule
LoRA          & 54m 38s & 15.56 GiB \\
DoRA          & 71m 19s & 15.56 GiB \\
AdaLoRA       & 58m 45s & 15.63 GiB \\
PiSSA         & 55m 52s & 15.56 GiB \\
OPLoRA        & 60m 56s & 15.57 GiB \\
PROLA         & 67m 18s & 15.56 GiB \\
\bottomrule
\end{tabular}
\end{table}

Table~\ref{tab:app_training_cost} reports wall-clock training time per epoch and peak per-GPU memory for the main Aurora PEFT configuration on two NVIDIA RTX 3090 GPUs.
The time is computed from the recorded training-only runtime divided by the number of training epochs, excluding validation and separate test evaluation.
PROLA's runtime and peak GPU memory usage fall within the range observed for the PEFT baselines.

\section{Qualitative Results}
\label{app:qualitative}

\textbf{Global WeatherBench~2 forecasts.}
We provide qualitative comparisons on the 2018 test split of
WeatherBench~2~\cite{rasp2024weatherbench}.
Predictions within each comparison use the same forecast initialization;
the global examples shown here are initialized at 18:00 UTC
on 10 January 2018.
For each map, all predictions and targets are shown after denormalization
to physical units.
Prediction and target panels for each variable and lead time share
the same color scale.
The absolute-error difference map is defined as
\begin{equation}
\Delta_{\mathrm{AE}}(\mathbf{s})
=
\left|
\hat{y}_{\mathrm{Ours}}(\mathbf{s}) - y(\mathbf{s})
\right|
-
\left|
\hat{y}_{\mathrm{Base}}(\mathbf{s}) - y(\mathbf{s})
\right|,
\end{equation}
where $\mathbf{s}$ is a grid point. For the displayed variable and lead $h$,
$\hat y_{\mathrm{Ours}}(\mathbf{s})$ and $y(\mathbf{s})$ are the corresponding
values of $\hat{\mathbf{y}}^{ST}_{t+h}$ and $\mathbf{y}^{\mathrm{HR}}_{t+h}$ at $\mathbf{s}$.
Here, $\mathrm{Base}$ denotes the comparison method shown in the
corresponding panel or column.
Thus, negative values indicate locations where PROLA has a smaller
absolute error than the compared baseline, while positive values
indicate locations where the baseline has a smaller absolute error.

\begin{figure}[t]
\centering
\includegraphics[width=0.8\linewidth]{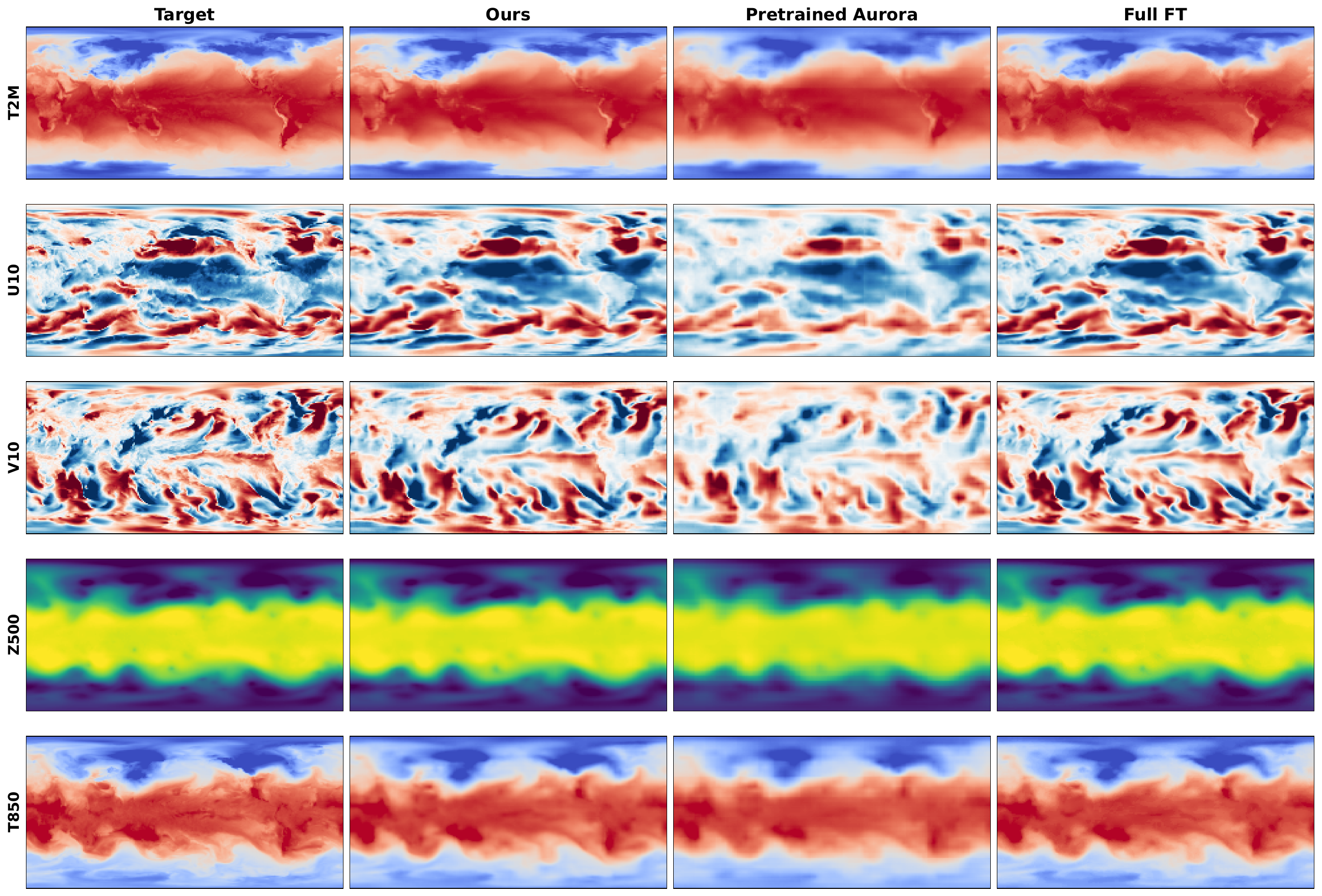}
\caption{Qualitative comparison at 2h lead time for multiple variables, showing the target, PROLA, pretrained Aurora, and Full FT predictions.}
\label{fig:supp_qual_overview_tplus2h}
\end{figure}

\begin{figure}[t]
\centering
\includegraphics[width=0.8\linewidth]{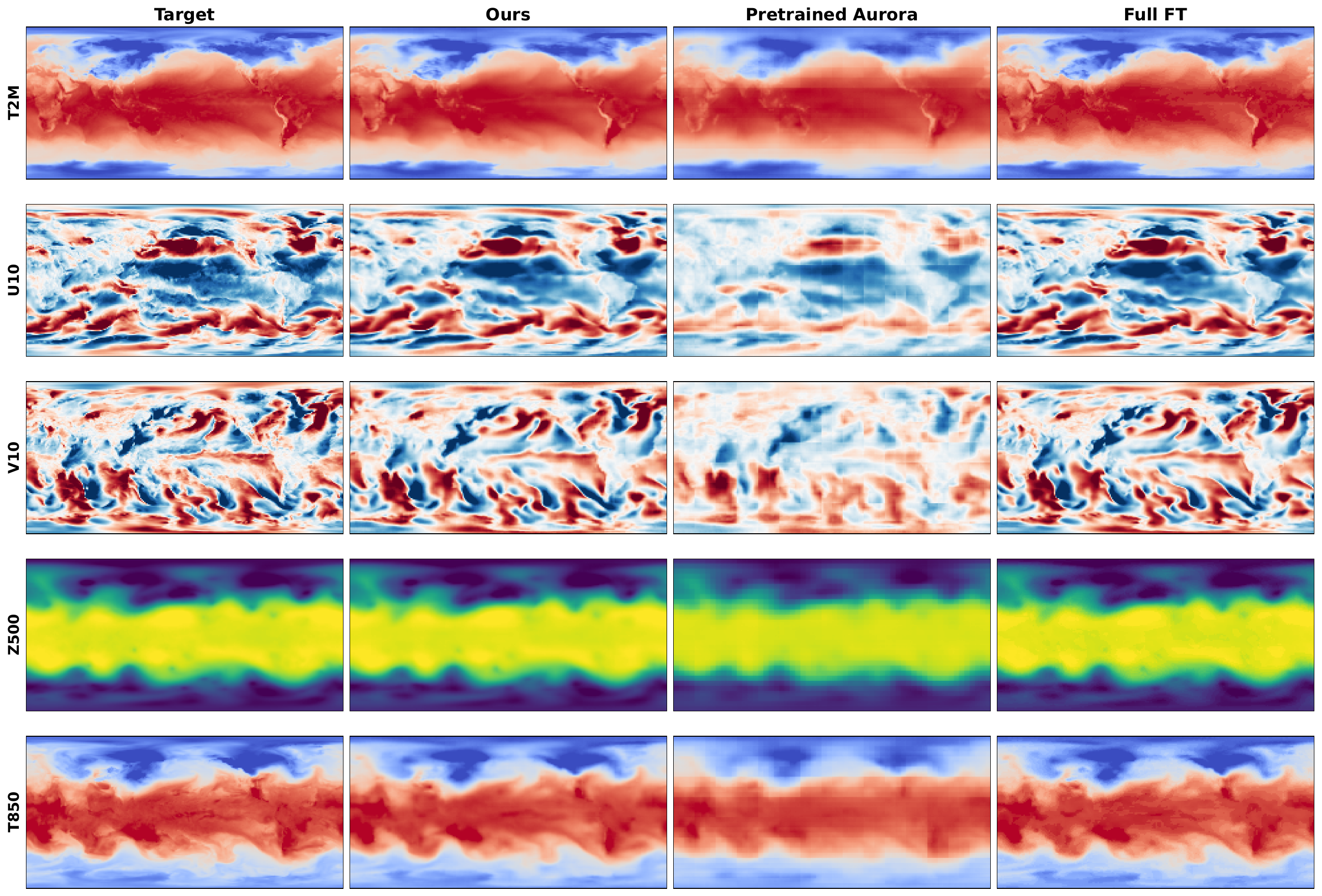}
\caption{Qualitative comparison at 4h lead time for multiple variables, showing the target, PROLA, pretrained Aurora, and Full FT predictions.}
\label{fig:supp_qual_overview_tplus4h}
\end{figure}

\begin{figure}[t]
\centering
\includegraphics[width=0.8\linewidth]{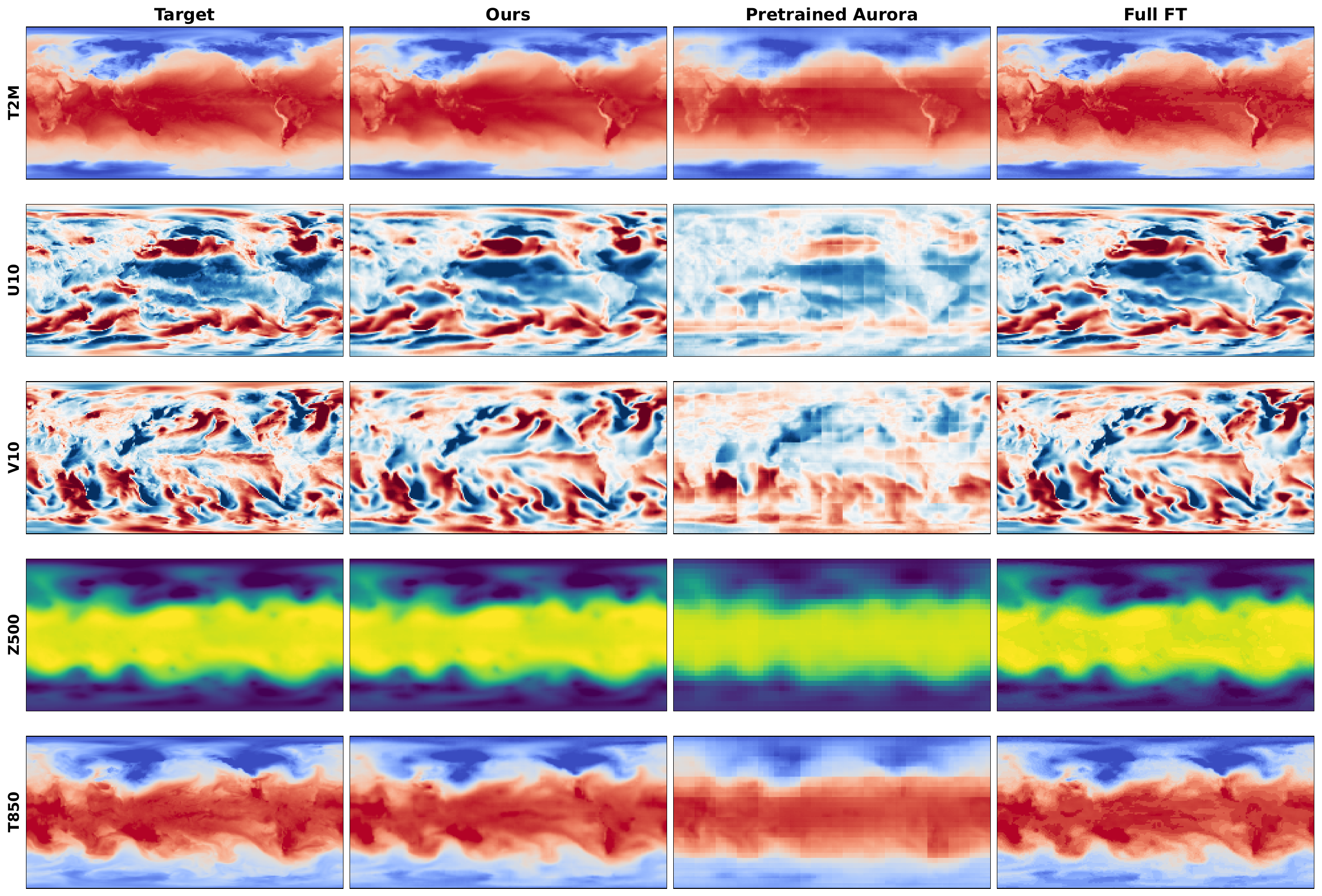}
\caption{Qualitative comparison at 6h lead time for multiple variables, showing the target, PROLA, pretrained Aurora, and Full FT predictions.}
\label{fig:supp_qual_overview_tplus6h}
\end{figure}

\begin{figure}[t]
\centering
\includegraphics[width=0.9\linewidth]{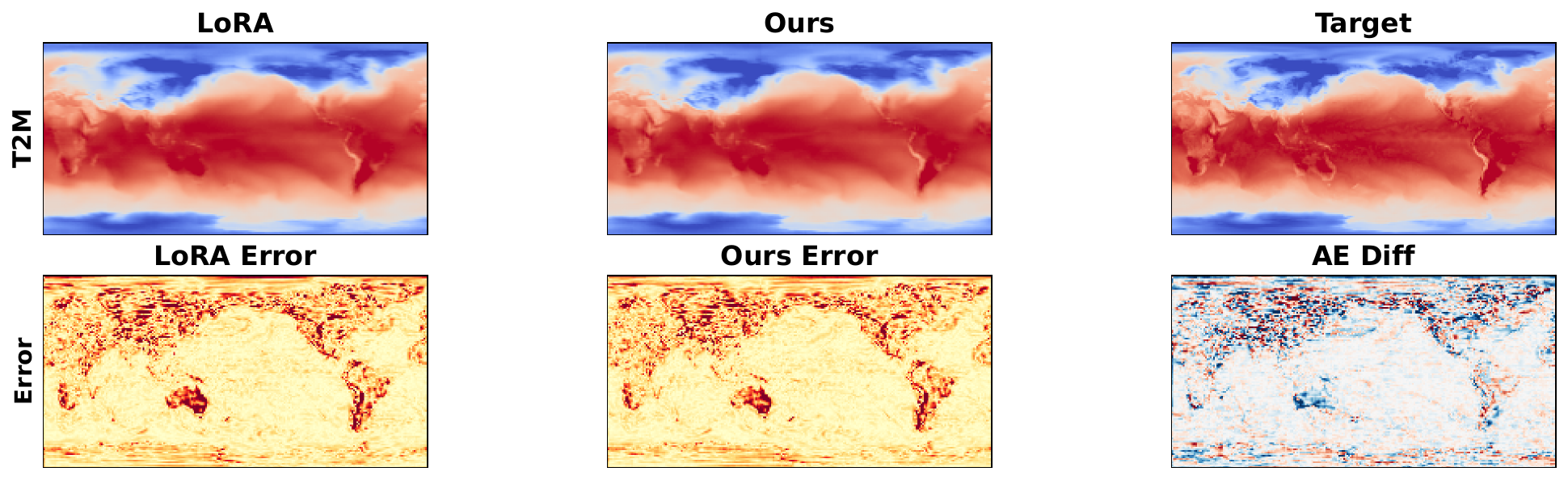}
\caption{T2M comparison at 2h lead time between LoRA and PROLA, with absolute-error maps and the pointwise absolute-error difference map.}
\label{fig:supp_qual_lora_t2m_tplus2h}
\end{figure}

\begin{figure}[t]
\centering
\includegraphics[width=0.9\linewidth]{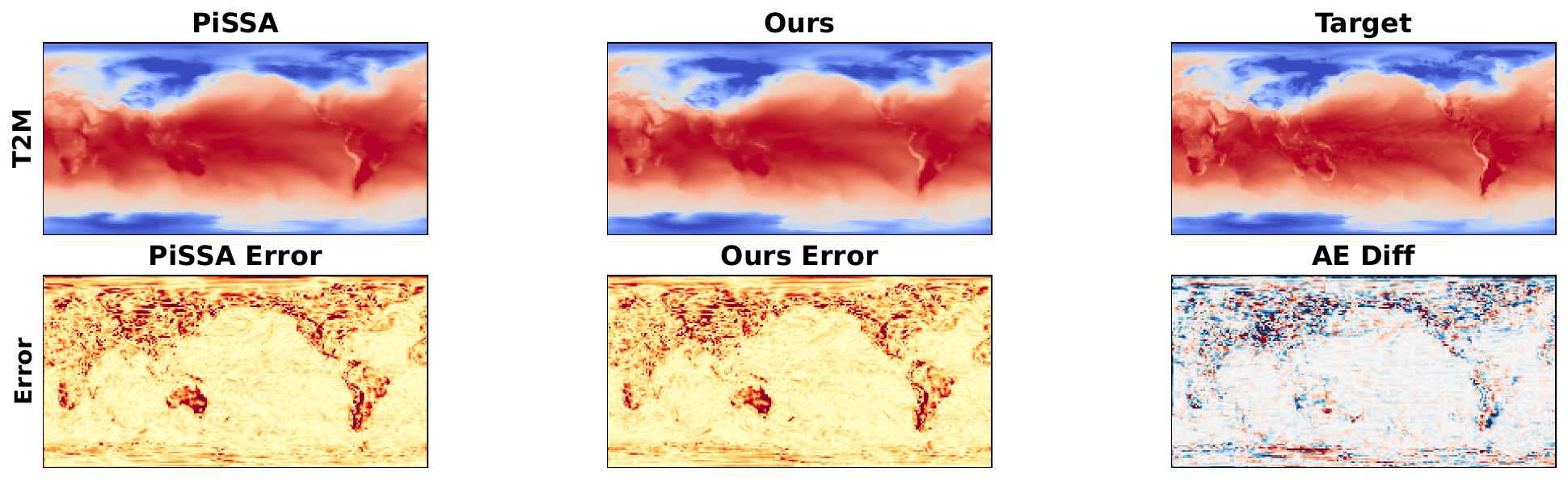}
\caption{T2M comparison at 2h lead time between PiSSA and PROLA, with absolute-error maps and the pointwise absolute-error difference map.}
\label{fig:supp_qual_pissa_t2m_tplus2h}
\end{figure}

\begin{figure}[t]
\centering
\includegraphics[width=0.9\linewidth]{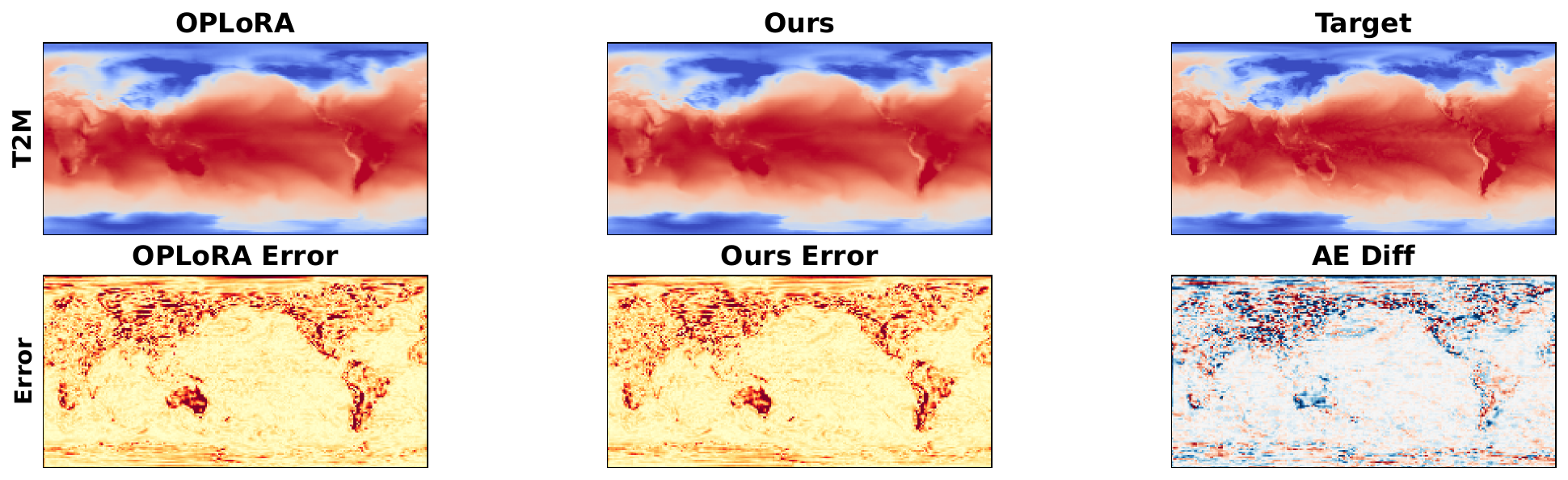}
\caption{T2M comparison at 2h lead time between OPLoRA and PROLA, with absolute-error maps and the pointwise absolute-error difference map.}
\label{fig:supp_qual_oplora_t2m_tplus2h}
\end{figure}

\begin{figure}[t]
\centering
\includegraphics[width=0.9\linewidth]{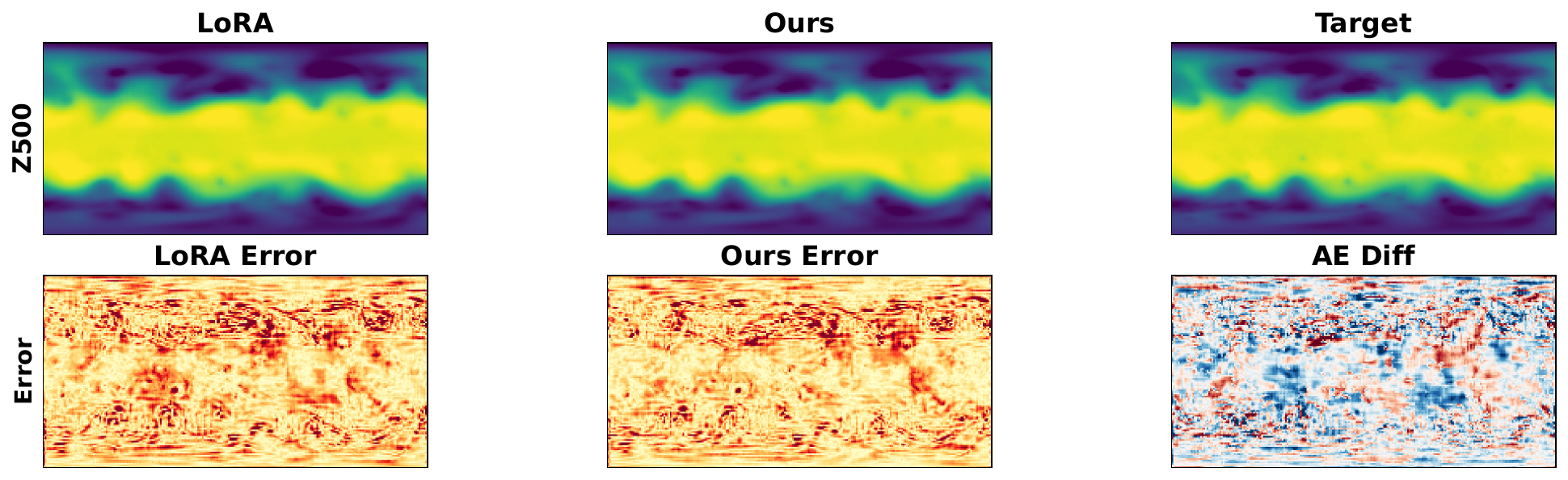}
\caption{Z500 comparison at 2h lead time between LoRA and PROLA, with absolute-error maps and the pointwise absolute-error difference map.}
\label{fig:supp_qual_lora_z500_tplus2h}
\end{figure}

\begin{figure}[t]
\centering
\includegraphics[width=0.9\linewidth]{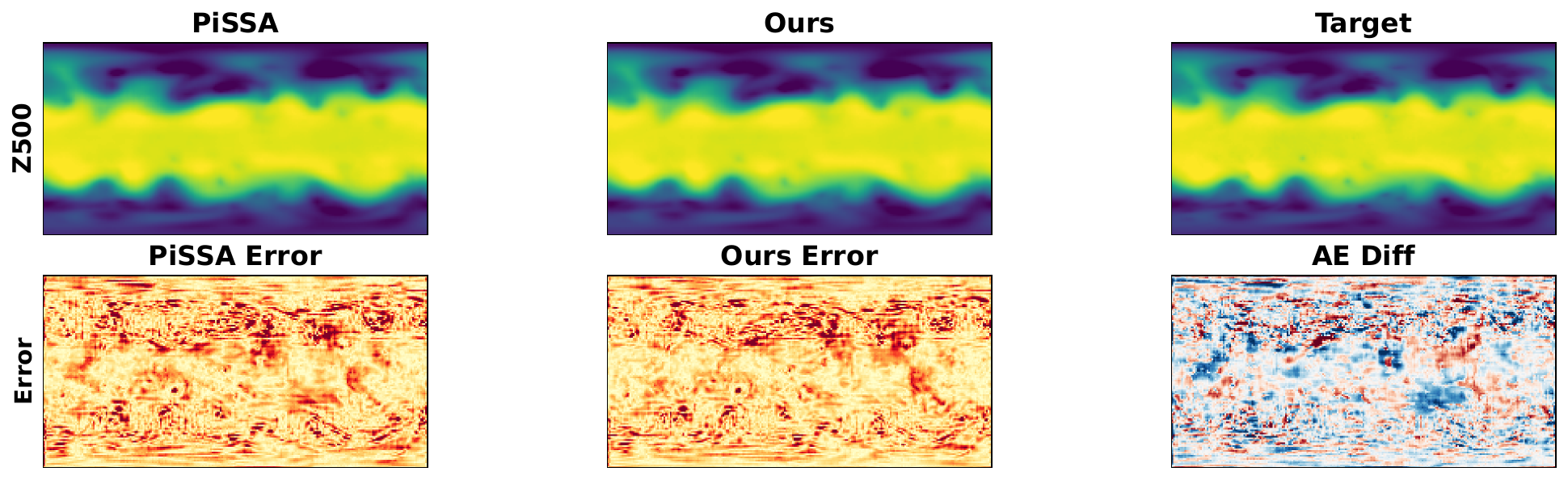}
\caption{Z500 comparison at 2h lead time between PiSSA and PROLA, with absolute-error maps and the pointwise absolute-error difference map.}
\label{fig:supp_qual_pissa_z500_tplus2h}
\end{figure}

\begin{figure}[t]
\centering
\includegraphics[width=0.9\linewidth]{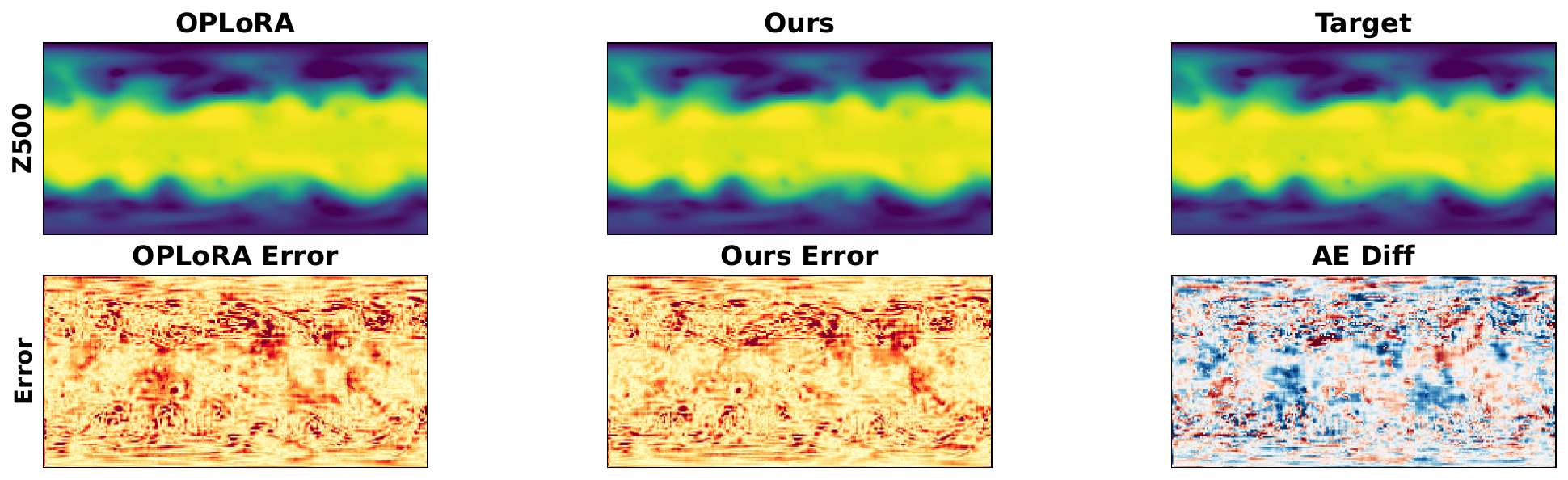}
\caption{Z500 comparison at 2h lead time between OPLoRA and PROLA, with absolute-error maps and the pointwise absolute-error difference map.}
\label{fig:supp_qual_oplora_z500_tplus2h}
\end{figure}

Figures~\ref{fig:supp_qual_overview_tplus2h},
\ref{fig:supp_qual_overview_tplus4h},
and~\ref{fig:supp_qual_overview_tplus6h} show qualitative prediction
results at 2h, 4h, and 6h lead times across the five reported variables,
comparing PROLA, the pretrained Aurora~\cite{bodnar2025foundation}
backbone, and Full FT against the ground-truth target.
The pretrained Aurora baseline spatially upsamples the current coarse
state and the pretrained 6h forecast, then interpolates between them
to obtain the intermediate 2h and 4h fields, without learned
correction heads.
Figures~\ref{fig:supp_qual_lora_t2m_tplus2h},
\ref{fig:supp_qual_pissa_t2m_tplus2h},
and~\ref{fig:supp_qual_oplora_t2m_tplus2h} show pairwise T2M comparisons
at 2h lead time between PROLA and three PEFT baselines
(LoRA~\cite{hu2022lora}, PiSSA~\cite{meng2024pissa},
and OPLoRA~\cite{xiong2026oplora}), each accompanied by absolute-error
maps and pointwise absolute-error difference maps.
Figures~\ref{fig:supp_qual_lora_z500_tplus2h},
\ref{fig:supp_qual_pissa_z500_tplus2h},
and~\ref{fig:supp_qual_oplora_z500_tplus2h} show the analogous Z500
comparisons at 2h lead time between PROLA and the same three
PEFT baselines.

\begin{figure*}[t]
\centering
\includegraphics[width=0.97\textwidth]{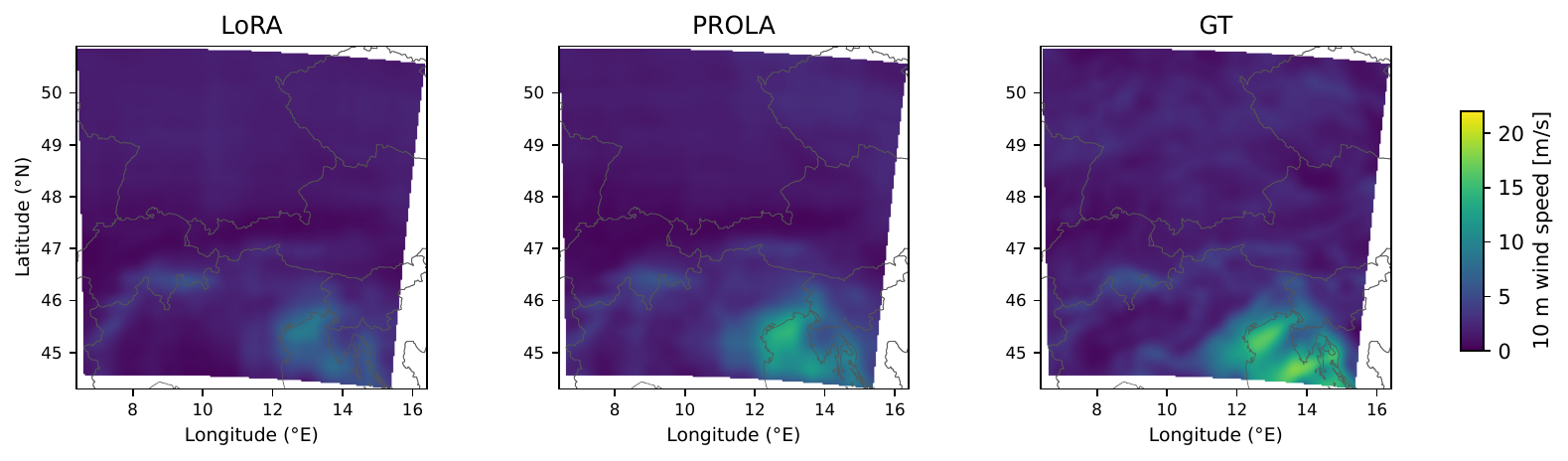}
\caption{\textbf{Qualitative ERA5-to-CERRA wind prediction at 6h.}
10\,m wind speed from LoRA, PROLA, and the CERRA target, valid at 06:00 UTC on 3 February 2018.
All panels share the same color scale and display smoothing.}
\label{fig:supp_cerra_wind_6h}
\end{figure*}

\textbf{Regional ERA5-to-CERRA forecasts.}
Figure~\ref{fig:cerra_qualitative} in the main paper compares LoRA
and PROLA on a T2M forecast initialized at 06:00 UTC on
4 February 2018 and valid six hours later.
Figure~\ref{fig:supp_cerra_wind_6h} provides an additional 10\,m
wind-speed comparison initialized at 00:00 UTC on 3 February 2018
and valid at 06:00 UTC.
Wind speed is computed as \(\sqrt{u_{10}^{2}+v_{10}^{2}}\) from the
predicted or target wind components.
The LoRA, PROLA, and CERRA target panels share the same geographic
extent, color scale, and display smoothing within each comparison.
Quantitative errors are computed on the native fields before
display smoothing.

\section{Additional Ablations}
\label{app:additional_ablations}

\begin{table*}[t]
\centering
\scriptsize
\setlength{\tabcolsep}{2.4pt}
\renewcommand{\arraystretch}{1.10}
\caption{\textbf{Direct prediction across PEFT methods.} We report latitude-weighted RMSE ($\downarrow$) at 1--6h. Best displayed values, including ties, are boldfaced.}
\label{tab:direct_full_leads}
\resizebox{0.6\textwidth}{!}{%
\begin{tabular}{ll|rrrrrr}
\toprule
Method & Variable & \multicolumn{6}{c}{RMSE $\downarrow$} \\
 &  & 1h & 2h & 3h & 4h & 5h & 6h \\
\midrule
 & Z500 & \textbf{101.90} & 72.91 & 58.91 & 72.14 & 98.30 & 129.60 \\
 & T850 & 1.009 & 0.973 & 0.979 & 1.044 & 1.118 & 1.214 \\
 & T2M & 1.419 & 1.303 & 1.256 & 1.230 & 1.294 & 1.394 \\
 & U10 & 1.426 & 1.438 & 1.481 & 1.567 & 1.652 & 1.755 \\
\multirow{-5}{*}{LoRA} & V10 & 1.376 & 1.417 & 1.484 & 1.585 & 1.677 & 1.781 \\
\midrule
 & Z500 & 104.90 & 73.76 & 56.47 & 68.45 & 95.47 & 128.00 \\
 & T850 & 1.009 & 0.955 & 0.942 & 0.992 & 1.058 & 1.152 \\
 & T2M & \textbf{1.416} & 1.261 & 1.191 & 1.134 & 1.185 & 1.292 \\
 & U10 & 1.332 & 1.292 & 1.272 & 1.298 & 1.327 & 1.392 \\
\multirow{-5}{*}{PiSSA} & V10 & 1.272 & 1.243 & 1.239 & 1.280 & 1.324 & 1.405 \\
\midrule
 & Z500 & 103.29 & 73.17 & 58.15 & 71.78 & 98.97 & 131.16 \\
 & T850 & 1.013 & 0.977 & 0.983 & 1.047 & 1.122 & 1.219 \\
 & T2M & 1.426 & 1.304 & 1.254 & 1.224 & 1.288 & 1.388 \\
 & U10 & 1.409 & 1.420 & 1.463 & 1.551 & 1.637 & 1.741 \\
\multirow{-5}{*}{OPLoRA} & V10 & 1.361 & 1.396 & 1.457 & 1.551 & 1.636 & 1.735 \\
\midrule
\rowcolor{cyan!8}
 & Z500 & 104.37 & \textbf{72.09} & \textbf{53.13} & \textbf{64.82} & \textbf{92.24} & \textbf{125.07} \\
\rowcolor{cyan!8}
 & T850 & \textbf{0.991} & \textbf{0.924} & \textbf{0.898} & \textbf{0.941} & \textbf{1.002} & \textbf{1.095} \\
\rowcolor{cyan!8}
 & T2M & 1.432 & \textbf{1.245} & \textbf{1.165} & \textbf{1.109} & \textbf{1.174} & \textbf{1.289} \\
\rowcolor{cyan!8}
 & U10 & \textbf{1.308} & \textbf{1.247} & \textbf{1.211} & \textbf{1.233} & \textbf{1.271} & \textbf{1.358} \\
\rowcolor{cyan!8}
\multirow{-5}{*}{PROLA} & V10 & \textbf{1.245} & \textbf{1.202} & \textbf{1.177} & \textbf{1.203} & \textbf{1.244} & \textbf{1.344} \\
\bottomrule
\end{tabular}}
\end{table*}

\begin{table*}[t]
\centering
\scriptsize
\setlength{\tabcolsep}{2.4pt}
\renewcommand{\arraystretch}{1.10}
\caption{\textbf{Full ablation results on scaffold-and-correct pipeline design.} We report latitude-weighted RMSE ($\downarrow$) and ACC ($\uparrow$) at 1--6h. Best displayed values, including ties, are boldfaced.}
\label{tab:ablation_head_full_leads}
\resizebox{\textwidth}{!}{%
\begin{tabular}{ll|rrrrrr|rrrrrr}
\toprule
Method / Variant & Variable & \multicolumn{6}{c|}{RMSE $\downarrow$} & \multicolumn{6}{c}{ACC $\uparrow$} \\
 &  & 1h & 2h & 3h & 4h & 5h & 6h & 1h & 2h & 3h & 4h & 5h & 6h \\
\midrule
 & Z500 & 104.37 & 72.09 & 53.13 & 64.82 & 92.24 & 125.07 & 0.995 & 0.997 & \textbf{0.999} & 0.998 & 0.996 & 0.992 \\
 & T850 & 0.991 & 0.924 & 0.898 & 0.941 & 1.002 & 1.095 & 0.980 & 0.983 & 0.984 & 0.982 & 0.980 & 0.976 \\
 & T2M & 1.432 & 1.245 & 1.165 & 1.109 & 1.174 & 1.289 & 0.960 & 0.970 & 0.974 & 0.976 & 0.974 & 0.968 \\
 & U10 & 1.308 & 1.247 & 1.211 & 1.233 & 1.271 & 1.358 & 0.946 & 0.951 & 0.954 & 0.952 & 0.949 & 0.942 \\
\multirow{-5}{*}{Direct (PROLA)} & V10 & 1.245 & 1.202 & 1.177 & 1.203 & 1.244 & 1.344 & 0.953 & 0.956 & 0.958 & 0.956 & 0.953 & 0.945 \\
\midrule
 & Z500 & 155.56 & 120.51 & 87.73 & 65.77 & 71.87 & 111.87 & 0.989 & 0.993 & 0.996 & 0.998 & \textbf{0.998} & 0.994 \\
 & T850 & 1.226 & 1.076 & 0.959 & 0.922 & 0.935 & 1.037 & 0.970 & 0.977 & 0.982 & 0.983 & 0.982 & 0.979 \\
 & T2M & 1.745 & 1.527 & 1.307 & 1.127 & 1.088 & 1.289 & 0.940 & 0.955 & 0.967 & 0.976 & 0.977 & 0.971 \\
 & U10 & 1.665 & 1.455 & 1.277 & 1.189 & 1.194 & 1.339 & 0.915 & 0.936 & 0.950 & 0.956 & 0.956 & 0.948 \\
\multirow{-5}{*}{w/o $C_s,C_t$} & V10 & 1.694 & 1.464 & 1.270 & 1.175 & 1.196 & 1.371 & 0.913 & 0.936 & 0.952 & 0.958 & 0.957 & 0.946 \\
\midrule
 & Z500 & 154.94 & 118.93 & 84.53 & 60.31 & 66.47 & 108.80 & 0.989 & 0.994 & 0.997 & 0.998 & \textbf{0.998} & 0.995 \\
 & T850 & 1.219 & 1.062 & 0.938 & 0.896 & 0.909 & 1.016 & 0.971 & 0.978 & 0.982 & \textbf{0.984} & \textbf{0.983} & 0.980 \\
 & T2M & 1.740 & 1.515 & 1.290 & 1.107 & \textbf{1.070} & 1.281 & 0.941 & 0.955 & 0.968 & 0.976 & \textbf{0.978} & 0.971 \\
 & U10 & 1.659 & 1.442 & 1.259 & 1.169 & 1.177 & 1.332 & 0.915 & 0.937 & 0.952 & 0.957 & \textbf{0.957} & 0.949 \\
\multirow{-5}{*}{w/o $C_t$} & V10 & 1.688 & 1.452 & 1.254 & 1.159 & 1.183 & 1.365 & 0.914 & 0.938 & 0.954 & 0.959 & \textbf{0.958} & 0.947 \\
\midrule
 & Z500 & 55.84 & 50.74 & 53.35 & 58.16 & 62.84 & 78.25 & \textbf{0.999} & \textbf{0.999} & \textbf{0.999} & 0.998 & \textbf{0.998} & \textbf{0.997} \\
 & T850 & 0.942 & 0.904 & 0.880 & 0.900 & 0.920 & 0.974 & 0.982 & \textbf{0.984} & 0.984 & \textbf{0.984} & \textbf{0.983} & 0.981 \\
 & T2M & 1.374 & 1.213 & 1.138 & 1.069 & 1.109 & \textbf{1.209} & \textbf{0.963} & \textbf{0.972} & 0.975 & \textbf{0.978} & 0.976 & \textbf{0.972} \\
 & U10 & 1.250 & 1.193 & 1.155 & 1.165 & 1.178 & \textbf{1.233} & 0.951 & 0.955 & 0.958 & 0.957 & 0.956 & \textbf{0.952} \\
\multirow{-5}{*}{w/o $C_s$} & V10 & 1.199 & 1.155 & 1.133 & 1.157 & 1.179 & 1.237 & 0.956 & 0.960 & 0.961 & 0.959 & \textbf{0.958} & 0.953 \\
\midrule
\rowcolor{cyan!8}
 & Z500 & \textbf{52.75} & \textbf{47.15} & \textbf{50.30} & \textbf{55.23} & \textbf{58.78} & \textbf{73.07} & \textbf{0.999} & \textbf{0.999} & \textbf{0.999} & \textbf{0.999} & \textbf{0.998} & \textbf{0.997} \\
\rowcolor{cyan!8}
 & T850 & \textbf{0.924} & \textbf{0.887} & \textbf{0.866} & \textbf{0.888} & \textbf{0.907} & \textbf{0.958} & \textbf{0.983} & \textbf{0.984} & \textbf{0.985} & \textbf{0.984} & \textbf{0.983} & \textbf{0.982} \\
\rowcolor{cyan!8}
 & T2M & \textbf{1.373} & \textbf{1.205} & \textbf{1.128} & \textbf{1.060} & 1.107 & 1.211 & \textbf{0.963} & \textbf{0.972} & \textbf{0.976} & \textbf{0.978} & 0.976 & \textbf{0.972} \\
\rowcolor{cyan!8}
 & U10 & \textbf{1.226} & \textbf{1.174} & \textbf{1.140} & \textbf{1.154} & \textbf{1.173} & 1.236 & \textbf{0.953} & \textbf{0.957} & \textbf{0.959} & \textbf{0.958} & \textbf{0.957} & \textbf{0.952} \\
\rowcolor{cyan!8}
\multirow{-5}{*}{Full pipeline} & V10 & \textbf{1.176} & \textbf{1.136} & \textbf{1.119} & \textbf{1.146} & \textbf{1.171} & \textbf{1.234} & \textbf{0.958} & \textbf{0.961} & \textbf{0.962} & \textbf{0.960} & \textbf{0.958} & \textbf{0.954} \\
\bottomrule
\end{tabular}}
\end{table*}

\textbf{Full pipeline ablations.}
Table~\ref{tab:direct_full_leads} extends the Direct prediction comparison
in Table~\ref{tab:ablation_head} to all lead times from 1h to 6h for
LoRA, PiSSA, OPLoRA, and PROLA.
Table~\ref{tab:ablation_head_full_leads} provides the full per-lead
pipeline ablation with PROLA.
The compared variants are Direct prediction, the scaffold-only model
without \(C_s\) and \(C_t\), the model without \(C_t\), the model without
\(C_s\), and the full scaffold-and-correct pipeline.
The full pipeline achieves the lowest RMSE across all five variables
at 3h, with consistent gains on Z500 and T850 across lead times.
Unless otherwise stated, the ablations use seed 0 and the same training schedule as the main comparison.

\begin{table*}[t]
\centering
\scriptsize
\setlength{\tabcolsep}{2.4pt}
\renewcommand{\arraystretch}{1.10}
\caption{\textbf{Full ablation results on PROLA branch composition and update scaling.} We report latitude-weighted RMSE ($\downarrow$) and ACC ($\uparrow$) at 1--6h. Best displayed values, including ties, are boldfaced. All configurations use $r_p+r_o=8$; checkmarks indicate update scaling.}
\label{tab:ablation_update_full_leads}
\resizebox{\textwidth}{!}{%
\begin{tabular}{lcccl|rrrrrr|rrrrrr}
\toprule
Configuration & $r_p$ & $r_o$ & Scaling & Variable & \multicolumn{6}{c|}{RMSE $\downarrow$} & \multicolumn{6}{c}{ACC $\uparrow$} \\
 &  &  &  &  & 1h & 2h & 3h & 4h & 5h & 6h & 1h & 2h & 3h & 4h & 5h & 6h \\
\midrule
 &  &  &  & Z500 & 55.57 & 50.45 & 53.66 & 58.98 & 63.55 & 78.55 & \textbf{0.999} & \textbf{0.999} & \textbf{0.999} & 0.998 & \textbf{0.998} & \textbf{0.997} \\
 &  &  &  & T850 & 0.941 & 0.910 & 0.894 & 0.918 & 0.937 & 0.987 & 0.982 & 0.983 & 0.984 & 0.983 & 0.982 & 0.980 \\
 &  &  &  & T2M & 1.383 & 1.258 & 1.193 & 1.125 & 1.152 & 1.248 & 0.963 & 0.969 & 0.973 & 0.976 & 0.975 & 0.970 \\
 &  &  &  & U10 & 1.288 & 1.240 & 1.209 & 1.225 & 1.241 & 1.294 & 0.948 & 0.952 & 0.954 & 0.953 & 0.952 & 0.947 \\
\multirow{-5}{*}{Principal} & \multirow{-5}{*}{8} & \multirow{-5}{*}{0} & \multirow{-5}{*}{--} & V10 & 1.227 & 1.194 & 1.178 & 1.204 & 1.228 & 1.290 & 0.954 & 0.957 & 0.958 & 0.956 & 0.954 & 0.949 \\
\midrule
 &  &  &  & Z500 & 54.94 & 49.80 & 53.06 & 58.35 & 62.77 & 77.71 & \textbf{0.999} & \textbf{0.999} & \textbf{0.999} & 0.998 & \textbf{0.998} & \textbf{0.997} \\
 &  &  &  & T850 & 0.935 & 0.904 & 0.887 & 0.911 & 0.930 & 0.979 & 0.982 & \textbf{0.984} & 0.984 & 0.983 & \textbf{0.983} & 0.981 \\
 &  &  &  & T2M & 1.378 & 1.251 & 1.184 & 1.116 & 1.144 & 1.241 & 0.963 & 0.970 & 0.973 & 0.976 & 0.975 & 0.970 \\
 &  &  &  & U10 & 1.275 & 1.228 & 1.198 & 1.214 & 1.232 & 1.289 & 0.949 & 0.953 & 0.955 & 0.953 & 0.952 & 0.948 \\
\multirow{-5}{*}{Principal} & \multirow{-5}{*}{8} & \multirow{-5}{*}{0} & \multirow{-5}{*}{\cmark} & V10 & 1.213 & 1.181 & 1.166 & 1.194 & 1.221 & 1.285 & 0.955 & 0.958 & 0.959 & 0.956 & 0.954 & 0.950 \\
\midrule
 &  &  &  & Z500 & 55.59 & 51.05 & 54.54 & 60.12 & 64.98 & 79.83 & \textbf{0.999} & \textbf{0.999} & \textbf{0.999} & 0.998 & \textbf{0.998} & \textbf{0.997} \\
 &  &  &  & T850 & 0.942 & 0.913 & 0.899 & 0.924 & 0.944 & 0.994 & 0.982 & 0.983 & 0.984 & 0.983 & 0.982 & 0.980 \\
 &  &  &  & T2M & 1.384 & 1.261 & 1.197 & 1.138 & 1.174 & 1.272 & 0.963 & 0.969 & 0.972 & 0.975 & 0.974 & 0.969 \\
 &  &  &  & U10 & 1.290 & 1.243 & 1.216 & 1.235 & 1.255 & 1.314 & 0.948 & 0.952 & 0.954 & 0.952 & 0.950 & 0.946 \\
\multirow{-5}{*}{Orthogonal} & \multirow{-5}{*}{0} & \multirow{-5}{*}{8} & \multirow{-5}{*}{--} & V10 & 1.233 & 1.201 & 1.188 & 1.217 & 1.246 & 1.314 & 0.954 & 0.956 & 0.957 & 0.955 & 0.953 & 0.947 \\
\midrule
 &  &  &  & Z500 & 54.63 & 49.55 & 52.81 & 58.11 & 62.32 & 76.67 & \textbf{0.999} & \textbf{0.999} & \textbf{0.999} & 0.998 & \textbf{0.998} & \textbf{0.997} \\
 &  &  &  & T850 & 0.934 & 0.903 & 0.886 & 0.910 & 0.929 & 0.977 & 0.982 & \textbf{0.984} & 0.984 & 0.983 & \textbf{0.983} & 0.981 \\
 &  &  &  & T2M & 1.409 & 1.254 & 1.175 & 1.108 & 1.155 & 1.260 & 0.961 & 0.970 & 0.973 & 0.976 & 0.974 & 0.970 \\
 &  &  &  & U10 & 1.275 & 1.223 & 1.189 & 1.204 & 1.223 & 1.284 & 0.949 & 0.953 & 0.956 & 0.954 & 0.953 & 0.948 \\
\multirow{-5}{*}{Orthogonal} & \multirow{-5}{*}{0} & \multirow{-5}{*}{8} & \multirow{-5}{*}{\cmark} & V10 & 1.203 & 1.169 & 1.157 & 1.188 & 1.216 & 1.278 & 0.956 & 0.959 & 0.960 & 0.957 & 0.955 & 0.950 \\
\midrule
 &  &  &  & Z500 & 53.55 & 48.22 & 51.36 & 56.31 & 60.04 & 74.38 & \textbf{0.999} & \textbf{0.999} & \textbf{0.999} & 0.998 & \textbf{0.998} & \textbf{0.997} \\
 &  &  &  & T850 & 0.930 & 0.894 & 0.875 & 0.898 & 0.917 & 0.967 & \textbf{0.983} & \textbf{0.984} & \textbf{0.985} & \textbf{0.984} & \textbf{0.983} & 0.981 \\
 &  &  &  & T2M & \textbf{1.347} & 1.207 & 1.143 & 1.080 & 1.120 & 1.221 & \textbf{0.965} & \textbf{0.972} & 0.975 & \textbf{0.978} & \textbf{0.976} & 0.971 \\
 &  &  &  & U10 & 1.248 & 1.198 & 1.165 & 1.179 & 1.196 & 1.256 & 0.951 & 0.955 & 0.958 & 0.956 & 0.955 & 0.950 \\
\multirow{-5}{*}{Both branches} & \multirow{-5}{*}{4} & \multirow{-5}{*}{4} & \multirow{-5}{*}{--} & V10 & 1.187 & 1.149 & 1.133 & 1.163 & 1.188 & 1.250 & 0.957 & 0.960 & 0.961 & 0.959 & 0.957 & 0.952 \\
\midrule
\rowcolor{cyan!8}
 &  &  &  & Z500 & \textbf{52.75} & \textbf{47.15} & \textbf{50.30} & \textbf{55.23} & \textbf{58.78} & \textbf{73.07} & \textbf{0.999} & \textbf{0.999} & \textbf{0.999} & \textbf{0.999} & \textbf{0.998} & \textbf{0.997} \\
\rowcolor{cyan!8}
 &  &  &  & T850 & \textbf{0.924} & \textbf{0.887} & \textbf{0.866} & \textbf{0.888} & \textbf{0.907} & \textbf{0.958} & \textbf{0.983} & \textbf{0.984} & \textbf{0.985} & \textbf{0.984} & \textbf{0.983} & \textbf{0.982} \\
\rowcolor{cyan!8}
 &  &  &  & T2M & 1.373 & \textbf{1.205} & \textbf{1.128} & \textbf{1.060} & \textbf{1.107} & \textbf{1.211} & 0.963 & \textbf{0.972} & \textbf{0.976} & \textbf{0.978} & \textbf{0.976} & \textbf{0.972} \\
\rowcolor{cyan!8}
 &  &  &  & U10 & \textbf{1.226} & \textbf{1.174} & \textbf{1.140} & \textbf{1.154} & \textbf{1.173} & \textbf{1.236} & \textbf{0.953} & \textbf{0.957} & \textbf{0.959} & \textbf{0.958} & \textbf{0.957} & \textbf{0.952} \\
\rowcolor{cyan!8}
\multirow{-5}{*}{Complete} & \multirow{-5}{*}{4} & \multirow{-5}{*}{4} & \multirow{-5}{*}{\cmark} & V10 & \textbf{1.176} & \textbf{1.136} & \textbf{1.119} & \textbf{1.146} & \textbf{1.171} & \textbf{1.234} & \textbf{0.958} & \textbf{0.961} & \textbf{0.962} & \textbf{0.960} & \textbf{0.958} & \textbf{0.954} \\
\bottomrule
\end{tabular}}
\end{table*}

\begin{table*}[t]
\centering
\scriptsize
\setlength{\tabcolsep}{2.4pt}
\renewcommand{\arraystretch}{1.10}
\caption{\textbf{Full ablation results on rank allocation and scaling components.} We report latitude-weighted RMSE ($\downarrow$) and ACC ($\uparrow$) at 1--6h. Best displayed values, including ties, are boldfaced. All configurations use $r_p+r_o=8$; checkmarks indicate update scaling.}
\label{tab:ablation_allocation_full_leads}
\resizebox{\textwidth}{!}{%
\begin{tabular}{lcccl|rrrrrr|rrrrrr}
\toprule
Configuration & $r_p$ & $r_o$ & Scaling & Variable & \multicolumn{6}{c|}{RMSE $\downarrow$} & \multicolumn{6}{c}{ACC $\uparrow$} \\
 &  &  &  &  & 1h & 2h & 3h & 4h & 5h & 6h & 1h & 2h & 3h & 4h & 5h & 6h \\
\midrule
 &  &  &  & Z500 & 53.90 & 48.22 & 51.46 & 56.54 & 60.16 & 74.19 & \textbf{0.999} & \textbf{0.999} & \textbf{0.999} & 0.998 & \textbf{0.998} & \textbf{0.997} \\
 &  &  &  & T850 & 0.933 & 0.897 & 0.876 & 0.899 & 0.919 & 0.970 & 0.982 & \textbf{0.984} & \textbf{0.985} & \textbf{0.984} & \textbf{0.983} & 0.981 \\
 &  &  &  & T2M & 1.385 & 1.229 & 1.153 & 1.086 & 1.127 & 1.225 & 0.962 & 0.971 & 0.974 & 0.977 & \textbf{0.976} & 0.971 \\
 &  &  &  & U10 & 1.245 & 1.193 & 1.161 & 1.176 & 1.193 & 1.249 & 0.951 & 0.955 & 0.958 & 0.956 & 0.955 & 0.951 \\
\multirow{-5}{*}{Principal-heavy} & \multirow{-5}{*}{6} & \multirow{-5}{*}{2} & \multirow{-5}{*}{\cmark} & V10 & 1.189 & 1.152 & 1.136 & 1.165 & 1.189 & 1.244 & 0.957 & 0.960 & 0.961 & 0.958 & 0.957 & 0.953 \\
\midrule
 &  &  &  & Z500 & 54.52 & 48.95 & 52.52 & 58.10 & 62.32 & 76.35 & \textbf{0.999} & \textbf{0.999} & \textbf{0.999} & 0.998 & \textbf{0.998} & \textbf{0.997} \\
 &  &  &  & T850 & 0.938 & 0.906 & 0.890 & 0.915 & 0.934 & 0.983 & 0.982 & 0.983 & 0.984 & 0.983 & 0.982 & 0.981 \\
 &  &  &  & T2M & 1.403 & 1.250 & 1.176 & 1.113 & 1.161 & 1.263 & 0.961 & 0.970 & 0.973 & 0.976 & 0.974 & 0.969 \\
 &  &  &  & U10 & 1.276 & 1.229 & 1.200 & 1.216 & 1.231 & 1.285 & 0.949 & 0.953 & 0.955 & 0.953 & 0.952 & 0.948 \\
\multirow{-5}{*}{Orthogonal-heavy} & \multirow{-5}{*}{2} & \multirow{-5}{*}{6} & \multirow{-5}{*}{\cmark} & V10 & 1.205 & 1.172 & 1.160 & 1.192 & 1.220 & 1.282 & 0.956 & 0.958 & 0.959 & 0.956 & 0.955 & 0.950 \\
\midrule
 &  &  &  & Z500 & 54.17 & 48.87 & 52.16 & 57.44 & 61.58 & 76.06 & \textbf{0.999} & \textbf{0.999} & \textbf{0.999} & 0.998 & \textbf{0.998} & \textbf{0.997} \\
 &  &  &  & T850 & 0.937 & 0.903 & 0.885 & 0.909 & 0.928 & 0.978 & 0.982 & \textbf{0.984} & 0.984 & 0.983 & \textbf{0.983} & 0.981 \\
 &  &  &  & T2M & 1.374 & 1.233 & 1.168 & 1.106 & 1.144 & 1.240 & 0.963 & 0.971 & 0.974 & 0.976 & 0.975 & 0.971 \\
 &  &  &  & U10 & 1.265 & 1.219 & 1.188 & 1.201 & 1.217 & 1.273 & 0.950 & 0.953 & 0.956 & 0.954 & 0.953 & 0.949 \\
\multirow{-5}{*}{$\rho$ only} & \multirow{-5}{*}{4} & \multirow{-5}{*}{4} & \multirow{-5}{*}{\cmark} & V10 & 1.202 & 1.165 & 1.151 & 1.180 & 1.204 & 1.264 & 0.956 & 0.959 & 0.960 & 0.957 & 0.956 & 0.951 \\
\midrule
 &  &  &  & Z500 & 54.08 & 48.76 & 51.94 & 57.06 & 61.05 & 75.45 & \textbf{0.999} & \textbf{0.999} & \textbf{0.999} & 0.998 & \textbf{0.998} & \textbf{0.997} \\
 &  &  &  & T850 & 0.934 & 0.900 & 0.882 & 0.906 & 0.925 & 0.976 & 0.982 & \textbf{0.984} & 0.984 & \textbf{0.984} & \textbf{0.983} & 0.981 \\
 &  &  &  & T2M & \textbf{1.358} & 1.219 & 1.155 & 1.093 & 1.137 & 1.235 & \textbf{0.964} & 0.971 & 0.974 & 0.977 & 0.975 & 0.971 \\
 &  &  &  & U10 & 1.256 & 1.208 & 1.176 & 1.191 & 1.208 & 1.267 & 0.950 & 0.954 & 0.957 & 0.955 & 0.954 & 0.950 \\
\multirow{-5}{*}{$\pi$ only} & \multirow{-5}{*}{4} & \multirow{-5}{*}{4} & \multirow{-5}{*}{\cmark} & V10 & 1.196 & 1.159 & 1.144 & 1.175 & 1.200 & 1.261 & 0.957 & 0.959 & 0.960 & 0.958 & 0.956 & 0.952 \\
\midrule
\rowcolor{cyan!8}
 &  &  &  & Z500 & \textbf{52.75} & \textbf{47.15} & \textbf{50.30} & \textbf{55.23} & \textbf{58.78} & \textbf{73.07} & \textbf{0.999} & \textbf{0.999} & \textbf{0.999} & \textbf{0.999} & \textbf{0.998} & \textbf{0.997} \\
\rowcolor{cyan!8}
 &  &  &  & T850 & \textbf{0.924} & \textbf{0.887} & \textbf{0.866} & \textbf{0.888} & \textbf{0.907} & \textbf{0.958} & \textbf{0.983} & \textbf{0.984} & \textbf{0.985} & \textbf{0.984} & \textbf{0.983} & \textbf{0.982} \\
\rowcolor{cyan!8}
 &  &  &  & T2M & 1.373 & \textbf{1.205} & \textbf{1.128} & \textbf{1.060} & \textbf{1.107} & \textbf{1.211} & 0.963 & \textbf{0.972} & \textbf{0.976} & \textbf{0.978} & \textbf{0.976} & \textbf{0.972} \\
\rowcolor{cyan!8}
 &  &  &  & U10 & \textbf{1.226} & \textbf{1.174} & \textbf{1.140} & \textbf{1.154} & \textbf{1.173} & \textbf{1.236} & \textbf{0.953} & \textbf{0.957} & \textbf{0.959} & \textbf{0.958} & \textbf{0.957} & \textbf{0.952} \\
\rowcolor{cyan!8}
\multirow{-5}{*}{Complete} & \multirow{-5}{*}{4} & \multirow{-5}{*}{4} & \multirow{-5}{*}{\cmark} & V10 & \textbf{1.176} & \textbf{1.136} & \textbf{1.119} & \textbf{1.146} & \textbf{1.171} & \textbf{1.234} & \textbf{0.958} & \textbf{0.961} & \textbf{0.962} & \textbf{0.960} & \textbf{0.958} & \textbf{0.954} \\
\bottomrule
\end{tabular}}
\end{table*}

\textbf{Full dual-branch ablations.}
Tables~\ref{tab:ablation_update_full_leads}
and~\ref{tab:ablation_allocation_full_leads} extend the PROLA design
ablation in Table~\ref{tab:ablation_prola} to all lead times from 1h to 6h.
The former compares principal-only, orthogonal-only, and dual-branch
configurations with and without update scaling.
The latter compares the \((r_p,r_o)=(4,4),(6,2),(2,6)\) rank splits
and the \(\rho\)-only and \(\pi\)-only scaling variants, whose definitions
are given in Appendix~\ref{app:allocation_diagnostics}.
All configurations use a total rank of eight.
The complete \(4+4\) configuration achieves the lowest RMSE across all
five variables at every reported lead time from 2h to 6h.

\begin{table*}[t]
\centering
\small
\setlength{\tabcolsep}{3pt}
\renewcommand{\arraystretch}{1.12}
\caption{\textbf{Effect of the total trainable rank.} Latitude-weighted RMSE ($\downarrow$) at 3h and 6h. PROLA uses equal principal and orthogonal ranks. Best displayed values within each rank budget are boldfaced.}
\label{tab:rank_budget}
\resizebox{0.8\textwidth}{!}{%
\begin{tabular}{l|rrrrr|rrrrr}
\toprule
Method & \multicolumn{5}{c|}{\textit{3h lead}} & \multicolumn{5}{c}{\textit{6h lead}} \\
 & Z500 & T850 & T2M & U10 & V10 & Z500 & T850 & T2M & U10 & V10 \\
\midrule
\rowcolor{gray!12}\multicolumn{11}{l}{\textit{Total rank $r=8$}} \\
PiSSA & 53.66 & 0.894 & 1.193 & 1.209 & 1.178 & 78.55 & 0.987 & 1.248 & 1.294 & 1.290 \\
OPLoRA & 54.54 & 0.899 & 1.197 & 1.216 & 1.188 & 79.83 & 0.994 & 1.272 & 1.314 & 1.314 \\
\rowcolor{cyan!8}
PROLA (4+4) & \textbf{50.30} & \textbf{0.866} & \textbf{1.128} & \textbf{1.140} & \textbf{1.119} & \textbf{73.07} & \textbf{0.958} & \textbf{1.211} & \textbf{1.236} & \textbf{1.234} \\
\midrule
\rowcolor{gray!12}\multicolumn{11}{l}{\textit{Total rank $r=128$}} \\
PiSSA & 51.35 & 0.872 & 1.124 & 1.137 & 1.112 & 74.56 & 0.962 & 1.194 & 1.248 & 1.239 \\
OPLoRA & 57.33 & 0.938 & 1.258 & 1.272 & 1.233 & 85.79 & 1.031 & 1.332 & 1.372 & 1.373 \\
\rowcolor{cyan!8}
PROLA (64+64) & \textbf{48.19} & \textbf{0.833} & \textbf{1.072} & \textbf{1.077} & \textbf{1.063} & \textbf{68.21} & \textbf{0.933} & \textbf{1.126} & \textbf{1.182} & \textbf{1.190} \\
\bottomrule
\end{tabular}}
\end{table*}

\textbf{Effect of the total rank.}
Table~\ref{tab:rank_budget} compares PiSSA, OPLoRA, and PROLA at total
ranks 8 and 128, with PROLA using \((r_p,r_o)=(4,4)\) and \((64,64)\),
respectively.
PROLA has the lowest RMSE across all five variables at both reported
lead times within each rank budget.
Increasing PROLA's rank reduces Z500 RMSE from 50.30 to 48.19 at 3h
and from 73.07 to 68.21 at 6h, with improvements also observed for
the other four variables.

\section{Limitations and Future Work}
\label{app:limitation}
PSWD is constrained by atmospheric chaos and the limited fine-scale information available in coarse inputs. Our weather evaluation covers a global reanalysis benchmark and regional prediction tasks, including ERA5-to-CERRA transfer to a distinct target dataset. Future work could extend this evaluation to broader regions and target sources, and incorporate physical constraints and probabilistic prediction to improve forecast consistency and reliability.

\section{Broader Impacts}
\label{app:impact}
This work aims to improve the efficiency and accuracy of high-resolution weather prediction from coarse global states. More accessible spatiotemporal downscaling may benefit applications such as renewable-energy planning, flood-risk assessment, aviation safety, and local weather-risk analysis, where fine spatial and temporal information is important. However, inaccurate or overconfident forecasts could negatively affect downstream decisions if used without proper uncertainty estimates or expert oversight. For real-world use, such models should be integrated with uncertainty estimates, operational validation protocols, and expert oversight to ensure reliable and responsible decision support.

\clearpage
\section*{NeurIPS Paper Checklist}

\begin{enumerate}

\item {\bf Claims}
    \item[] Question: Do the main claims made in the abstract and introduction accurately reflect the paper's contributions and scope?
    \item[] Answer: \answerYes
    \item[] Justification: In Abstract and Introduction.

\item {\bf Limitations}
    \item[] Question: Does the paper discuss the limitations of the work performed by the authors?
    \item[] Answer: \answerYes
    \item[] Justification: In the appendix.

\item {\bf Theory assumptions and proofs}
    \item[] Question: For each theoretical result, does the paper provide the full set of assumptions and a complete (and correct) proof?
    \item[] Answer: \answerYes
    \item[] Justification: In the appendix.

    \item {\bf Experimental result reproducibility}
    \item[] Question: Does the paper fully disclose all the information needed to reproduce the main experimental results of the paper to the extent that it affects the main claims and/or conclusions of the paper (regardless of whether the code and data are provided or not)?
    \item[] Answer: \answerYes
    \item[] Justification: In Experiments and appendix.

\item {\bf Open access to data and code}
    \item[] Question: Does the paper provide open access to the data and code, with sufficient instructions to faithfully reproduce the main experimental results, as described in supplemental material?
    \item[] Answer: \answerYes
    \item[] Justification: We release the code.

\item {\bf Experimental setting/details}
    \item[] Question: Does the paper specify all the training and test details (e.g., data splits, hyperparameters, how they were chosen, type of optimizer) necessary to understand the results?
    \item[] Answer: \answerYes
    \item[] Justification: In Experiments and appendix.

\item {\bf Experiment statistical significance}
    \item[] Question: Does the paper report error bars suitably and correctly defined or other appropriate information about the statistical significance of the experiments?
    \item[] Answer: \answerYes
    \item[] Justification: We report mean and standard deviation over three seeds for the main comparisons.

\item {\bf Experiments compute resources}
    \item[] Question: For each experiment, does the paper provide sufficient information on the computer resources (type of compute workers, memory, time of execution) needed to reproduce the experiments?
    \item[] Answer: \answerYes
    \item[] Justification: In the appendix.

\item {\bf Code of ethics}
    \item[] Question: Does the research conducted in the paper conform, in every respect, with the NeurIPS Code of Ethics \url{https://neurips.cc/public/EthicsGuidelines}?
    \item[] Answer: \answerYes
    \item[] Justification: We follow NeurIPS Code of Ethics.

\item {\bf Broader impacts}
    \item[] Question: Does the paper discuss both potential positive societal impacts and negative societal impacts of the work performed?
    \item[] Answer: \answerYes
    \item[] Justification: In the appendix.

\item {\bf Safeguards}
    \item[] Question: Does the paper describe safeguards that have been put in place for responsible release of data or models that have a high risk for misuse (e.g., pre-trained language models, image generators, or scraped datasets)?
    \item[] Answer: \answerNA
    \item[] Justification: We do not release any datasets.

\item {\bf Licenses for existing assets}
    \item[] Question: Are the creators or original owners of assets (e.g., code, data, models), used in the paper, properly credited and are the license and terms of use explicitly mentioned and properly respected?
    \item[] Answer: \answerYes
    \item[] Justification: In the appendix.

\item {\bf New assets}
    \item[] Question: Are new assets introduced in the paper well documented and is the documentation provided alongside the assets?
    \item[] Answer: \answerYes
    \item[] Justification: We provide code in the supplementary material.

\item {\bf Crowdsourcing and research with human subjects}
    \item[] Question: For crowdsourcing experiments and research with human subjects, does the paper include the full text of instructions given to participants and screenshots, if applicable, as well as details about compensation (if any)?
    \item[] Answer: \answerNA
    \item[] Justification: The paper does not involve crowdsourcing nor research with human subjects.

\item {\bf Institutional review board (IRB) approvals or equivalent for research with human subjects}
    \item[] Question: Does the paper describe potential risks incurred by study participants, whether such risks were disclosed to the subjects, and whether Institutional Review Board (IRB) approvals (or an equivalent approval/review based on the requirements of your country or institution) were obtained?
    \item[] Answer: \answerNA
    \item[] Justification: The paper does not involve research with human subjects.

\item {\bf Declaration of LLM usage}
    \item[] Question: Does the paper describe the usage of LLMs if it is an important, original, or non-standard component of the core methods in this research? Note that if the LLM is used only for writing, editing, or formatting purposes and does \emph{not} impact the core methodology, scientific rigor, or originality of the research, declaration is not required.
    \item[] Answer: \answerYes
    \item[] Justification: We evaluate Qwen3.5-9B on BIRD in appendix.

\end{enumerate}

\end{document}